%% file: PCT_IJRR.tex
\documentclass[Afour,sageh,times]{sagej}

\usepackage{multirow}
\usepackage{moreverb,url}
\usepackage{wrapfig}
\usepackage{newtxmath}
\usepackage{subfigure}

\newtheorem{lemma}{Lemma}

\newtheorem{theorem}{Theorem}        

\usepackage[colorlinks,bookmarksopen,bookmarksnumbered,citecolor=red,urlcolor=red]{hyperref}

\newcommand\BibTeX{{\rmfamily B\kern-.05em \textsc{i\kern-.025em b}\kern-.08em
T\kern-.1667em\lower.7ex\hbox{E}\kern-.125emX}}

\def\volumeyear{2016}

\begin{document}

\runninghead{Smith and Wittkopf}

\definecolor{green}{rgb}{0, 0.5, 0}
\definecolor{orange}{rgb}{1.0, 0.6, 0.2}
\definecolor{red}{rgb}{1.0, 0.0, 0.0}
\definecolor{blue}{rgb}{0.0, 0.0, 1.0}
\definecolor{teal}{rgb}{0.0, 0.4, 0.4}
\definecolor{purple}{rgb}{0.65,0,0.65}
\definecolor{saffron}{rgb}{0.95,0.75,0.2}
\definecolor{turquoise}{rgb}{0.0,0.5,0.5}
\definecolor{black}{rgb}{0,0,0}
\newcommand{\zh}[1]{{\color{orange}#1}}
\newcommand{\kx}[1]{{\color{red}#1}}
\newcommand{\rev}[1]{{\color{black}#1}}

\title{Deliberate Planning of 3D Bin Packing on Packing Configuration Trees
    }

\author{Hang Zhao\affilnum{2}, Juzhan Xu\affilnum{3}, Kexiong Yu\affilnum{1},  Ruizhen Hu\affilnum{3}, Chenyang Zhu\affilnum{1}, Bo  Du\affilnum{2}, and Kai Xu\affilnum{1,4}}

\affiliation{\affilnum{1}National University of Defense Technology, China, 
\affilnum{2}Wuhan University, China, 
\affilnum{3}Shenzhen University, China,
\affilnum{4}Xiangjiang Laboratory,  China
}

\corrauth{Kai Xu, 
National University of Defense Technology. China.
}

\email{kevin.kai.xu@gmail.com}

\input{abstract.tex}


\maketitle

\input{intro.tex}

\input{related.tex}
\input{method.tex}

\input{result.tex}

\input{conclusion.tex}


\bibliographystyle{SageH}
\bibliography{reference}

\input{appendix.tex}

\end{document}

%% file: abstract.tex
\begin{abstract}

The online 3D Bin Packing Problem (3D-BPP) has widespread applications in industrial automation and has aroused enthusiastic research interest recently. Existing methods usually solve the problem with limited resolution of spatial discretization, and/or cannot deal with complex practical constraints well. We propose to enhance the practical applicability of online 3D-BPP via learning on a novel hierarchical representation—packing configuration tree (PCT). PCT is a full-fledged description of the state and action space of bin packing which can support packing policy learning based on deep reinforcement learning (DRL). 
The size of the packing action space is proportional to the number of leaf nodes, i.e., candidate placements, making the DRL model easy to train and well-performing even with continuous solution space. 
We further discover the potential of PCT as tree-based planners in deliberately solving packing problems of industrial significance, including large-scale packing and different variations of the BPP setting.
A recursive packing method is proposed to decompose large-scale packing into regular sub-trees while a spatial ensemble mechanism 
integrates local solutions into a global one. 
For different BPP variations with additional decision variables, such as lookahead, buffering, and offline packing, we propose a unified planning framework enabling various problem solving based on a pre-trained PCT model with no additional adaptation. 
Extensive evaluations demonstrate that our method outperforms existing online BPP baselines and is versatile in incorporating various practical constraints. Driven by PCT, the planning process excels across large-scale problems and diverse problem variations, with performance improving as the problem scales up and the decision variables grow. To verify our method, we develop a real-world packing robot for industrial warehousing, with careful designs accounting for constrained placement and transportation stability. Our packing robot operates reliably and efficiently on unprotected pallets at 9.8 seconds per box. It achieves averagely 19 boxes per pallet with 57.4\% space utilization for large-size boxes.

\end{abstract}

%% file: intro.tex
\section{Introduction}

As one of the most classic combinatorial optimization problems, the 3D bin packing problem (3D-BPP) usually refers to packing a set of cuboid-shaped items $i \in \mathcal{I}$, with sizes $s^x_i, s^y_i, s^z_i$ along $x,y,z$ axes, respectively,
into the maximum space utilization of bin $C$ with sizes $S^x, S^y, S^z$, in an axis-aligned fashion. Traditional 3D-BPP assumes that all the items to be packed are known a priori~\citep{MartelloPV00}, which is also called \textit{offline} BPP. The problem is known to be strongly NP-hard ~\citep{de2003greedy}.
However, in many real-world application scenarios, e.g., logistics or warehousing~\citep{WangH19}, the upcoming items cannot be fully observed; only the current item to be packed is observable. Packing items without the knowledge of all upcoming items is referred to as \textit{online} BPP~\citep{Seiden02}.

Due to its obvious practical usefulness, online 3D-BPP has received increasing attention recently. Given the limited knowledge, the problem cannot be solved by usual search-based methods. Different from offline 3D-BPP where the items can be placed in an arbitrary order,  online BPP must place items following their coming order, which imposes additional constraints. 
Online 3D-BPP is usually solved with either heuristic methods~\citep{ha2017online} or learning-based ones~\citep{ZhaoS0Y021}, with complementary pros and cons. Heuristic methods are generally not limited by the size of the action space, but they find difficulties in handling complex practical constraints such as packing stability.
Learning-based approaches typically outperform heuristic methods, particularly under complex constraints. However, their convergence is challenging when the action space is large, which limits their applicability due to the restricted resolution of spatial discretization  \citep{ZhaoS0Y021}.

We propose to enhance learning-based online 3D-BPP towards practical applicability through learning with a novel hierarchical representation–\emph{packing configuration tree (PCT)}. 
PCT is a dynamically growing tree where the internal nodes describe the space configurations of packed items and leaf nodes the packable placements of the current item. It is a full-fledged description of the state and action space of bin packing which can support packing policy learning based on deep reinforcement learning (DRL). We extract state features from PCT using graph attention networks \citep{VelickovicCCRLB18} which encode the spatial relations of all configuration nodes. The state feature is input into the actor and critic networks of the DRL model. The actor network, designed based on a pointer mechanism, weighs the leaf nodes and outputs the final placement.

During training, PCT grows under the guidance of heuristics such as \emph{corner point}~\citep{MartelloPV00}, \emph{extreme point}~\citep{CrainicPT08}, and \emph{empty maximal space}~\citep{ha2017online}. 
Although PCT is expanded with heuristic rules, confining the solution space to what the heuristics could explore, our DRL model learns a discriminant fitness function (the actor network) for the candidate placements, resulting in an effective and robust packing policy exceeding the heuristic methods.
 Furthermore, the size of the packing action space is proportional to the number of leaf nodes, making the DRL model easy to train and well-performing even with continuous solution space.

PCT was published in  ICLR 2022 \citep{zhao2022learning}, which is the first learning-based method that successfully solves online 3D-BPP with continuous solution space and achieves strong performance.
We believe that its potential extends beyond regular online packing and further discover its capability as tree-based planners to deliberately solve packing problems of industrial significance, including large-scale packing and different variations of BPP setting.
We propose \emph{recursive packing} to decompose the tree structure of large-scale online 3D-BPP as regular sub-trees to individually solve them. 
The obtained local solutions are then integrated into a global one through an effective \emph{spatial ensemble} method.
In addition to online packing, PCT's enhanced representation of packing constraints and flexible action space can be extended to other mainstream BPP settings, such as lookahead, buffering, and offline packing. 
We propose a unified planning framework enabling various problem solving based on a pre-trained PCT model with no additional adaptation.

We establish a real-world packing robot in an industrial warehouse, carefully designed to meet \emph{constrained placement}~\citep{choset2005principles} and \emph{transportation stability}~\citep{hof2005condition} requirements.
Unlike laboratory setups with protective container walls~\citep{yang2021packerbot, xu2023neural}, our system operates under industrial standards with boxes (items) directly placed onto unprotected pallets. 
Even minor robot-object collisions during placement can destabilize the static stack, and dynamic transportation by Automated Guided Vehicles (AGVs) or human workers may further challenge the stack's stability.
To satisfy constrained placement, our system incorporates a modular end-effector capable of actively adjusting its shape to maximize gripping force while minimizing collision risk. 
To ensure transportation stability, we perform physics-based verification via test-time simulation to account for real-world uncertainties. Each placement is evaluated under multiple sets of disturbances, with the simulation accelerated by GPU-based batch parallelism~\citep{makoviychuk2021isaac}.
Combined with an asynchronous decision-making pipeline that overlaps decision time with robot execution, our packing robot operates efficiently and reliably on unprotected pallets in industrial settings,  with a cycle time of 9.8 seconds per box and averagely 19 boxes per pallet (57.4\% space utilization for large boxes).
A dynamic packing video is provided at  \href{https://drive.google.com/file/d/18yhruSGh5Jj7e4Zqr06BgeUxOh3cWEwq/view?usp=sharing}{URL}.

Our works make the following contributions (those which are newly introduced in this paper are marked with the bullet symbol of `*'):
\begin{itemize}
    \item We propose a full-fledged tree description 
    for  online 3D-BPP, 
    which further enables efficient packing policy learning based on DRL.
    \item PCT is the first learning-based method that successfully solves online 3D-BPP with continuous solution space, achieving state-of-the-art performance.
    \item[*] We propose recursive packing to decompose large-scale packing problems and spatial ensemble to integrate local solutions into a global one.
    \item[*] \rev{We propose a unified planning framework to solve different BPP variations, based on a pre-trained PCT model with no additional adaptation.}
    \item[*] We develop an industrial packing robot that meets constrained placement and transportation stability,  operating reliably on standard unprotected pallets.
\end{itemize}

%% file: related.tex
\section{Related Work}
\label{gen_inst}

\subsection{3D Bin Packing Problems}\quad
Given a single bin $C$ and a set of items \(\mathcal{I}\), the objective of 3D-BPP~\citep{MartelloPV00}  is to maximize the space utilization. Its basic constraints can be formulated as follows:
\begin{align}
\text{Maximize: } & \quad \sum_{i=1}^N v_i \quad  \quad v_i = s^x_i \cdot s^y_i \cdot s^z_i, \\ 
\text{Subject to: } & \quad
p_i^d + s_i^d \leq p_j^d + S^d (1-e_{ij}^d), \quad \forall i \neq j, 
\label{eq:non-overlapping} \\ 
& \quad 0 \leq p_i^d \leq S^d - s_i^d, 
\label{eq:containment} 
\end{align}
where \(p_i\) denotes the Front-Left-Bottom (FLB) coordinate of item \(i \in \mathcal{I}\), and \(N\) is the total number of items after packing. The variable \(d \in \{x, y, z\}\) represents the axis. If item \(i\) is placed before item \(j\) along axis \(d\), the value of \(e_{ij}^d\) is 1; otherwise, it is 0. Equations~(\ref{eq:non-overlapping}) and (\ref{eq:containment}) represent the non-overlapping constraint and containment constraint~\citep{MartelloPV00}, respectively.

The early interest in 3D-BPP mainly focused on its offline setting.
Offline 3D-BPP assumes that all items are known a priori and can be placed in an arbitrary order.  
\citet{MartelloPV00} first solved this problem with an exact branch-and-bound approach. 
Limited by the exponential worst-case complexity of exact approaches, lots of heuristic and meta-heuristic algorithms are proposed to get an approximate solution quickly, such as guided local search \citep{FaroePZ03}, tabu search \citep{CrainicPT09}, and hybrid genetic algorithm \citep{KangMW12}. 
\citet{hu2017solving} decompose the offline 3D-BPP into packing order decisions and online placement decisions. This two-step fashion is widely accepted and followed by~\citet{DuanHQGZWX19},~\citet{HuXCG0020}, and~\citet{Attend2Pack}.

Although offline 3D-BPP has been well studied, their search-based approaches cannot be directly transferred to the online setting.
As a result, many heuristic methods have been proposed.
For reasons of simplicity and good performance, the deep-bottom-left (DBL) heuristic~\citep{KarabulutI04} has long been the preferred choice. 
\citet{ha2017online} sort the empty spaces with this DBL order and place the item into the first fit one.
\citet{WangH19a} propose a Heightmap-Minimization method to minimize the volume increase of the packed items as observed from the loading direction.
\citet{HuXCG0020} optimize the empty spaces available for the packing future with a Maximize-Accessible-Convex-Space method.  

\subsection{Learning-based Online Packing}\quad
The heuristic methods are intuitive to implement and can be easily applied to various online 3D-BPP scenarios. \rev{However, the price of good flexibility is that these methods perform poorly since they cannot be adapted to specific problem distributions.
In later sections, we show that even the best-performing heuristics underperform our learning-based PCT method by at least 11.5\% across all online 3D-BPP test instances.}
Designing new heuristics for specific classes of 3D-BPP is heavy work since this problem has an NP-hard solution space where many situations need to be premeditated manually by trial and error. Substantial domain knowledge is also necessary to ensure safety and reliability. To automatically generate a policy that works well on specified online 3D-BPP, \cite{Generalized} and \cite{ZhaoS0Y021} employ  DRL  to solve this problem; however,  their methods only work in discrete and small coordinate spaces. Despite their limitations, these works are soon followed for logistics robot implementation \citep{HongKL20, yang2021packerbot,zhao2021learning}.  Referring to \citet{hu2017solving}, \citet{Attend2Pack} adopt an online placement policy for offline packing needs. 
All these  methods work in a grid world with limited discretization accuracy,  which reduces their practical applicability.

PCT is the first learning-based method to successfully solve online 3D-BPP in  continuous solution space. 
Its core idea is to identify a finite set of candidates from the continuous domain and use DRL to determine the best solution. 
This candidate-based packing mechanism has been widely adopted in subsequent work. 
\cite{yuan2023towards} and \cite{pan2023adjustable}  optimize PCT policies from specific perspectives about performance variance and the worst performance.
TAP-NET++~\citep{xu2023neural} extends this approach by simultaneously calculating the attention between multiple candidates and multiple items to address buffering packing~\citep{puche2022online}.
\cite{zhao2023learning}  theoretically prove the local optimality of limited candidates for packing irregularly shaped items, using DRL to further determine the global solution.

\subsection{Practical Constraints of Industrial Packing}\quad

The majority of literature on 3D-BPP~\citep{MartelloPV00} focuses primarily on 
basic non-overlapping constraint (\ref{eq:non-overlapping})  and containment constraint (\ref{eq:containment}).
Failing to consider essential real-world constraints, such as stability~\citep{RamosOL16}, these algorithms have limited industrial applicability.
\citet{zhao2021learning} propose a fast quasi-static equilibrium estimation method tailored for DRL training and test their learned policies with real logistics boxes.
A key limitation of their approach is the use of a heightmap (the upper frontier of packed items) state representation, similar to \citet{Attend2Pack}, which overlooks the underlying constraints between packed items. 
The lack of spatial information makes the problem a partially observable Markov Decision Process (POMDP)~\citep{spaan2012partially},  which complicates DRL training and limits performance on more complex practical 3D-BPP instances involving constraints like isle friendliness and load balancing~\citep{GzaraEY20}.
PCT overcomes these limitations by explicitly storing the necessary packing configuration in a tree structure and using graph attention networks \citep{VelickovicCCRLB18} to capture their relationships. This allows PCT to better represent the packing state and enhance the performance of DRL.

Large-scale packing plays a common role in industrial production. 
For truck packing~\citep{egeblad2007fast}, hundreds of items must be packed online before long-distance transportation.
However, DRL faces challenges when applied to large-scale combinatorial optimization (CO) tasks~\citep{KoolHW19, qiu2022dimes}. On the one hand,  learning via trial and error struggles to collect sufficient samples in the enormous NP-hard space. On the other hand, long sequences of decision-making lead to learning instability~\citep{sutton2018reinforcement}. While recent studies demonstrate that graph-based neural solvers exhibit problem scale  generalizability~\citep{yu2024disco}, performance degradation is still observed due to test distribution mismatches. 
Leveraging the structured packing representation of PCT, we can decompose the large-scale packing into regular sub-problems to effectively solve them and integrate local solutions into a global one.
 This deliberate planning achieves state-of-the-art performance on large-scale packing, with performance continuing to improve as the problem scale increases.

In industrial applications, strictly online BPP~\citep{Seiden02} is not the only demand.  Packing settings along with additional decision variables usually need to be considered.
For example, lookahead packing~\citep{grove1995online} 
allows for observing upcoming items in advance for enabling better placement of the current one.  Buffering packing~\citep{puche2022online} temporarily stores incoming items in a buffer, allowing the robot to select any one of them within reach.
Offline packing~\citep{MartelloPV00, demisse2012mixed} receives complete item information from the central control system to schedule placement order.
While various solvers have been proposed recently to address these settings, they typically rely on additional parameterized  modules~\citep{hu2017solving, DuanHQGZWX19} to handle the extra constraints and decision variables, increasing training overhead and decreasing transferability.
The advantages of PCT—better constraint representation and more flexible decision-making—can also benefit these settings. 
We induce these problems into non-conflicting planning processes within a unified framework, 
achieving consistently state-of-the-art performance across various BPP settings.

%% file: method.tex
\section{Method}
\label{headings}

\begin{figure*}[t]
    \begin{center}
    \centerline{\includegraphics[width=0.96\textwidth]{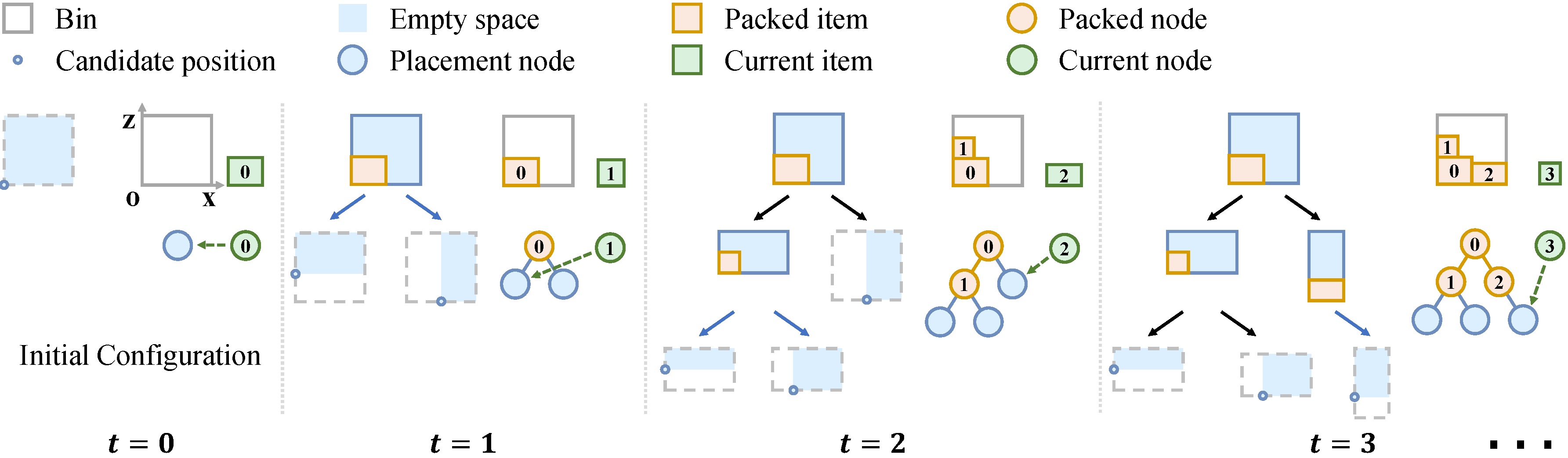}}
    \vspace{-4pt}
    \caption[font = \small]{
        PCT expansion illustrated using a 2D example (in $xoz$ plane) for simplicity, and the number of allowed orientations  $|\textbf{O}|$ is 1 (see Appendix~\ref{section:appendix_expanding} for the 3D version).
        A newly added item introduces a series of empty spaces and new candidate placements are generated, e.g., the left-bottom corner of the empty space.  
    }
    \vspace{-30pt}
\label{fig:PackingTree}
\end{center}
\end{figure*}

We begin by introducing our learning-based packing solver, PCT, and demonstrating its enhanced representation of packing constraints and flexible action space in Section~\ref{subsection:packingTree}.
In Section~\ref{subsection:markov}, we formalize PCT-based packing
as a Markov Decision Process (MDP) 
for policy learning. 
Using pre-trained PCT models, we then perform deliberate planning to solve packing problems with industrial significance: large-scale packing and different variations of the BPP setting.
In Section~\ref{subsection:recursive_packing}, we present a recursive packing method to tackle large-scale challenges, along with a spatial ensemble mechanism to integrate local solutions into a global one. \rev{In Section~\ref{subsection:uniform_packing}, we introduce a unified planning framework that solves different BPP variations with no additional adaptation.
}

\subsection{Packing Configuration Tree}
\label{subsection:packingTree}

When a rectangular item $n_t$ is added to a given packing with position $(p^x_n, p^y_n, p^z_n)$ at time step $t$, it introduces a series of new candidate positions where future items can be accommodated, as illustrated in Figure~\ref{fig:PackingTree}. 
Combined with the axis-aligned orientation $o \in \textbf{O}$ for $n_t$ based on existing positions, we get candidate placements (i.e., position and orientation).
The packing process can be seen as a placement node being replaced by a packed item node, and new candidate placement nodes are generated as children. As the packing time step $t$ goes on, these nodes are iteratively updated and a dynamic \emph{packing configuration tree} is formed,  
denoted as $\mathcal{T}$.
The internal node set $\textbf{B}_t\in\mathcal{T}_t$ represents the space configurations of packed items, and the leaf node set $\textbf{L}_t\in\mathcal{T}_t$ the packable candidate placements. During the packing, leaf nodes that are no longer feasible, e.g., covered by packed items, will be removed from $\textbf{L}_t$. When there is no packable leaf node that makes $n_t$ satisfy the constraints of placement, the packing episode ends. Without loss of generality, we stipulate a top-down packing within a single bin \citep{wang2021dense}.

Traditional 3D-BPP literature only cares about the remaining placements for accommodating the current item $n_t$, their packing policies can be written as  $\pi(\textbf{L}_t | \textbf{L}_t, n_t)$. If we want to promote this problem for practical demands, 3D-BPP needs to satisfy more complex practical constraints which also act on $\textbf{B}_t$. 
Taking packing stability for instance, a newly added item $n_t$ has possibly force and torque effects on the whole item set $\textbf{B}_t$ \citep{RamosOL16}. 
The addition of $n_t$ should make $\textbf{B}_t$ a stable spatial distribution so that more items can be added in the future. 
Therefore, our packing policy over $\textbf{L}_t$ is defined as $\pi(\textbf{L}_t|\mathcal{T}_t, n_t)$, which means 
probabilities of selecting leaf nodes from  $\textbf{L}_t$ given $\mathcal{T}_t$ and $n_t$.
For online packing, we hope to find the best leaf node selection policy to expand the PCT with more relaxed constraints so that more future items can be appended. 

\paragraph{\textbf{Leaf Node Expansion}}
The performance of online 3D-BPP policies has a strong relationship with the choice of leaf node expansion schemes, which incrementally calculate new candidate placements introduced by the just placed item $n_t$. 
A good expansion scheme should reduce the number of solutions to be explored while not missing too many feasible packings. Meanwhile, polynomials computability is also expected.
Designing such a scheme from scratch is non-trivial. Fortunately, several placement rules independent from particular packing problems have been proposed, such as \textit{Corner Point} ~\citep{MartelloPV00}, \textit{Extreme Point}~\citep{CrainicPT08}, and \textit{Empty Maximal Space}~\citep{ha2017online}. We extend these schemes, which have proven to be accurate and efficient, to our PCT expansion. Their performance will be reported in Section~\ref{subsection:action_result}.

\paragraph{\textbf{Tree Representation}}
Given the bin configuration $\mathcal{T}_t$ and the current item $n_t$, the packing policy can be parameterized as $\pi(\textbf{L}_t|\mathcal{T}_t,n_t)$. The tuple $(\mathcal{T}_t, n_t)$ can be treated as a graph and encoded by Graph Neural Networks (GNNs) \citep{1555942}. Specifically, the PCT keeps growing with time step $t$ and cannot be embedded by spectral-based approaches  \citep{BrunaZSL13} requiring a fixed graph structure. We adopt non-spectral Graph Attention Networks (GATs) \citep{VelickovicCCRLB18}, which require no graph structure priori.

The raw space configuration nodes $\textbf{B}_t, \textbf{L}_t, n_t$ are presented by descriptors in different formats.
We use three independent node-wise Multi-Layer Perceptron (MLP) blocks to project these heterogeneous descriptors into the homogeneous node features: 
${\hat{\textbf{h}}} = \{\phi_{\theta_B}(\textbf{B}_t),  \phi_{\theta_L}(\textbf{L}_t), \phi_{\theta_n}(n_t)\} \in \mathbb{R}^{d_h \times N}$, $d_h$ is the dimension of node feature and $\phi_\theta$ is an MLP block with parameters $\theta$.
The feature number $N = |\textbf{B}_t|+|\textbf{L}_t|+1$  is a  variable. The GAT layer is used to transform $\hat{\textbf{h}}$ into high-level node features. 
The Scaled Dot-Product Attention~\citep{VaswaniSPUJGKP17} is applied to each node for calculating the relation weight of one node to another.  
These relation weights are normalized and used to compute the linear combination of features $\hat{\textbf{h}}$. The  feature of node $i$ embedded by the GAT layer can be represented as:
\vspace{-4pt}
\begin{align} 
    \text{GAT}(\hat{h}_i) = W^O \sum^N_{j=1} &softmax\left(\frac{(W^Q\hat{h}_i)^T W^K\hat{h}_j}{\sqrt{d_{k}}}\right)W^V\hat{h}_j, 
\end{align}
where $W^Q \in \mathbb{R}^{d_k \times d_h} $, $W^K \in \mathbb{R}^{d_k \times d_h} $, $W^V \in \mathbb{R}^{d_v \times d_h} $, and  $W^O \in \mathbb{R}^{d_h \times d_v} $ are projection matrices. $d_k$ and $d_v$ are dimensions of projected features. The softmax operation
normalizes the  relation weight between node $i$ and node $j$.
The initial feature $\hat{\textbf{h}}$ is embedded by a GAT layer and the skip-connection operation~\citep{VaswaniSPUJGKP17} is followed to get the final output features $\textbf{h}$:
\vspace{-4pt}
\begin{equation}  
	 \textbf{h}'  =  \hat{\textbf{h}} + \text{GAT}(\hat{\textbf{h}}), \quad \textbf{h}  = \textbf{h}' + \phi_{FF} (\textbf{h}'), 
     \label{eq:layer} 
     \vspace{-4pt}
\end{equation}
where $\phi_{FF}$ is a node-wise Feed-Forward MLP with output dimension $d_h$ 
and $\textbf{h}'$  is an intermediate variable.
\rev{Since each node needs to calculate its relation with all other nodes, the complexity of this GAT operator is $O(N^2)$.}
Equation~(\ref{eq:layer}) can be seen as an independent block and can be repeated multiple times with different parameters. 
We don't extend \text{GAT} to employ the multi-head attention mechanism~\citep{VaswaniSPUJGKP17} since we find that additional attention heads cannot help the final performance. We execute Equation~(\ref{eq:layer}) once and we set $d_v = d_k$. More implementation details are provided in  Appendix~\ref{section:appendix_para}. 

\paragraph{\textbf{Leaf Node Selection}}
Given the node features $\textbf{h}$, we need to decide the leaf node indices for accommodating the current item $n_t$. Since the leaf nodes vary as the PCT keeps growing over time step $t$, we use a pointer  mechanism~\citep{VinyalsFJ15} which is context-based attention over variable inputs 
to select a leaf node from $\textbf{L}_t$.
We still adopt Scaled Dot-Product Attention for calculating pointers, the global context feature $\bar{h}$ is aggregated by a mean operation on $\textbf{h}$: $\bar{h} = \frac{1}{N} \sum_{i = 1}^N h_i$. 
The global feature $\bar{h}$ is projected to a query $q$ by matrix $W^q \in \mathbb{R}^{d_{k} \times d_h}$ and the leaf node features $\textbf{h}_\textbf{L}$ are utilized to calculate a set of keys $k_\textbf{L}$ by $W^k \in \mathbb{R}^{d_{k} \times d_h}$. 
The compatibility  $\textbf{u}_\textbf{L}$ of the query with all keys are:
\vspace{-4pt}
\begin{equation}
    q = W^q \bar{h}, \quad k_i = W^k h_i, \quad u_{i} = \frac{q^T k_i}{\sqrt{d_{\text{k}}}}. 
\vspace{-6pt}
\end{equation}
Here $h_i$ only comes from $\textbf{h}_\textbf{L}$. The compatibility vector $\textbf{u}_\textbf{L}  \in \mathbb{R}^{|\textbf{L}_t|}$ represents the leaf node selection logits. \rev{This pointer calculation possesses a complexity of $O(|\textbf{L}_t|)$.
}
The probability distribution over the PCT leaf nodes $\textbf{L}_t$ is: 
\vspace{-4pt}
\begin{equation}\label{eqnatt}
    \pi_\theta (\textbf{L}_t|\mathcal{T}_t, n_t) = softmax\left(c_{clip}\cdot\tanh\left(u_\textbf{L}\right)\right).
\vspace{-4pt}
\end{equation}
Following \cite{BelloPL0B17}, the compatibility logits are clipped with tanh,  where the range is controlled by hyperparameter $c_{clip}$, and finally normalized by softmax. 
\rev{ We provide the computational costs of GAT and the pointer mechanism in Appendix~\ref{section:moreResults}.
}

\subsection{Markov Decision Process Formulation}
\label{subsection:markov}
The online 3D-BPP decision at time step $t$ only depends on the current tuple $(\mathcal{T}_t, n_t)$ and can be formulated as a Markov Decision Process (MDP), which is constructed with state $\mathcal{S}$, action $\mathcal{A}$, transition $\mathcal{P}$, and reward $R$. We solve this MDP with an end-to-end DRL agent. The MDP model is formulated as follows:

\paragraph{\textbf{State}}
The state $s_t$ at time step $t$ is represented as $s_t = (\mathcal{T}_{t}, n_t)$, where $\mathcal{T}_t$ consists of the internal nodes $\textbf{B}_t$ and the leaf nodes $\textbf{L}_t$. Each internal node  
$b \in \textbf{B}_t$ is a spatial configuration of size $(s_b^x, s_b^y, s_b^z)$  and coordinate $(p_b^x, p_b^y, p_b^z)$ corresponding to a packed item. The current item $n_t$ is a size tuple $(s_n^x, s_n^y, s_n^z)$. Extra properties will be appended to $b$ and $n_t$ for specific packing preferences, such as density, item category, etc. 
The descriptor for leaf node $l \in \textbf{L}_t$ is a placement vector of size $(s_o^x, s_o^y, s_o^z)$ and position coordinate $(p^x, p^y, p^z)$, where $(s_o^x, s_o^y, s_o^z)$ indicates the sizes of $n_t$ along each dimension after an axis-aligned orientation $o \in \textbf{O}$. Only the packable leaf nodes that satisfy placement constraints are provided.

\paragraph{\textbf{Action}}
The action $a_t \in \mathcal{A}$ is the index of the selected leaf node $l$, denoted as $a_t = index(l)$. The action space $\mathcal{A}$ has the same size as $\textbf{L}_t$.
A surge of learning-based methods \citep{ZhaoS0Y021} directly learn their policy on a grid world through discretizing the full coordinate space,
where $|\mathcal{A}|$ grows explosively with the accuracy of the discretization.
 Different from existing works, our action space solely depends on the leaf node expansion scheme and the packed items $\textbf{B}_t$. Therefore, our method can be used to solve online 3D-BPP with continuous solution space. We also find that even if only an intercepted subset $\textbf{L}_{sub} \in \textbf{L}_t$ is provided, our method can still maintain a good performance. 

\paragraph{\textbf{Transition}}
The transition $\mathcal{P}(s_{t+1}|s_t)$ is jointly determined by the current policy $\pi$ and the probability distribution of sampling items. 
Our online sequences are generated on the fly from an item set $\mathcal{I}$ in a uniform distribution. 
The transferability of our method on item sampling distributions different from the training one is discussed in  Appendix~\ref{section:moreResults}.

\paragraph{\textbf{Reward}}
Our reward function $R$ is defined as $r_t = c_{r} \cdot w_t$ once $n_t$ is inserted into PCT as an internal node successfully; otherwise, $r_t = 0$ and the packing episode ends. Here, $c_r$ is a constant and $w_t$ is the weight of $n_t$.  
The choice of $w_t$ depends on the customized needs. For simplicity and clarity, unless otherwise noted, we set $w_t$ as the volume $v_t$ of $n_t$, where $v_t= s^x_n \cdot s^y_n \cdot s^z_n$.

\paragraph{\textbf{Training Method}}
A DRL agent seeks a policy $\pi(a_t|s_t)$ to maximize the accumulated discounted reward. Our DRL agent is trained with the ACKTR method~\citep{wu2017scalable}.
The actor weighs the leaf nodes $\textbf{L}_t$ and 
outputs the policy distribution $\pi_\theta(\textbf{L}_t|\mathcal{T}_t, n_t)$. 
The critic maps the global context $\bar{h}$ into a state value prediction to predict how much accumulated discount reward the agent can get from $t$ and helps the training of the actor.
The action $a_t$ is sampled from the distribution $\pi_\theta(\textbf{L}_t|\mathcal{T}_t, n_t)$ for training, and we take the argmax of the policy for the test.

\begin{figure}
    \centering
    \includegraphics[width=0.8\linewidth]{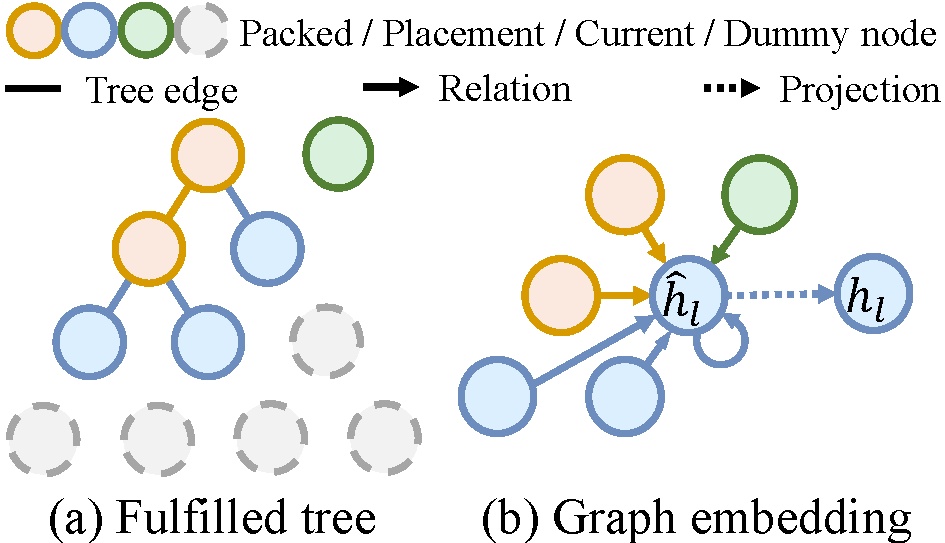}
    \vspace{-4pt}
    \caption[font = \small]{\centering   Batch calculation for PCT. 
    }
    \vspace{-12pt}
    \label{fig:fullConnect}
\end{figure}  
ACKTR runs multiple parallel processes for gathering on-policy training samples. 
The node number $N$ of each sample varies with the time step $t$ and the packing sequence of each process. For batch calculation, we fulfill PCT to a fixed length with dummy nodes, as illustrated by Figure~\ref{fig:fullConnect} (a). These redundant nodes are eliminated by \textit{masked attention}~\citep{VelickovicCCRLB18} during the feature calculation of GAT. The aggregation of $\textbf{h}$ only happens on the eligible nodes. 
For preserving node spatial relations, state $s_t$ is embedded by \text{GAT} as a fully connected graph as Figure~\ref{fig:fullConnect} (b), without any inner mask operation.
More implementation details are provided in Appendix~\ref{section:appendix_para}.

\subsection{Recursive Packing for Large-Scale BPP}
\label{subsection:recursive_packing}

The enormous NP-hard solution space and the long sequence of decision-making make learning large-scale packing policies a formidable challenge. 
We propose \emph{recursive packing}, which decomposes the large-scale $\mathcal{T}$  into a set of smaller sub-trees $\mathbf{T} = \{ \mathcal{T}^1, ..., \mathcal{T}^n \}$.
These sub-problems are solved in parallel using a pre-trained PCT model $\pi_\theta$, and the local solutions are then integrated to tackle the original $\mathcal{T}$. This approach alleviates the problem scale challenges while preserving the solution quality of $\pi_\theta$.

\begin{figure}[htp]
    \centering
    \includegraphics[width=\linewidth]{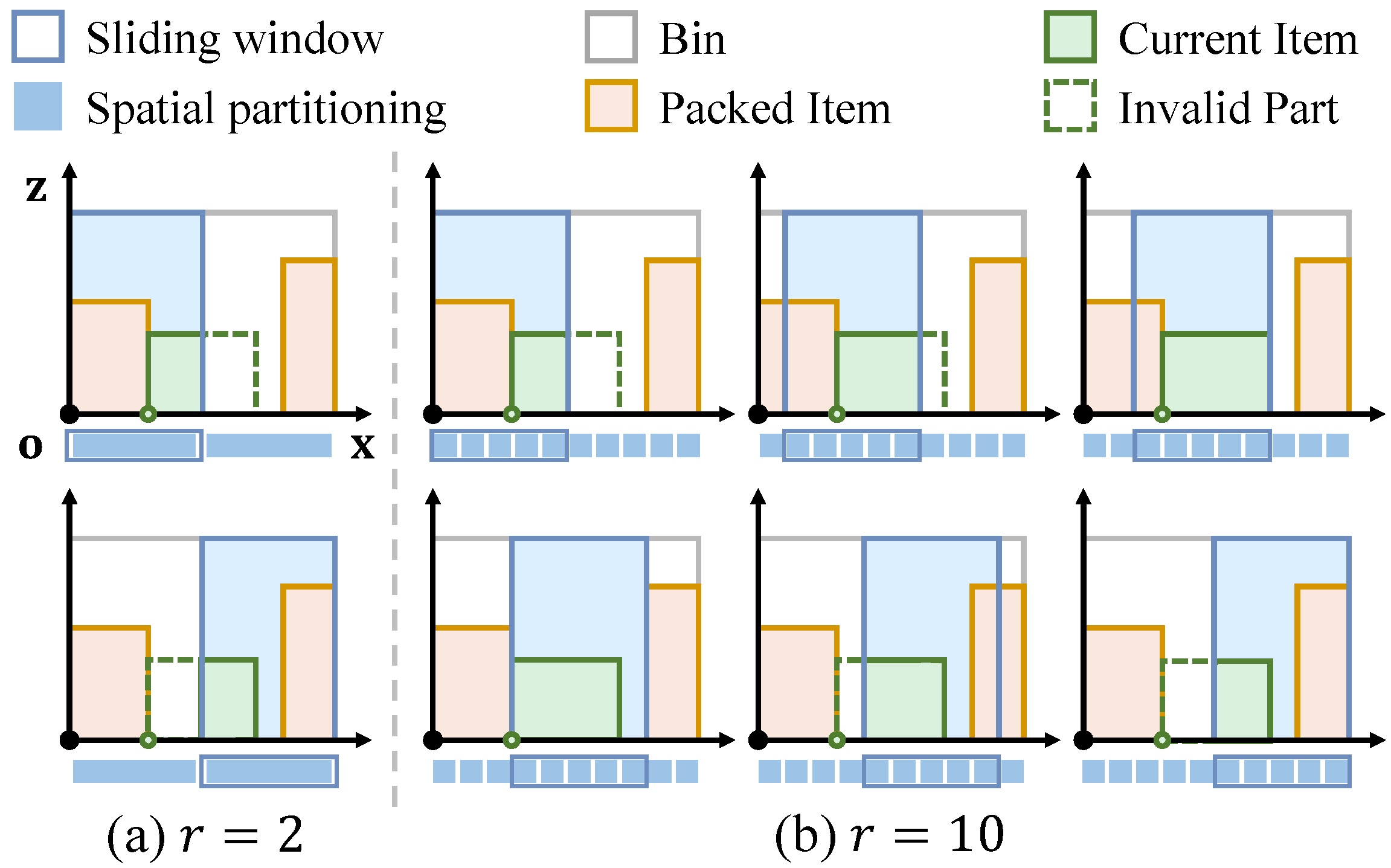}
    \vspace{-16pt}
    \caption[font = \small]{
        Problem decomposition based on sliding windows,  illustrated in the $xoz$ plane.
        Low-resolution decomposition (a) results in the loss of solutions, while fine-grained partitioning (b) clearly increases computational overhead.
    }
    \label{fig:sub_container}
\end{figure}

For problem decomposition, an intuitive approach is to 
 divide the bin \( C \) into uniform sections with resolution $r$ along each dimension, 
  and maintain a sliding window that traverses the entire bin space in a convolution-like manner,  resulting in sub-bins $\mathbf{c} = \{ c^1, ..., c^n \}$.
Each $c^i$, along with its overlapped packed items $\mathbf{B}^i$ and empty space $\mathbf{L}^i$, is treated as a sub-problem $\mathcal{T}^i$.
While this decomposition is intuitive, it's not aware of the in-bin spatial distribution,
 leading to potential degradation in solution quality. Figure~\ref{fig:sub_container} (a) gives a demonstration for this, solutions that exist in the original bin \( C \) are no longer feasible in any of the decomposed sub-bin $c^i$.
Although finer-grained partitioning can, to some extent, avoid this (Figure~\ref{fig:sub_container} (b)), it also leads to an $O(r^6)$ space complexity, significantly increasing the number of sub-problems and computational costs. 
Even more, for problems requiring decisions in continuous domains, fine-grained partitioning is inherently infeasible.

\begin{figure}[tp]
    \centering
    \includegraphics[width=0.96\linewidth]{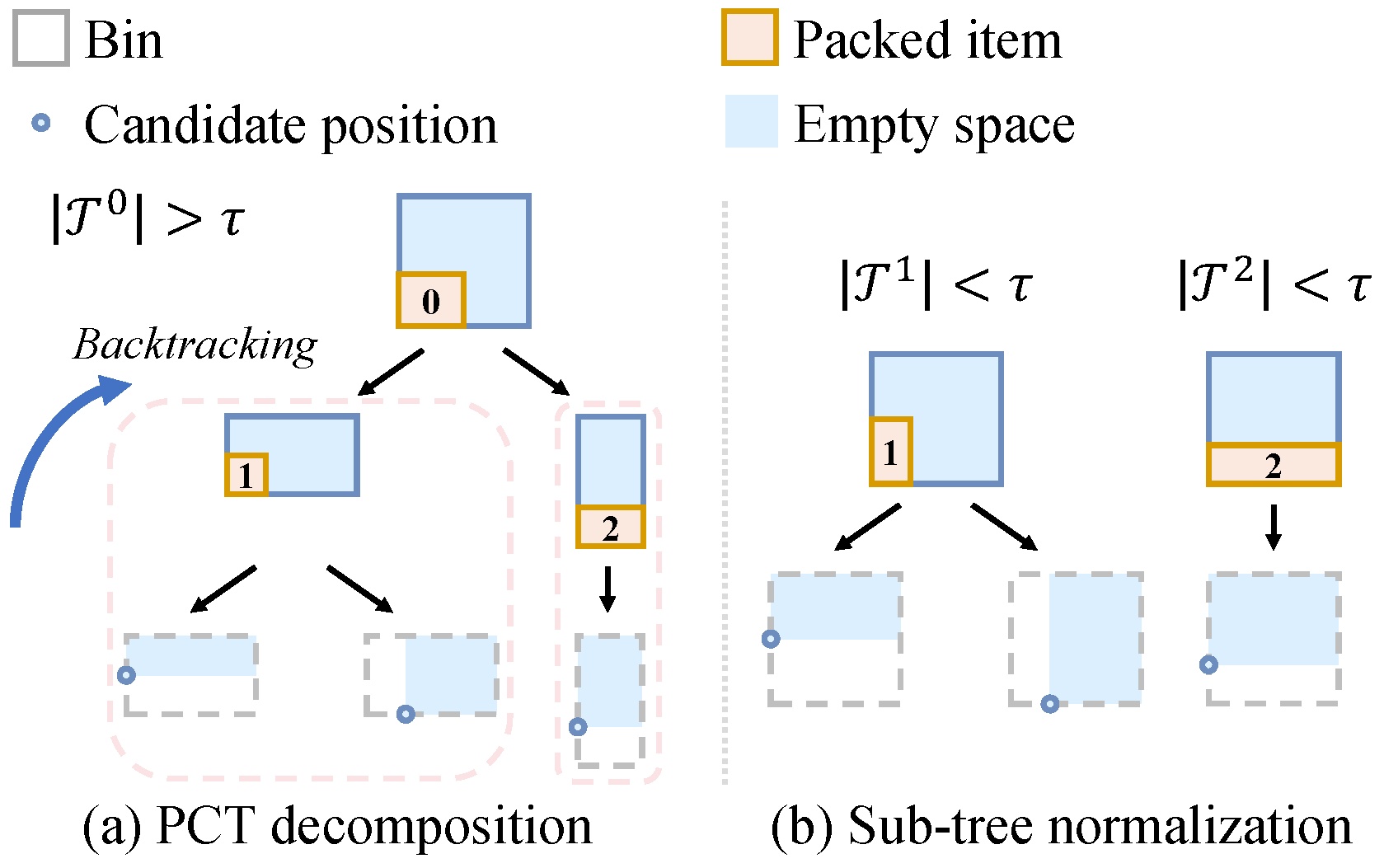}
    \vspace{-4pt}
    \caption[font = \small]{
        Recursive packing. 
        The PCT \(\mathcal{T}^0\) with item 0 as the root exceeds the given threshold \(\tau=2\). (a) We backtrack from the leaf node upwards until the maximum size of sub-trees, \(\{\mathcal{T}^1, \mathcal{T}^2\}\),  is smaller than \(\tau\). (b) Each sub-tree is normalized so that its dimensions can be mapped back to the original bin \(C\).
    }
    \label{fig:recursive_packing}
\end{figure}

However, for PCT, the decomposition of large-scale packing is natural and efficient. We adopt Empty Maximal Space (EMS), where each (previous) leaf node corresponds to a cube-shaped empty space that can be treated as a sub-bin.
Given the current item \( n \) and a large-scale \( \mathcal{T} \), the decomposition begins at a random leaf node  \( l \in \mathcal{T} \) and backtracks upward. The backtrack stops at an internal node $b^v$ whose sub-tree size $|\mathcal{T}^v|$ exceeds a set threshold $\tau$. At this point, the historical EMS $l^v$ accommodating $b^v$ can be treated as a sub-bin $c^v$. 
After sub-bin determination, we detect whether any node of $\mathcal{T}$  overlaps with \( c^v \), and these
 overlaps are inherited as internal and leaf nodes of \( \mathcal{T}^v \)   now viewed as a new sub-problem. This backtrack repeats iteratively until all leaf nodes $\mathbf{L}$ are assigned to at least one sub-tree, ensuring that all possible solutions are retained and resulting in the sub-problem set \( \mathbf{T} \).
Figure~\ref{fig:recursive_packing} (a) provides an illustration of recursive packing with $\tau = 2$.

Since PCT supports decision-making in continuous domain,  the configuration node of \( \mathcal{T}^v  \) can be normalized back to the size of the original bin \( C \), resulting in \( \hat{\mathcal{T}^v} \).
This enables the pre-trained policy $\pi_\theta$ to well adapt to sub-problems and generate high-quality solutions.
The  nodes   $b^v, l^v, n^v  \in \mathcal{T}^v $  can be normalized as:
\vspace{-4pt}
\begin{equation}
    \begin{aligned}
        \hat{b}^v &= (b^v - \text{FLB}(c^v)) \cdot S / s^v, \\
        \hat{l}^v &= (l^v - \text{FLB}(c^v)) \cdot S / s^v, \\
        \hat{n}^v &= n \cdot S / s^v,
    \end{aligned}
\vspace{-8pt}
\end{equation}
where \( S \in \mathbb{R}^3 \) represents the size of the original bin \( C \) and \( s^v  \in \mathbb{R}^3 \)  the size of sub-bin \( c^v \). Function \( \text{FLB}(c^v) \) denotes the FLB coordinates of \( c^v \). This normalization for \( \mathcal{T}^v \) is  illustrated in Figure~\ref{fig:recursive_packing} (b).

\begin{figure}[tp]
    \centering
    \includegraphics[width=0.9\linewidth]{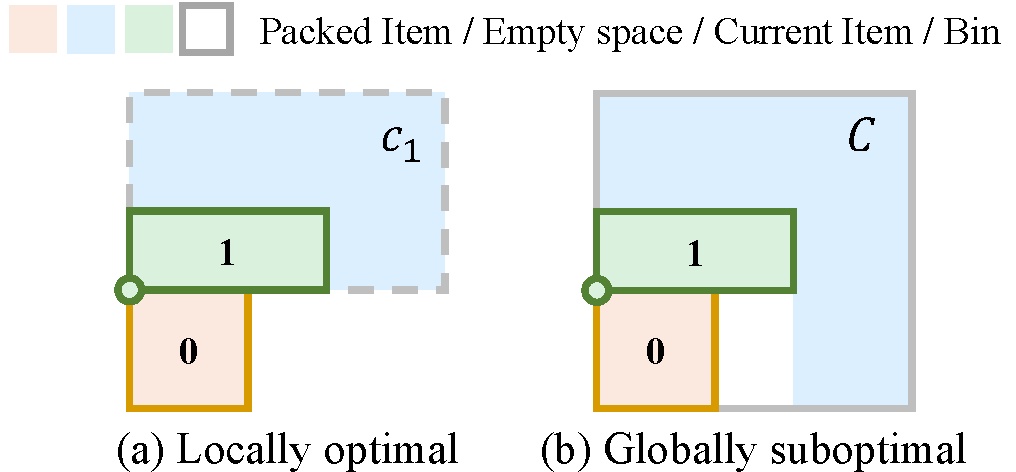}
    \vspace{-6pt}
    \caption[font = \small]{
        From the view of sub-bin $c_1$ (a), the current item 
        is optimally placed, but for the global view of bin $C$ (b), unused space beneath the item exists, which can no longer be utilized.
    }
    \label{fig:suboptimal}
\end{figure}

\paragraph{\textbf{Spatial Ensemble}}
Given sub-problems \( \mathbf{T} \),
 the solution for \( \mathcal{T}_i \in \mathbf{T} \) can be generated using a pre-trained policy \( \pi_\theta(\cdot|\hat{\mathcal{T}_i}, \hat{n}) \). 
 These solutions need to be integrated to conduct a global placement.
\cite{ZhaoS0Y021} propose evaluating placement quality in multi-bin decision scenarios utilizing the learned state value approximator 
 \( V(c_t, n_t) = \mathbb{E}\left[ \sum_{t}^{\infty} \gamma^k r_{t+k} \right] \), which captures the cumulative space utilization achievable within bin \( c \) in the future after placing the current item \( n \) in \( c \). Here \( \gamma \to  1\) is the reward discount factor.
 However, such a multi-bin decision approach
 assumes that bins are independent and do not affect each other, which obviously no longer holds in recursive packing.
 Moreover, the sub-bin only provides local placement evaluation, which does not necessarily represent global optimality. As illustrated in Figure~\ref{fig:suboptimal} (a), from the view of sub-bin \( c_1 \), the current item is compactly placed. However, from the global view of  bin \( C \)  (Figure~\ref{fig:suboptimal} (b)),  unused space exists beneath the current item, which cannot be utilized for subsequent packing due to the top-down robot packing requirement,  indicating a sub-optimal solution.

Based on the above observations, we propose \emph{spatial ensemble}, which evaluates a placement via ensembling multiple sub-bins' views.
We denote \( \Phi(l, c^i) \) as the score function for evaluating the value of leaf node \( l \) within sub-bin \( c^i \); the larger the better.
This \( \Phi \) function can be defined with any customized criteria to represent industrial preferences.
The optimal placement $l^*$  is determined by selecting the leaf node with the best worst score across all sub-bins:
\vspace{-4pt}
\begin{equation}
    l^* = \arg\max_{l \in \textbf{L}} \min_{c_i \in \mathbf{c}} \Phi(l, c_i).
   \label{eq:ensemble}
\vspace{-4pt}
\end{equation}

It is important to note that  \( \Phi(l, c_i) \) are not comparable across different sub-bins.
For example, using the state value function \( V(\cdot) \), 
the score for nearly full sub-bins approaches zero, while the score for an empty sub-bin may be close to volume  \(| C |\). This discrepancy in score ranges introduces comparison unfairness.
To avoid this, we replace  \( \Phi(l, c_i) \) from absolute values to ascending rank orders among all leaf nodes in a sub-bin \( c_i \), denoted as \( \tilde{\Phi}(l, c_i) = \operatorname{rank}_{\textbf{L}_i}(\Phi(l, c_i)) \).
The same leaf node \( l \) can appear in multiple sub-bins, and we select its worst rank as its final evaluation, with  $l^* = \arg\max_{l \in \textbf{L}} \min_{c_i \in \mathbf{c}} \tilde{\Phi}(l, c_i)$.

To maximize space utilization, we adopt the action probabilities output by the policy \( \pi(\cdot) \) as 
packing preferences \( \Phi \). 
Since items within sub-bins no longer follow a fixed size distribution, we introduce a multi-scale training mechanism. 
During training, the policy is randomly exposed to item distributions of large, medium, and small sizes. After each packing episode, the size distribution changes, so the policy must adapt to these variations during training and enhance its adaptability to unseen sizes at test time.

\subsection{Uniform Planning for BPP Variations}
\label{subsection:uniform_packing}

\begin{figure}[htp]
    \centering
    \includegraphics[width=\linewidth]{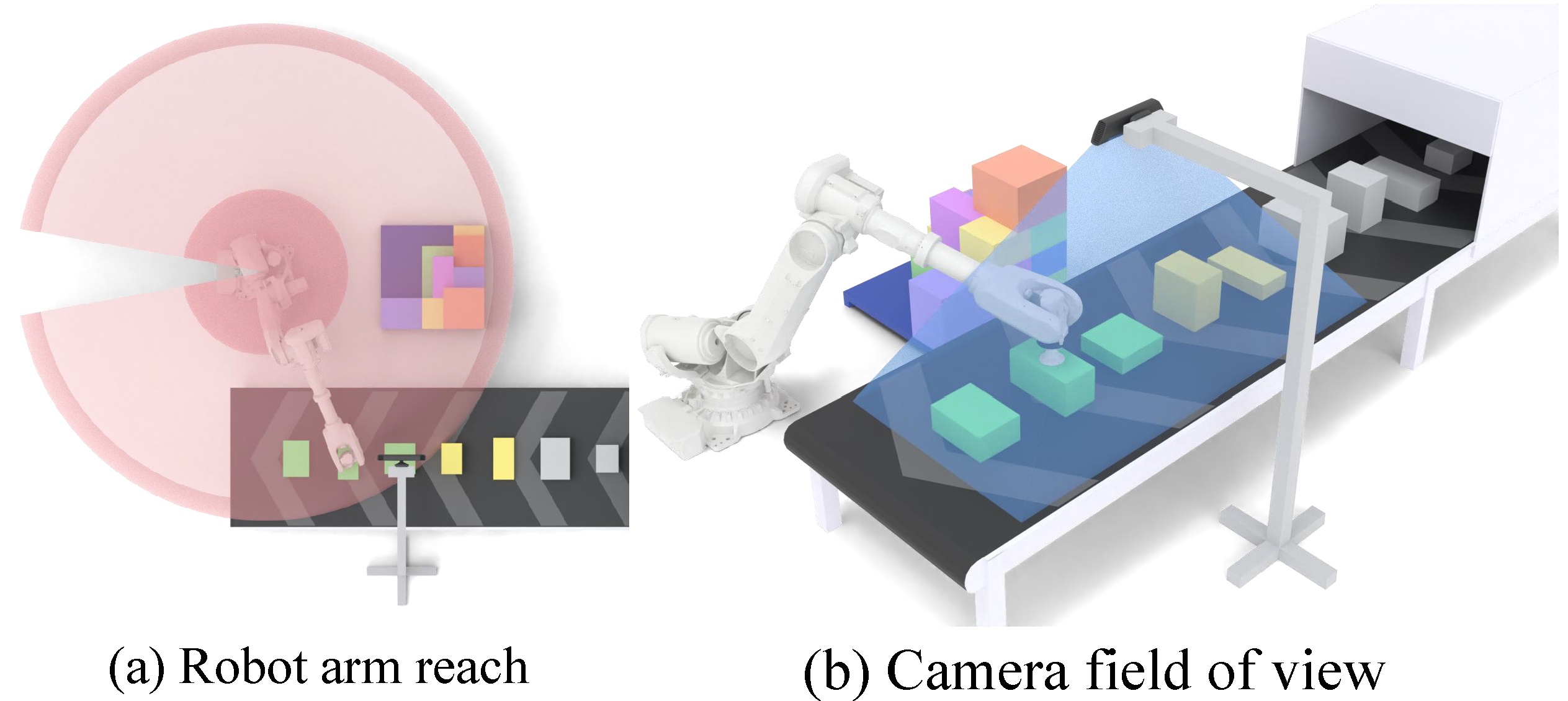}
    \vspace{-12pt}
    \caption[font = \small]{
        Item operation attributes for packing. The green items within the robotic arm's reach are selectable. Yellow items are previewed items and gray items are unknown.
    }
    \label{fig:packing_variants}
\end{figure}

Industrial packing takes various settings,  and training a dedicated model for each is difficult to transfer to the others. A unified framework for solving different BPP variations is highly desirable.
\cite{BED-BPP} provide a clear classification
of mainstream packing problems based on item operation attributes. For robot packing, items are categorized as selectable, previewed, or unknown,  determined by the camera field of view \( Fov_c \) and the robot's arm reach \( R_r \), as illustrated in Figure~\ref{fig:packing_variants}. 
Selectable items are within the robot's reach \(R_r\), as indicated by green items in the pink region of Figure~\ref{fig:packing_variants} (a).
Previewed items (yellow) are within the camera's view but outside the robot’s reach, represented by the blue cone in Figure~\ref{fig:packing_variants} (b).
Unknown items lie outside $Fov_c$ and are colored in gray.
The total item number is \( |\mathcal{I}| \), with the number of selectable, previewed, and unseen items represented by \( s = |R_r| \), \( p = |Fov_c| - |R_r| \), and \( u = |\overline{Fov_c}| \), respectively. 
A summary of the classification of mainstream packing problems is provided in Table~\ref{tab:problem_classification}.

\begin{table}[ht!]
    \caption{Mainstream packing problems.  ``Sel.'', ``Prev'',  and  ``Un.'' denote previewed, selectable,  and unseen items.}
    \label{tab:problem_classification}
    \centering
    \footnotesize
    \resizebox{0.49\textwidth}{!}{
    \begin{tabular}{l|l|l|l}
        \toprule
        Sel. &  Prev. &  Un. & \multicolumn{1}{c}{Problem} \\
       \midrule
       $s=1$  & $p=0$ &  $u>0$ & Online packing~\citep{Seiden02}  \\
       $s=1$  & $p>0$ &  $u>0$ &  Lookahead packing~\citep{grove1995online}  \\
       $s>1$  & $p=0$ &  $u>0$ & Buffering packing~\citep{puche2022online}    \\
       $s=|\mathcal{I}|$  & $p=0$ & $u=0$ & Offline packing~\citep{MartelloPV00}  \\
       \bottomrule
       
\end{tabular}
    }
\vspace{-4pt}
\end{table}

\begin{figure*}[htp]
    \centering
    \includegraphics[width=\linewidth]{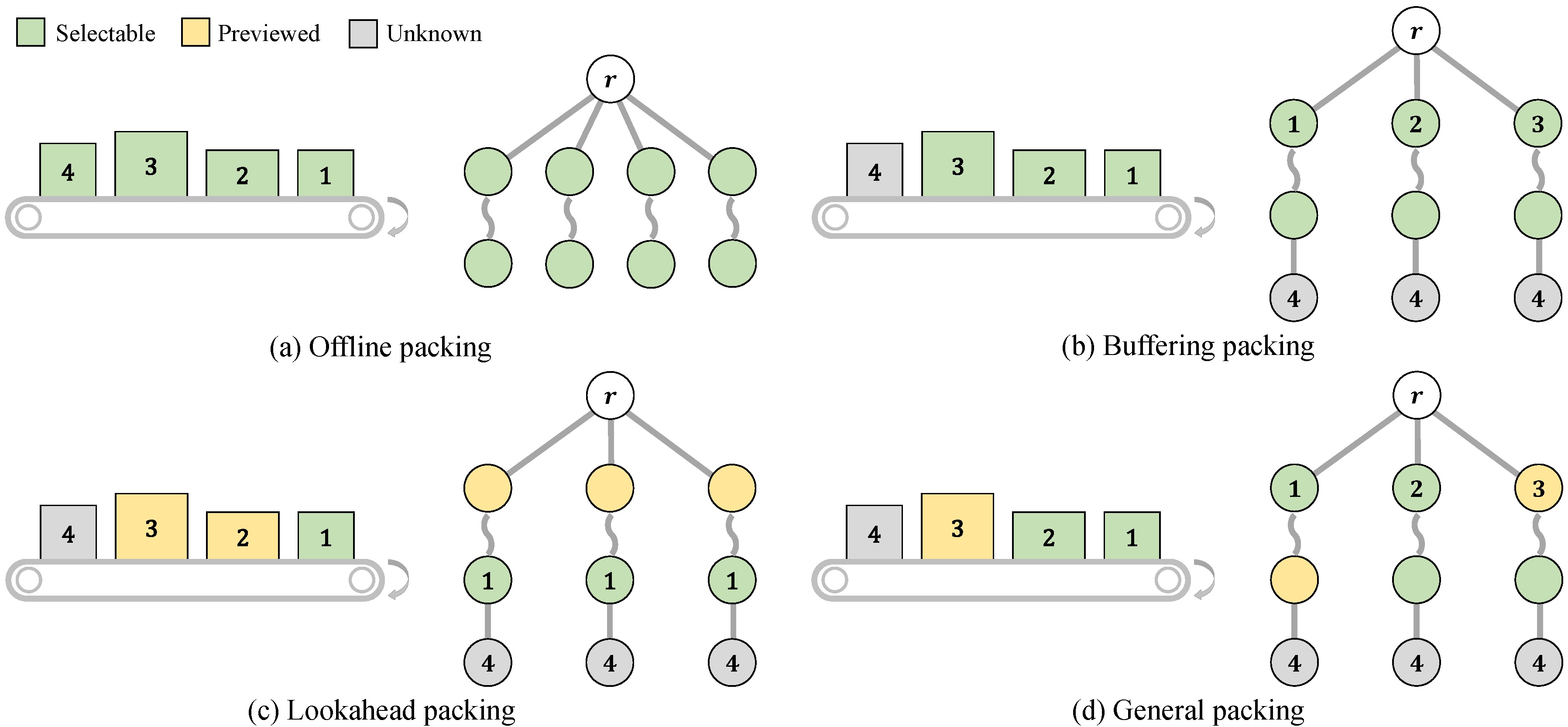}
    \vspace{-14pt}
    \caption{
        Mainstream BPP variations–offline (a),  buffering (b), and lookahead (c)–can be modeled as conflict-free search trees. \rev{We solve them 
via a unified framework (d)  powered by a pre-trained PCT model with no additional adaptation.
 }   
 }
    \vspace{-10pt}
    \label{fig:tree_variants}
\end{figure*}

A genuine problem-solving process involves the repeated use of available information to initiate exploration, which discloses, in turn, more information until a way to attain the solution is finally discovered \citep{newell1959report}.
We propose modeling various packing as model-based planning (MBP)~\citep{mayne2000constrained, silver2016mastering}, where different item operation attributes are explicitly represented as distinct planning constraints. This allows different BPP variations to be solved within a unified framework and eliminates the need to introduce or adjust decision modules.

We first formalize offline 3D-BPP, where all items are selectable, as a planning problem. 
An intuitive approach is to perform a traversal tree search over all packing orders and positions for all items \( \mathcal{I} \). Each path of the search tree represents a possible solution, 
and we choose the one with the highest accumulated space utilization \(\sum_{i = 0}^{|\mathcal{I}|} v_i \) for execution. This brute force search has a computational complexity of \( O(|\mathcal{I}|! \cdot |\mathcal{A}|^{|\mathcal{I}|}) \), where  \( |\mathcal{A}| \) represents action space size. 
Leveraging the pre-trained policy model  \( \pi_\theta \), which determines item position, the search can be simplified to only consider item order, as exhibited in Figure~\ref{fig:tree_variants} (a), lowering the complexity to \( O(|\mathcal{I}|!) \). 
The constraints for  planning selectable items  are as follows:
\begin{enumerate}
    \item Enumerate the placement order of items, with path node locations predicted by \( \pi_\theta \).
    \item     Items are placed based on the node sequence of the planned path. The planning is only conducted once. 
\end{enumerate}

Buffering packing can be considered a direct extension of offline packing, where unknown items should be additionally considered. These items cannot be explicitly included as tree nodes.
We use the state value function \( V(\cdot) \) to implicitly
estimate their distribution and future values.
 For \( s \) items within  \( Fov_c \), their placement order can still be enumerated during the planning, while items outside of \( Fov_c \) are all modeled as a single leaf node in the path end,  as shown in Figure~\ref{fig:tree_variants} (b). The value of each path is calculated by the sum of item volumes in the path and the value of the leaf node, i.e., \(\sum_{i = 0}^{s} v_i + V(\cdot) \).
 Since future arrivals exist, the planning follows the principle of MBP,
 where only the first node of the selected path is executed, and the search reinitializes when a new item arrives. 
 The additional planning constraints when introducing unknown items are summarized as:
\begin{enumerate}
    \item All unknown items are modeled as leaf nodes of the search tree with their values estimated by  \( V(\cdot) \).
    \item For each time step, only the first path node is executed, and planning restarts in the next.
\end{enumerate}

Now we discuss lookahead packing when items in range \( Fov_c - R_r \) exist.
These $p$ items can be directly incorporated in the search tree for enumeration (Figure~\ref{fig:tree_variants} (c)), accounting for future item arrival before packing the current one. The path with the highest score \( \sum_{i = 0}^{s+p} v_i + V(\cdot) \) is selected. However, previewed items cannot be really placed, introducing additional search constraints:
\begin{enumerate}
    \item Following top-down packing,  the policy \( \pi_\theta \) must not place selectable items above previewed items.
    \item For the selected path, only its first selectable node is executed, even if it starts with a previewed node.
\end{enumerate}

We can find that these planning constraints for different operation attributes are compatible and can be integrated into a unified  framework,  which can be applied to general packing scenarios where \( p \geq 0\), \( s \geq 1 \), and \( u \geq 0 \), as illustrated in Figure~\ref{fig:tree_variants} (d). 
We denote this unified search tree as  \textbf{T}ree \textbf{o}f \textbf{P}acking (\textbf{ToP}).
To ensure that the planning meets the real-time requirements of industrial packing,
we introduce Monte Carlo Tree Search (MCTS)~\citep{silver2016mastering} which reduces the time complexity of the brute-force search from \( O(|p+s|!) \) to \( O((p+s)\cdot m) \). Here \( m \) is the number of sampled paths. 
During planning, MCTS at adjacent time steps may share the same part of paths (i.e., item sequences). We maintain a global cache that stores previously visited paths to avoid redundant computations. This approach nearly halves the decision time costs.

%% file: result.tex
\section{Experiments}

In this section, we first present PCT performance combined with different leaf node expansion schemes. We then highlight the advantages of the structured packing representation, including improved node spatial relation representations and a more flexible action space.
Next, we validate the effectiveness of PCT-driven planners in solving industrial packing problems, specifically large-scale packing and different variations of the BPP setting.
Finally, we introduce our real-world packing robot in an industrial warehouse, carefully designed to meet constrained placement and transportation stability.

\paragraph{\textbf{Baselines}}
Although there are very few online packing implementations publicly available, we still do our best to collect or reproduce various online 3D-BPP algorithms, both heuristic and learning-based, from potentially relevant literature.
We help the heuristics to make pre-judgments of 
placement constraints, e.g., stability, in case of premature downtime.
The learning-based agents are trained until there are no significant performance gains.
All methods are implemented in Python and tested on 2000  instances with a desktop computer equipped with a Gold 5117 CPU and a GeForce TITAN V GPU.
We evaluate performance based on the average space utilization and the average number of packed items, denoted as ``Uti.'' and ``Num.'' , respectively. We also report the variance  ($\times 10^{-3}$) of ``Uti.'',  denoted as ``Var.''.  We include the ``Gap'' metric, which indicates the difference relative to the best ``Uti.'' across all methods.

\paragraph{\textbf{Datasets}}
Some baselines \citep{KarabulutI04, WangH19a} need to traverse the entire coordinate space to find the optimal solution, and the running costs explode as the spatial discretization accuracy increases. To ensure that all algorithms are runnable within a reasonable period, we use the discrete dataset proposed by \citet{ZhaoS0Y021} without special declaration. The bin sizes $S^d$ are set to 10 with $d \in \{x, y, z\}$, and the item sizes $s^d \in \mathbb{Z}^+$ are not greater than $S^d / 2$ to avoid over-simplification. Our performance on the continuous dataset will be reported in Section~\ref{subsection:continous}.
Considering that there are many practical scenarios of 3D-BPP, we choose three representative ones:

$Setting\,1$: Following \citet{zhao2021learning},  stability of $\textbf{B}_t$ is verified when  $n_t$ is placed. For robot manipulation convenience, only two horizontal orientations ($|\textbf{O}| = 2$) are allowed for top-down placement. 

$Setting\,2$: Following \citet{MartelloPV00}, item $n_t$ only needs to satisfy Constraints~(\ref{eq:non-overlapping})  and~(\ref{eq:containment}).  Arbitrary orientation ($|\textbf{O}| = 6$) is allowed here. This is the most common setting in the 3D-BPP literature.

$Setting\,3$: Building on $setting\,1$, each item $n_t$ is assigned an additional density property $\rho$ uniformly sampled from $(0,1]$.  This density information is incorporated into the descriptors of both $\textbf{B}_t$ and $n_t$.

\subsection{Performance of PCT Policies}
\label{subsection:action_result}

We first report the performance of  PCT combined with different leaf node expansion schemes.  Three existing schemes, which have proven to be both efficient and effective, are adopted here: Corner Point (CP), Extreme Point (EP), and Empty Maximal Space (EMS). 
These schemes are all related to boundary points of packed items $b \in \textbf{B}_t$ along the $d$ axis. 
We combine these boundary points to get the superset,
namely \textit{Event Point} (EV).
See Appendix~\ref{section:appendix_expanding} for details and learning curves.
We incorporate these schemes into our PCT model. 
Although the number of generated leaf nodes is reasonable, we only randomly intercept a subset $\textbf{L}_{sub_t} \subset \textbf{L}_t$ if $|\textbf{L}_t|$ exceeds a certain length, for saving computing resources. 
This interception length is constant during training and determined by a grid search (GS) during the test. See Appendix~\ref{section:appendix_para} for  details.
The performance comparisons are summarized in Table~\ref{tab:baselines}. 
\input{./table/performance_comparison.tex}

Although PCT grows under the guidance of heuristics, the combinations of PCT with EMS and EV still learn effective policies outperforming all baselines by a large margin across all settings. Note that the closer the space utilization is to 1, the more difficult online 3D-BPP is.
It is interesting to see that policies guided by EMS and EV even exceed the performance of the full coordinate space (FC), which is expected to be optimal. This demonstrates that a good leaf node expansion scheme reduces the complexity of packing and helps DRL agents achieve better performance.
To prove that the interception of $\textbf{L}_t$ will not harm the final performance, we train agents with full leaf nodes derived from the EV scheme (EVF), and the test performance is slightly worse than the intercepted cases. We conjecture that the interception keeps the final performance may be caused by two reasons. First, 
sub-optimal solutions for online 3D-BPP exist even in the intercepted set $\textbf{L}_{sub}$. In addition, the randomly chosen leaf nodes force the agent to make new explorations in case the policy $\pi$ falls into the local optimum. 

The performance of \citet{zhao2021learning} deteriorates quickly in $setting\,2$ and $setting\,3$  due to the multiplying orientation space and insufficient state representation separately. 
Running costs, scalability performance, behavior understanding, and visualized results can be found in Appendix~\ref{section:moreResults}.
We also repeat the same experiment as~\citet{Attend2Pack}, which packs items sampled from a pre-defined item set $|\mathcal{I}|$ = 64 in $setting\,2$. While the method of \citet{Attend2Pack} packs on average $15.6$ items and achieves $67.0\%$ space utilization, our method packs $19.7$ items with a space utilization of $83.0\%$. 
Although  EV sometimes yields better performance, we provide a detailed explanation in Appendix~\ref{section:appendix_expanding} that its computational complexity is quadratic, whereas EMS's complexity is linear to internal nodes $|\textbf{B}|$.  Therefore, we choose EMS as the default scheme.
\rev{We provide a theoretical analysis of PCT in Appendix~\ref{sec:theory}:  it selects candidates that are locally optimal in a tightness measure,  while the DRL agent further determines the global decision, improving learning efficiency and performance.
}

\subsection{Benefits of Tree Presentation}

Here we verify that the PCT representation does help online 3D-BPP tasks.
For this, we embed each space configuration node independently like PointNet~\citep{QiSMG17} to prove that the node spatial relations help the final performance.
We also deconstruct the tree structure into node sequences and embed them with  Ptr-Net~\citep{VinyalsFJ15}, which selects a member from serialized inputs, to indicate that the graph embedding fashion fits our tasks well.
We have verified that an appropriate choice of $\textbf{L}_t$ makes DRL agents easy to train, then we remove the internal nodes $\textbf{B}_t$ from $\mathcal{T}_t$, along with its spatial relations with other nodes, to prove $\textbf{B}_t$ is also a necessary part. 
We choose EV as the leaf node expansion scheme here. The comparisons are summarized in Table~\ref{tab:graph_embedding}.

\input{./table/presentation.tex}
If we ignore the spatial relations between the PCT nodes or only treat the state input as a flattened sequence, the performance of the learned policies will be severely degraded. 
The presence of $\textbf{B}$ functions more on $setting\,1$ and $setting\,3$  since $setting\,2$ allows items to be packed in any empty spaces without considering constraints with internal nodes. 
This also confirms that a complete PCT representation is essential for online 3D-BPP of practical needs.

\subsection{Performance on Continuous Dataset}
\label{subsection:continous}

The most concerning issue about online 3D-BPP is its solution space limit. Given that most learning-based methods can only work in a limited, discrete space, we directly test our method in a continuous bin with sizes $S^d =1$ to demonstrate our superiority. 
Due to the lack of public datasets for online 3D-BPP issues, we generate item sizes through a uniform distribution  $s^d\sim U(a, S^d / 2)$, where $a$ is set to 0.1 in case endless items are generated. 

\begin{wrapfigure}{r}{0.3\linewidth}
    \centering
    \vspace{-10pt}
    \includegraphics[width=1\linewidth]{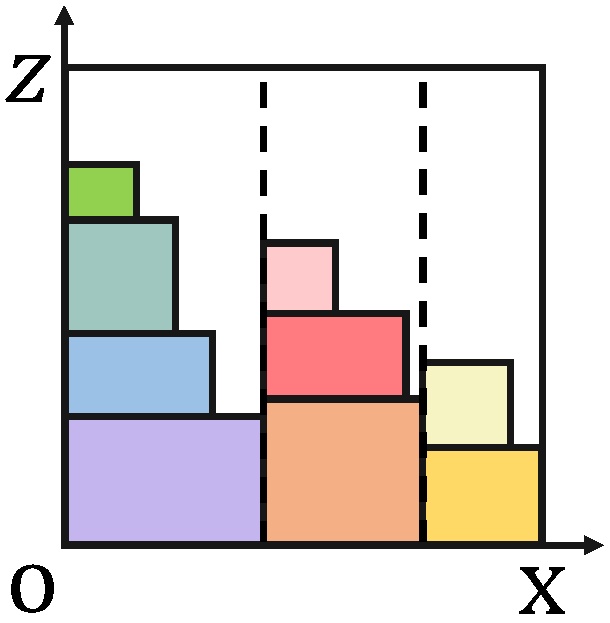}
    \vspace{-20pt}
    \label{fig:shrink_area}
\end{wrapfigure} 
Specifically, for 3D-BPP instances where stability is considered, the diversity of item size $s^z$ needs to be controlled. If all subsets of $\textbf{B}_t$ meet:
\begin{equation}
    \sum_{i \in \textbf{B}_{sub1}} s^z_i \neq \sum_{i \in \textbf{B}_{sub2}  }s^z_i, 
\end{equation}
where $\textbf{B}_{sub1} \neq \textbf{B}_{sub2},  \textbf{B}_{sub1}, \textbf{B}_{sub2} \in \textbf{B}_t$.
This means the partition problem~\citep{korf1998complete} has no solutions, and any packed items cannot form a new plane for providing support in the direction of gravity, leading to the available packing areas shrinking. 
Excessive diversity of $s^z$  will degenerate 3D-BPP  into 1D-BPP as presented in the above toy demo. To prevent this degradation from leading to the underutilization of the bins, we sample $s^z$ from a finite set $\{0.1,0.2,\dots,0.5\}$ on $setting\,1$ and $setting\,3$ in this section.

\input{./table/performance_continuous.tex}

We find that some heuristic methods like OnlineBPH~\citep{ha2017online} also have the potential to work in the continuous domain. We improve these methods as our baselines. Another intuitive approach for online packing with continuous domain is driving a DRL agent to sample actions from a Gaussian distribution (GD) and output continuous coordinates directly. 
The test results are summarized in
Table~\ref{tab:continuous}. 
Although the infinite continuous-size item set  ($|\mathcal{I}|=\infty$) increases the difficulty of the problem and reduces the performance of all methods, 
our method still performs the best among all competitors. The DRL agent which directly outputs continuous actions cannot even converge, and their variance is not considered.
Our work is the first learning-based method that solves online 3D-BPP with continuous solution space successfully.

\subsection{More Complex Practical Constraints}
\label{subsection:morecomplex}

To further demonstrate that PCT effectively handles complex constraints, we conduct experiments extending PCT to online 3D-BPP with practical constraints, including isle friendliness, load balancing, and load bearing constraints proposed by \cite{GzaraEY20}, kinematic constraints \citep{martello2007algorithm}, bridging constraints \citep{shin2016reconciling}, and height uniformity.
\rev{We denote  each constraint’s objective by $f(\cdot)$  and normalize it as 
$\hat{f}(\cdot) = f(\cdot) / \bar{f}$, where $\bar{f}$ is the average sum of constraint rewards tested under a random policy.  This 
places the sum of constraint rewards on the same level as the space utilization reward. We define the item weight as \( w_t = \text{max}(0, v_t + c \cdot \hat{f}(\cdot)) \). The max operator ensures non-negative item weights and encourages the policy to always pack as many items as possible. The constant \( c \) balances the importance of the constraints against space utilization. The constraints considered are as follows:
}

\input{./table/complex_constraints.tex}

$\bullet$ \textit{Isle Friendliness} stipulates that items of the same category should be packed as closely as possible.  The constraint reward is  $f = - \text{dist}(n_t, \textbf{B}_t)$,  where $\text{dist}(n_t, \textbf{B}_t)$ measures the average distance between $n_t$ and the items of the same category in  $\textbf{B}_t$. 
 Category information is appended to the descriptors of $\textbf{B}_t$ and $n_t$. Four item categories are tested here.

 $\bullet$ \textit{Load Balancing}  dictates that the packed items should have an even mass distribution. The constraint is \( f = - \text{var}(n_t, \textbf{B}_t) \), where \( \text{var}(n_t, \textbf{B}_t) \) is the variance of the mass distribution of the packed items on the bottom of the bin. 

$\bullet$ \textit{Load-bearing Constraint} considers that items placed above do not exert excessive weight on those below.
The force $\text{bear}(b, \textbf{B}_t)$ on each item \( b \) is simulated using physics engine~\citep{coumans2016pybullet}. The constraint  term is \( f = - \mathbb{E}_{b\in{\textbf{B}_t}}\text{bear}(b, \textbf{B}_t) \).

$\bullet$ \textit{Kinematic Constraint} minimizes the impact of placed items on the robot's subsequent motions. 
Instead of time-consuming motion planning~\citep{gorner2019moveit},
we use the safe position reward \( V_{\text{safe}} \)  \citep{zhao2021learning} as constraint, with $f =  V_{\text{safe}}$.

$\bullet$ \rev{ \textit{Bridging Constraint} requires packing items in a  staggered manner to form bridges, 
improving stability by distributing the center of gravity~\citep{page1981biaxial}.}
The number of items \( b \in \textbf{B}_t \) that contribute to  bridging $n_t$  is summed up as \( \text{bridge}(n_t, \textbf{B}_t) \), 
and the constraint reward is  \( f = \text{bridge}(n_t, \textbf{B}_t) \).

$\bullet$ \textit{Height Uniformity}  ensures even height distribution of items in the bin. 
 The constraint reward is  \( f = - H_{\text{var}} \), where \( H_{\text{var}} \) is the variance of the heightmap  after packing  $n_t$.

\rev{
The introduction of constraint rewards \(\hat{f}(\cdot) \) transforms the policy’s objective into a multi-objective optimization problem.
To ensure that space utilization remains the primary focus,  we set \( c = 0.1 \) for all constraints.
}
We adopt the learning-based method CDRL in Section~\ref{subsection:action_result}  as the baseline, as heuristic methods primarily aim to maximize space utilization and are not flexible enough to handle additional constraints. The results are summarized in Table~\ref{tab:practical}. PCT demonstrates strong adaptability to multiple complex constraints, achieving consistently higher objective scores. Its improvement over random placement (Imp.)  clearly outperforms the baseline method, demonstrating that PCT  effectively captures complex task constraints and is suitable for solving online packing in practical applications.
\rev{We conduct a parameter sensitivity analysis for \( c \) in Appendix~\ref{section:moreResults},  comparing values  of \(0.1\), \( 1.0 \), and \( 10.0 \). These correspond to cases where constraint rewards are weighted less than, the same level as, or greater than the space utilization objective, respectively.
}

\subsection{Recursive Packing for Large Problems  }

We validate the effectiveness of recursive packing for large problem scales. 
To keep the validation focusing on problem decomposition and solution integration, 
we follow the standard packing setup~\citep{MartelloPV00} of $setting\,2$.
The method is tested on continuous domains with bin dimensions set to $S^d = 1$. The packing scale $\bar{N}$  is maintained by sampling the item size from a uniform distribution $\mathcal{U}(0, (8 / \bar{N})^{\frac{1}{3}})$.
Sub-problems are decomposed recursively with a threshold $\tau$, and our spatial ensemble method integrates local solutions produced by pre-trained PCT models $\pi_\theta$.
The policy $\pi_\theta$ is exposed to multiple distributions of item sizes during training, specifically, $\mathcal{N}(0.3, 0.1^2)$, $\mathcal{N}(0.1, 0.2^2)$, and $\mathcal{N}(0.5, 0.2^2)$. 
\rev{
The principle for determining $\tau$ is to ensure that the scale of decomposed sub-problems is regular and familiar to  $\pi_\theta$.   
To this end, we generate 2000 random sequences from  $\mathcal{U}(0.1, 0.5)$ to evaluate  $\pi_\theta$, and find the 95th percentile of packed item number to be 30.4. 
This indicates that 95\% of the sequences contain no more than 30.4 items, and we set  $\tau = 30$.
}

The experiments are conducted on  $\bar{N} \in \{200, 500, 1000\}$. To date, $\bar{N} = 500$ is the largest scale for learning-based online packing.
For each $\bar{N}$, 
100 test sequences are randomly generated, and all methods are tested on the same data.   
The test results are summarized in Table~\ref{tab:large_scale}.
We compare with baselines which can operate in the continuous domain, including  BR~\citep{ZhaoS0Y021}, OnlineBPH \citep{ha2017online}, and LSAH \citep{hu2017solving}. Additionally, we evaluate the performance of the PCT model trained on the corresponding scale $\bar{N}$, labeled as PCT$^*$, and the performance of transferring a pre-trained PCT model $\pi_\theta$ to large problem scales, labeled as PCT$^\dag$.

\input{table/large_scale.tex}

As illustrated in Table~\ref{tab:large_scale}, recursive packing excels on large problem scales, achieving consistently the best performance.
Notably, as the problem scale $\bar{N}$ increases, its performance continues to improve. 
Among all tested methods, only recursive packing and LSAH exhibit such improvements as the problem scale grows, with recursive packing significantly outperforming LSAH. 
We visualize the large-scale packing results of recursive packing in Appendix~\ref{subsection:visulization}. 
For the smaller scale $\bar{N} = 200$, both PCT$^*$ and PCT$^\dag$ behave satisfactorily. 
However, as $\bar{N}$ increases,  PCT$^*$ quickly deteriorates, confirming training instability brought by long-sequence decision making and underscoring the necessity of problem decomposition. 
\rev{A computational cost analysis in Appendix~\ref{section:moreResults} also demonstrates that directly training PCT$^*$ at larger scales incurs prohibitive overhead.} 
PCT$^\dag$ performs consistently across different problem scales, verifying its generalization ability.

To validate the necessity of spatial ensemble, we compare it with other solution integration schemes.
These alternatives lack the inter-bin comparison. Instead, they directly select the sub-bin \( c \) with locally the highest score \( \Phi \):
\vspace{-4pt}
\begin{equation}
    c = \arg\max_{c_i \in \mathcal{C}} \Phi(c_i).
\vspace{-4pt}
\end{equation}
The pre-trained policy \( \pi \) then determines the placement in the selected sub-bin \( c \). 
We test the following functions \( \Phi \):

$\bullet$ \textit{Maximum State Value}: The sub-bin \( c \) with the highest state value \( V(\cdot) \) is selected to place the current item \( n \). 
A higher state value indicates higher future capacity. 
 The score function is \( \Phi(c_i) = V(\hat{\mathcal{T}}_i) \), where \( \hat{\mathcal{T}}_i \) is the normalized PCT representation within $c_i$.

 $\bullet$ \textit{Maximum Volume}: Similar to the  minimum cost priority principle \citep{dijkstra2022note}, the sub-bin \( c \) with the largest volume is selected. 
A larger volume indicates better filling of a sub-bin. 
From a divide-and-conquer perspective, effectively completing sub-tasks improves overall task performance. The score function is \( \Phi(c_i) = \sum_{\hat{b}_j \in \hat{\textbf{B}}_i} \hat{s}^x_j \cdot \hat{s}^y_j \cdot \hat{s}^z_j \), where \( \hat{s} \in \mathbb{R}^3 \) is the normalized size of node \( \hat{b} \in \hat{\textbf{B}}_i \).

$\bullet$ \textit{Maximum Return}: Inspired by the A* algorithm \citep{hart1968formal} which considers both cost and future profits, we sum up the state value and volume of a sub-bin \( c \). The score function is $\Phi(c_i) = V(\hat{\mathcal{T}}_i) + \sum_{\hat{b}_j \in \hat{\textbf{B}}_i} \hat{s}^x_j \cdot \hat{s}^y_j \cdot \hat{s}^z_j$, reflecting the total reward of placing \( n \) in  \( c \).

$\bullet$ \textit{Minimum Surface Area}: 
Unlike previous local evaluations, 
a global score function is introduced to minimize the surface area of packed items.
A smaller surface area indicates a more compact stack. 
The function is: $\Phi(c_i) = -(\tilde{S}^x_i \cdot \tilde{S}^y_i + \tilde{S}^x_i \cdot \tilde{S}^z_i + \tilde{S}^y_i \cdot \tilde{S}^z_i)$, where $\tilde{S}^d_i$ represents the maximum stack dimensions along axis $d$ after placing item $n$ in sub-bin $c_i$.

\input{table/fusion_function.tex}

The results are summarized in Table~\ref{tab:funsion_function}. 
Locally evaluating sub-bins without inter-bin comparison leads to sub-optimal performance. 
The minimum surface area alternative, while it evaluates globally, does not directly correlate with the objective of maximizing space utilization.
These results emphasize the importance of effectively integrating local solutions. In contrast, our spatial ensemble method leverages inter-bin comparison to obtain a global solution, resulting in a significant performance advantage across different $ \bar{N} $.

\begin{figure}[t!]
    \centering
    \includegraphics[width=0.98\linewidth]{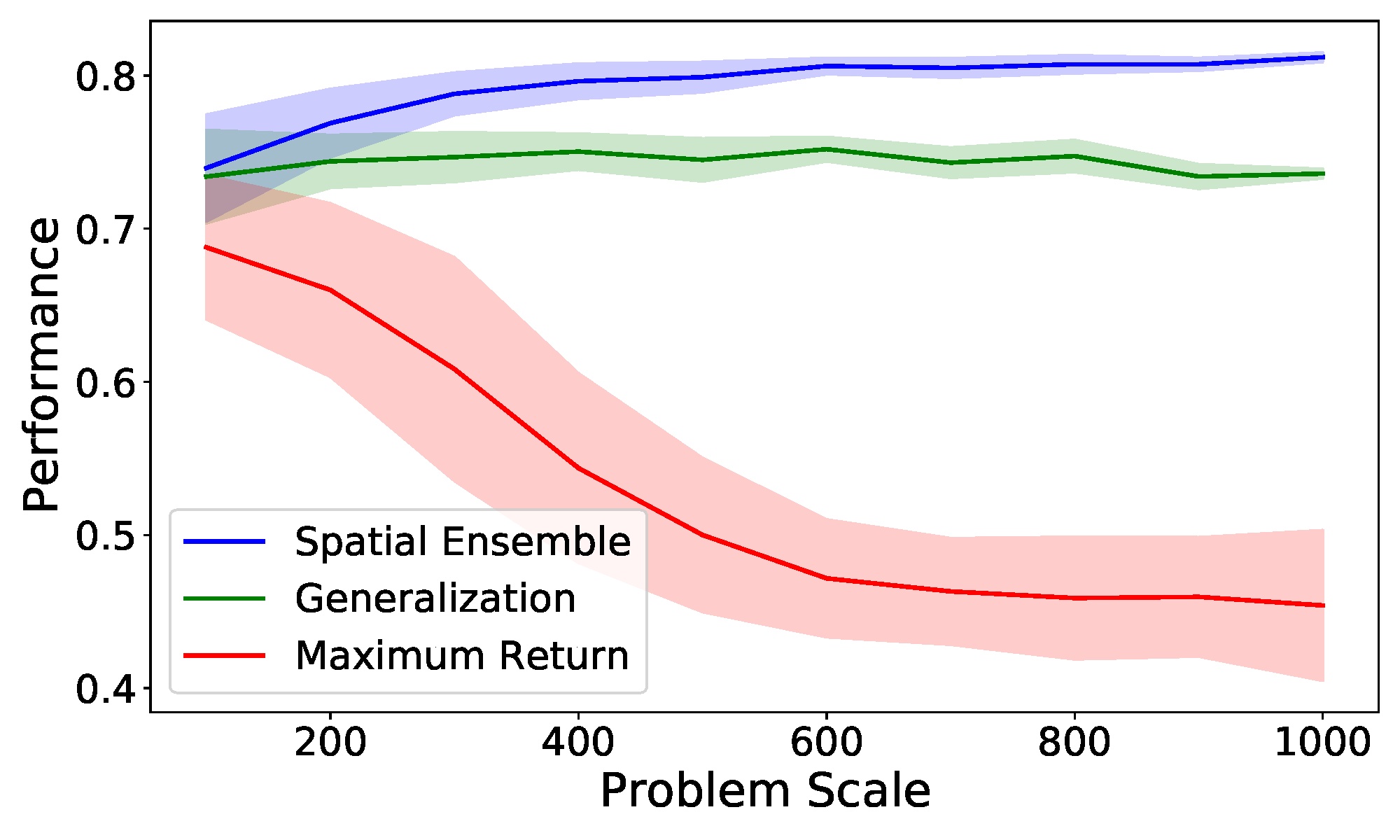}
    \vspace{-6pt}
    \caption[font = \small]{
        Packing performance across different problem scales, with shaded regions around each curve representing performance variance; the wider the larger.
    }
    \label{fig:problem_scale}
    \vspace{-4pt}
\end{figure}

\begin{figure}[t!]
    \centering
    \begin{minipage}{0.49\textwidth}
    \includegraphics[width=\linewidth]{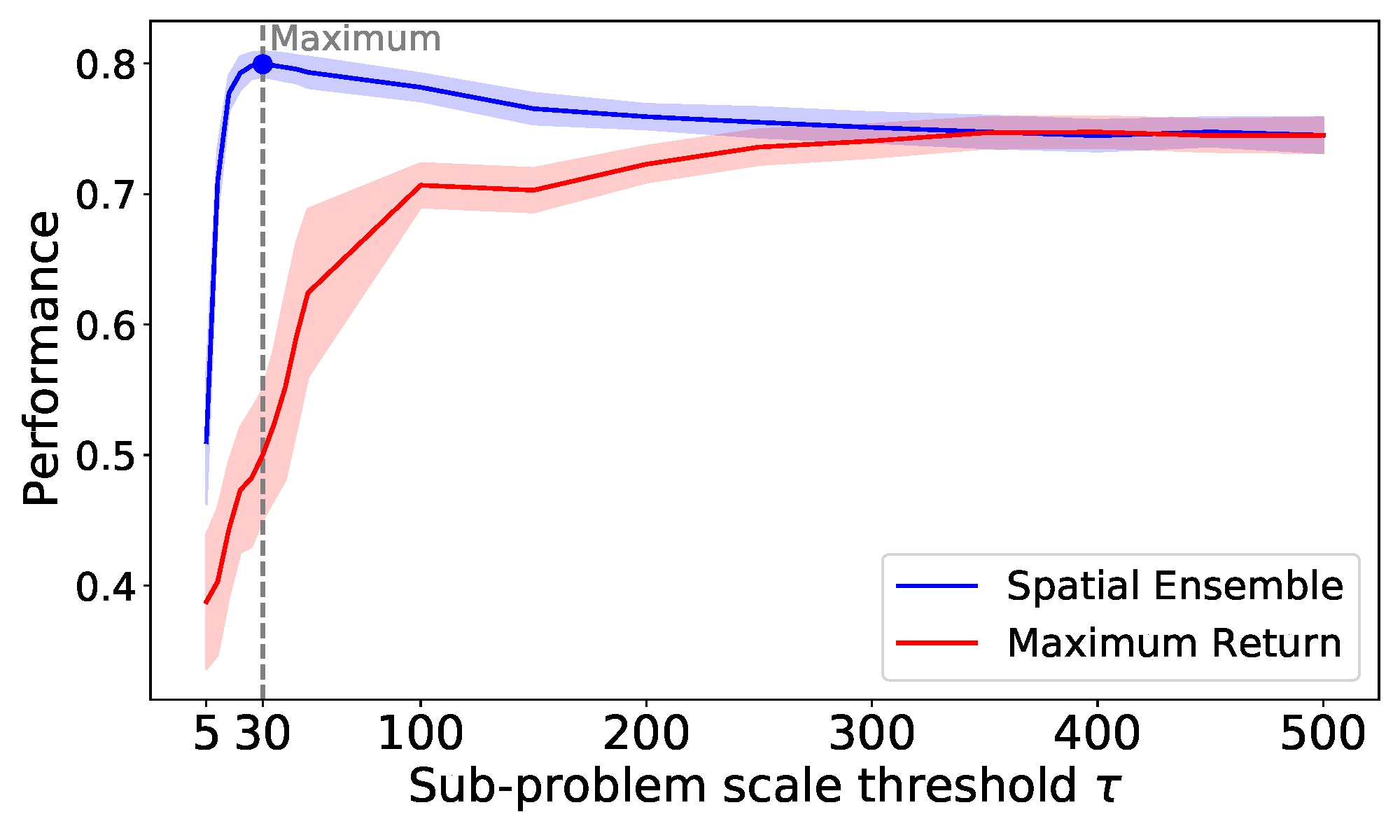}
    \vspace{-20pt}
    \caption[font = \small]{
            \rev{Packing performance for varying sub-problem scale thresholds $\tau$, tested on average problem scale $\bar{N} = 500$.  Spatial Ensumble performance peaks around $\tau = 30$,   aligning with the sub-problem scale frequently encountered by the base policy.}
    }
    \vspace{6pt}
    \label{fig:box_bound}
    \end{minipage}
    \begin{minipage}{0.48\textwidth}
        \captionof{table}{
            \rev{ Quantitative performance of recursive packing with Spatial Ensumble for solution integration 
             across different  $\tau$.  
            } 
        }
        \vspace{-10pt}
        \begin{center}
        \begin{small}
        \label{tab:tau_value}
        \setlength{\tabcolsep}{0.49em}
        \renewcommand{\arraystretch}{1}
        \resizebox{\linewidth}{!}{
            \begin{tabular}{cccccccccc}
                \toprule
                $\tau$ & 5 & 10 & 15 & 20 & 25 & 30 & 35 & 40 & 45
                 \\
                 \midrule
                 Uti. $\uparrow$ & 51.0\% & 71.0\%  & 77.7\% & 79.3\% & 79.8\% &  \textbf{79.9}\% & 79.8\% & 79.7\% &  79.5\%\\
                 \bottomrule
                 \toprule
                50 & 100 & 150 & 200 & 250 & 300 & 350 & 400 & 450 & 500
                 \\
                 \midrule
                 79.3\% & 78.2\% & 76.5\% & 75.9\% & 75.5\% & 75.1\% & 74.8\% & 74.4\% & 74.1\% & 74.5\% 
                 \\ 
                \bottomrule
                \end{tabular}
        }
        \end{small}
        \end{center}
        \end{minipage}
\end{figure}

We also make comparisons with PCT$^\dag$.
Its performance consistently behaves worse than recursive packing. We provide performance curves in  Figure~\ref{fig:problem_scale}, with wider shaded areas indicating higher variance. While recursive packing consistently improves with increasing problem scale,  PCT$^\dag$ maintains performance similar to the training scale across all problem scales. We also visualize the performance of the integration method of maximum return, which performs worse as the problem scale grows and exhibits significant variance all the time.

We explore the impact of the sub-problem decomposition threshold \( \tau \) on final performance, with experiments conducted on \( \bar{N} = 500 \).
 As observed in Figure~\ref{fig:box_bound}, increasing \( \tau \), which makes the sub-problem scale approach the original problem, causes both recursive packing and maximum return to degrade to the direct generalization performance of the pre-trained policy. The performance of recursive packing improves with finer decomposition, while the maximum return integration declines. 
 This highlights the importance of the solution integration choice.
\rev{The performance advantage persists as \( \tau \) approaches 30. Further reducing \( \tau \) causes the sub-problem scale to diverge from those encountered during $\pi_\theta$ training, leading to a gradual performance decline, as shown in Figure~\ref{fig:box_bound}. We provide quantitative results in Table~\ref{tab:tau_value} to detail the relationship between \( \tau \) and final performance.}

\subsection{ToP Results on BPP Variations}

We evaluate the effectiveness of our unified planning framework, ToP, across different BPP variations. 
We compare ToP with state-of-the-art algorithms for each setting to validate its consistent superiority. 
We first conduct comparisons on packing forms with online properties with $u > 0$.
For lookahead packing with \( s = 1, p > 0 \), we compare ToP with CDRL~\citep{zhao2021learning}. For buffering packing with \(s > 1, p = 0 \), we compare it with TAP-NET++~\citep{xu2023neural}. For general packing, where \(  s \geq 1, p \geq 0\), we adopt the O3DBP method proposed by \cite{BED-BPP}.
Since most baselines operate in discrete domains, this experiment is conducted on the discrete dataset.
Performance comparisons on the continuous-domain ICRA stacking challenge datasets are provided in Appendix~\ref{section:moreResults}.

\input{table/various_settings.tex}

\begin{figure}[tp]
    \centering
    \includegraphics[width=\linewidth]{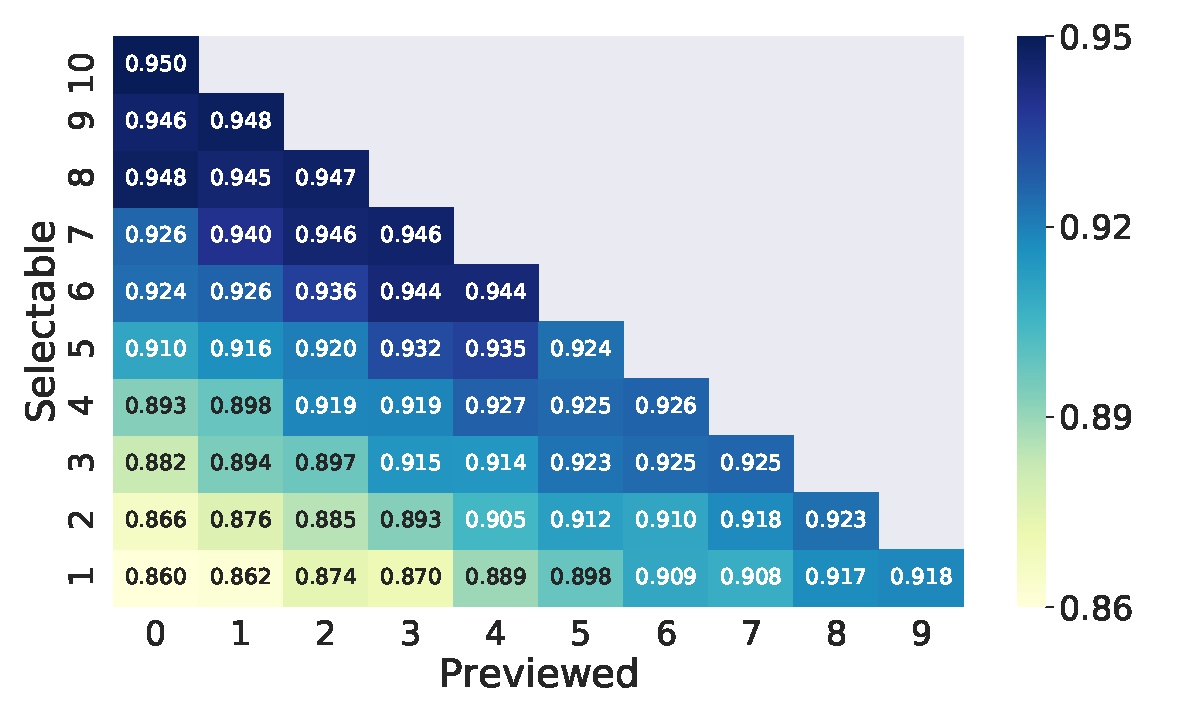}
    \vspace{-20pt}
    \caption[font = \small]{
     Asymptotic planning performance with space utilization labeled in each grid.
    }
    \vspace{-14pt}
    \label{fig:preview_select}
\end{figure}

The comparison results across different packing settings are summarized in Table~\ref{tab:various_setting}. 
ToP consistently delivers the best space utilization.
We visualize ToP performance with varying selectable item number  \( s \) and previewed item number \( p \) in Figure~\ref{fig:preview_select}, based on experiments conducted on $setting\,2$.
The heatmap clearly reveals that, as decision variables increase, the packing performance also improves.
 Visualizations of ToP results across different BPP variations can be found in Appendix~\ref{subsection:visulization}. 

\input{table/offline_setting.tex}

We also compare ToP on the widely studied offline BPP \citep{MartelloPV00}, where \(s = |\mathcal{I}| \) and \(p = u = 0 \). 
Baselines include the traditional optimization solver  
Gurobi~\citep{gurobi2018} and learning-based methods including RCQL~\citep{li2022one}, Attend2Pack~\citep{Attend2Pack}, and TAP-NET++~\citep{xu2023neural}. The total number of items  \( |\mathcal{I}| = 50 \).
Unlike online packing \( (u > 0) \), where items arrive continuously and intermediate decisions must be made, offline packing allows iterative optimization for better solutions.
To ensure fairness, each method’s decision time is capped at 600 seconds. 
The results, summarized in Table~\ref{tab:offline_setting}, illustrate that ToP outperforms all baselines. Traditional solvers often get stuck in local optima, leading to low-quality solutions within the time limit.

\begin{figure}
    \centering
    \includegraphics[width=\linewidth]{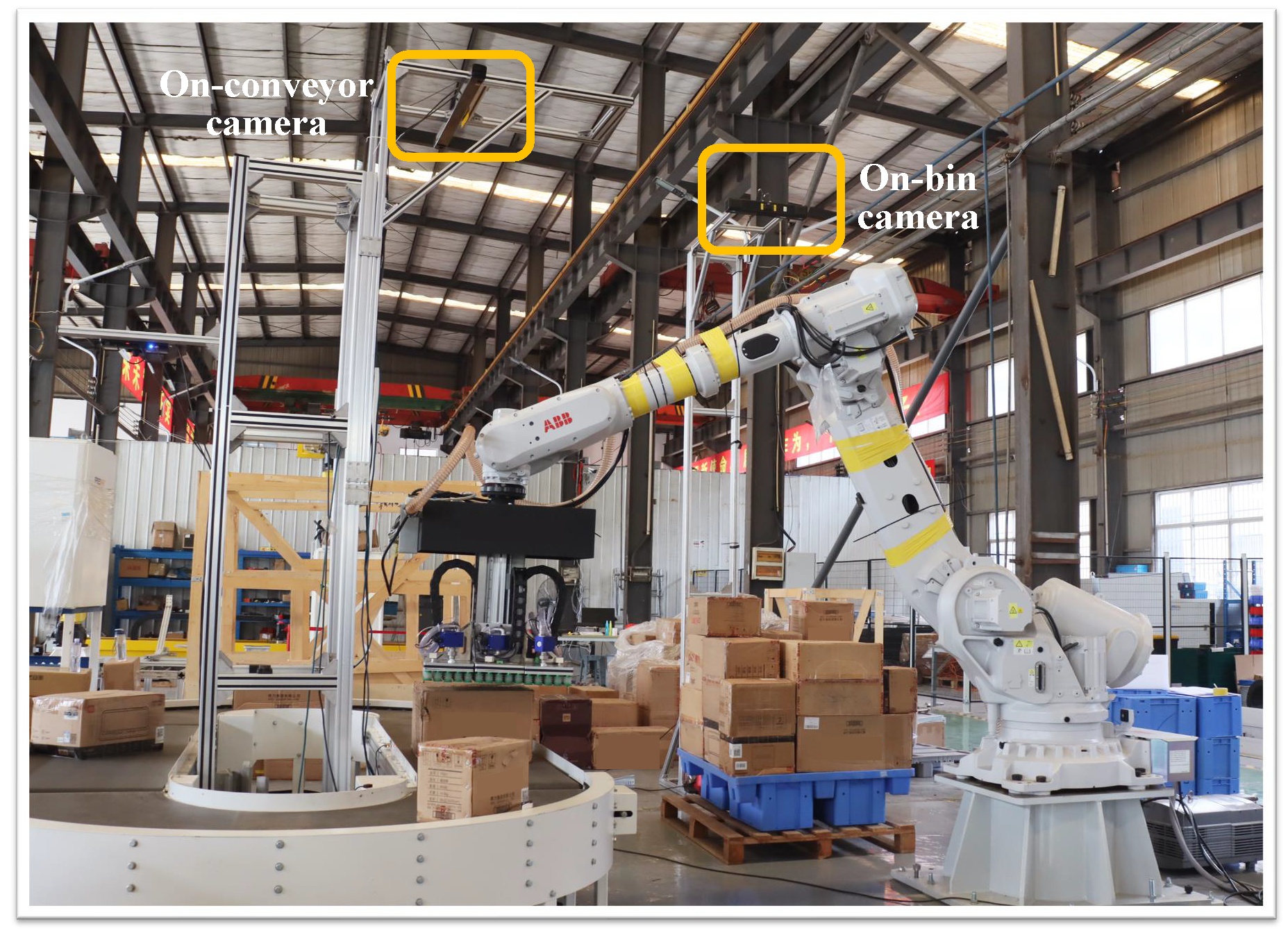}
    \vspace{-14pt}
    \caption{Our real-world packing system. The on-conveyor camera detects targets. The on-bin one monitors possible drifts.
    }
    \vspace{-14pt}
    \label{fig:realrobot}
\end{figure} 

 \subsection{Real-World  Packing Robot}  
 \label{subsection:realRobot}

We develop a real-world packing robot in an industrial warehouse, as presented in Figure \ref{fig:realrobot}. 
The system adopts an ABB IRB 6700 robotic arm with a 210kg load capacity.
Boxes (items) are delivered via a conveyor belt, which stops upon box detection by a photoelectric sensor.
A key challenge for robotic packing is safely and efficiently placing boxes into constrained spaces where objects are positioned closely. Even minor robot-object collisions can destabilize the stack, leading to serious production failure. 
Moreover, unlike laboratory packing scenarios with protective container walls~\citep{yang2021packerbot, xu2023neural}, industrial packing often omits such protection for efficiency, making the constrained placement problem~\citep{choset2005principles} more challenging.
Solving it with motion planning~\citep{gorner2019moveit} requires precise scene modeling and is time-consuming, violating practical demands.

\begin{figure*}[t]
    \centering
    \includegraphics[width=\linewidth]{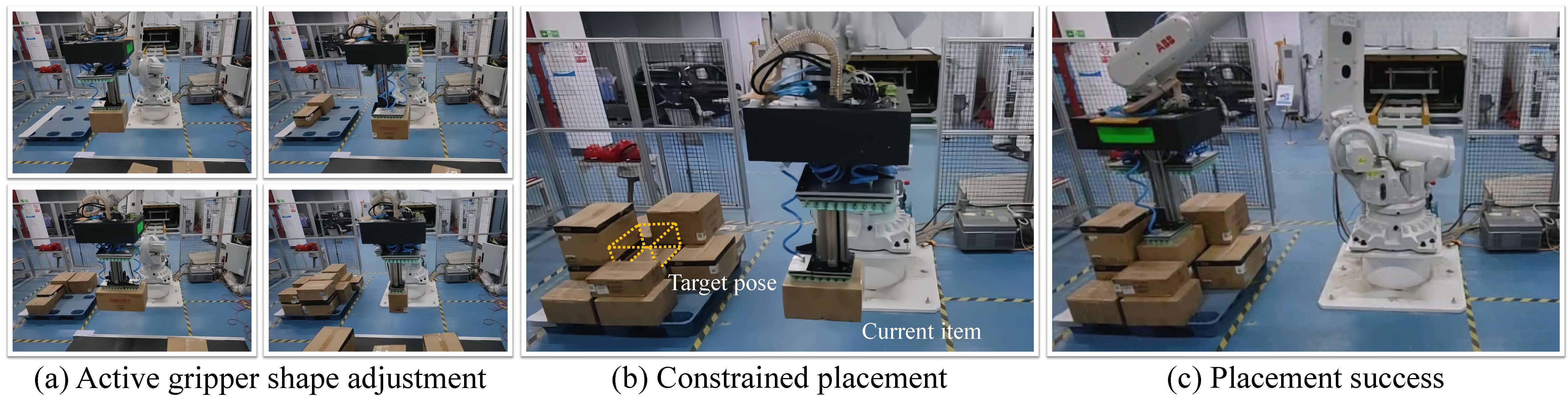}
    \vspace{-10pt}
    \caption{
    (a) The gripper actively adjusts its shape to adapt to the constrained placement requirements.
    (b) The robot grips and transfers boxes to the target pose where the surrounded boxes exist.
    Oversized end-effectors may cause collisions with placed boxes. (c) Our modular gripper dynamically adapts its shape to achieve constrained placement. 
    }
    \vspace{-10pt}
    \label{fig:constraint_placement}
\end{figure*} 

 Instead of motion planning, we adopt a simple top-down placement manner with a flexible modular gripper which can actively adjust its shape to satisfy constrained placements. 
 Each module is equipped with multiple suction cups for gripping flat items. 
 Modules vary in size, allowing the gripper to adjust its own shape based on the target box's size, as illustrated in Figure~\ref{fig:constraint_placement} (a).
 The shape adjustment principle is simplified to maximize box coverage without exceeding the box's top boundary, providing sufficient gripping force while avoiding robot-object collisions during placement, as demonstrated in Figure~\ref{fig:constraint_placement} (b) and (c).
 This ensures both efficiency and safety in industrial scenarios.
\rev{ We provide more details of our modular gripper and  its simplified motion planning in Appendix~\ref{subsection:Hardware}.}

 The warehouse stocks 8000 Stock Keeping Units (SKUs), including large-sized boxes ranging up to 80  $\times$  80 $\times$ 60cm, with a maximum weight of 30kg. 
 The unprotected pallet measures 120 $\times$ 100cm, and the maximum stack height is 140cm.
 An RGB-D camera (PhoXi 3D Scanner XL) captures the top surface of incoming boxes. 
 \rev{Box detection and segmentation are performed using Mask R-CNN \citep{HeGDG17}, achieving a 99.95\% recognition success rate for boxes on the conveyor.
An eye-to-hand calibration~\citep{dekel2020optimal}  is performed between the on-conveyor camera and the robot base to transform the top-center coordinates of the detected items into the robot's coordinate system for  grasping.
}
 The modular gripper automatically adjusts its configuration 
 while approaching the target box, ensuring no delay in the packing cycle. 
 The gripper has a 40cm lifting stroke, so surrounding boxes must not exceed 40cm above the top surface of the target placement. Placement candidates violating this constraint are discarded.

\begin{figure}
    \centering
    \includegraphics[width=\linewidth]{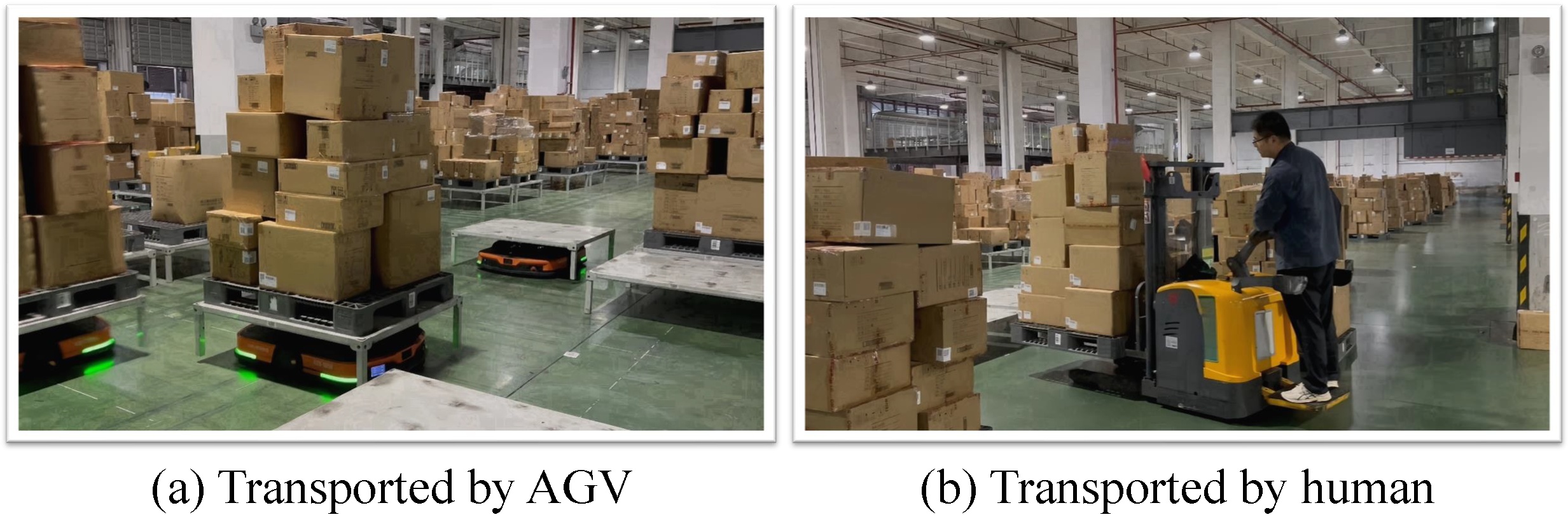}
    \vspace{-20pt}
    \caption{Transportation of packed boxes in a warehouse.}
    \label{fig:dynamic_transport}
\end{figure}

  \begin{figure}
    \centering
    \includegraphics[width=\linewidth]{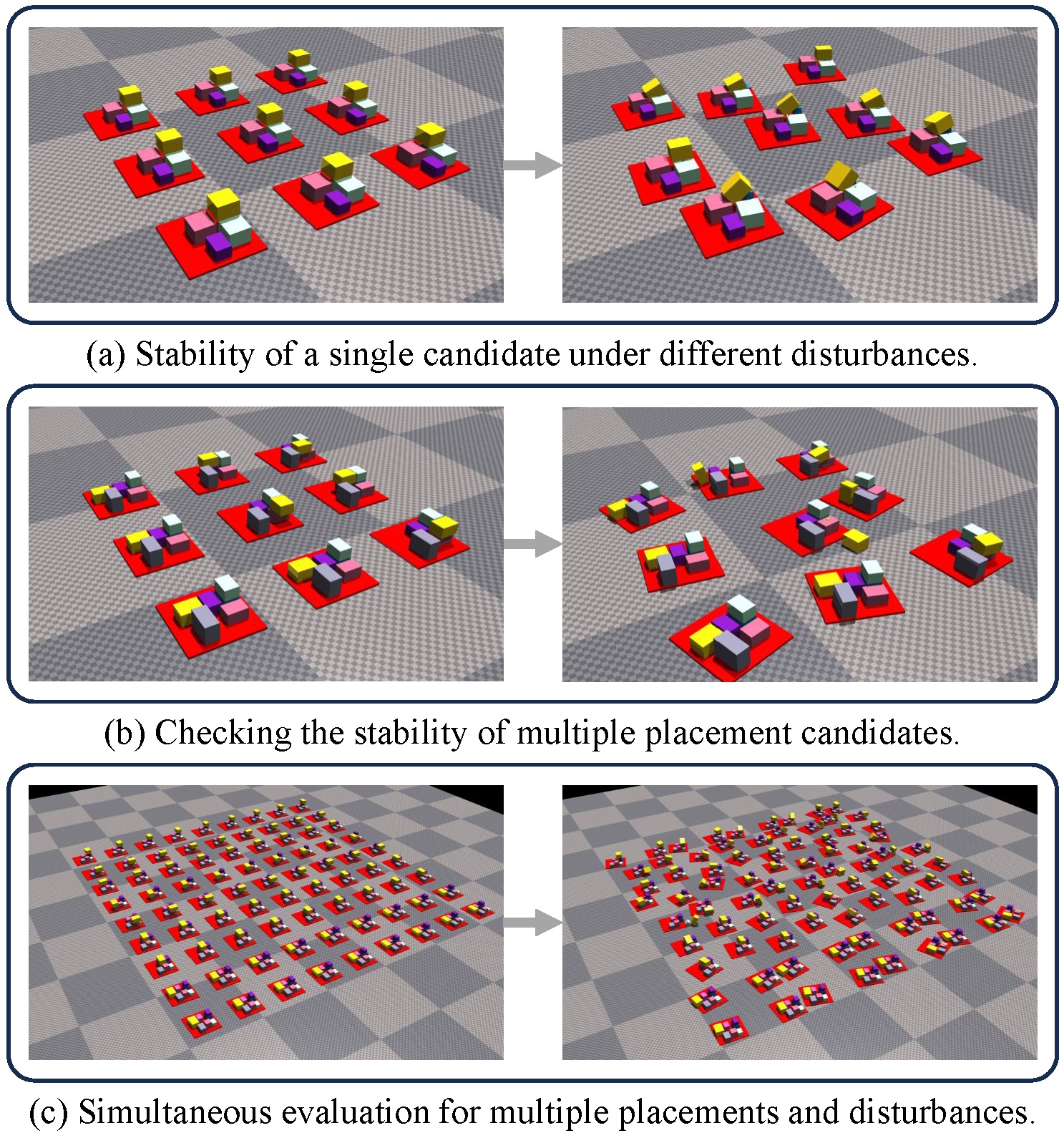}
    \vspace{-20pt}
    \caption{Simultaneously sampling multiple disturbances to simulate the real-world dynamic transportation.} 
    \vspace{-10pt}
    \label{fig:ensumble_simulation}
\end{figure} 

Aside from constrained placement, another major challenge in industrial packing lies in ensuring the stability of packed boxes during dynamic transportation~\citep{hof2005condition}. As exhibited in Figure~\ref{fig:dynamic_transport}, packed boxes are moved by Automated Guided Vehicles (AGVs) or human workers, involving passive motions like lift, acceleration/deceleration, and rotation of the stack. Most existing packing research relies on quasi-static equilibrium for stability evaluation~\citep{WangH19a, zhao2021learning}, which does not hold under dynamic conditions.

We model real-world transportation uncertainties through physics-based verification.
Ideally, if a real-world disturbance set $d$ causes stack $\mathbf{B}$ to collapse, a physical simulator $\mathcal{E}$,  as $\mathbf{B}$'s digital twin, should predict this and reject unstable placements.
However, real-world disturbances cannot be captured or recorded. To this, we randomly sample multiple disturbance sets to evaluate the stability for each placement, as illustrated in Figure~\ref{fig:ensumble_simulation} (a).  
Each disturbance set is a combination of 10 randomly generated translations and rotations, with translations in the range $[15, 20]$cm along the  $x$ and $y$ axes and rotations in the range $[-10^\circ, 10^\circ]$ for the $z$ axis.
The simulated linear and angular velocities are set to 6m/s and $30^\circ$/s, respectively. If any disturbance causes the stack to collapse, we consider a placement $l$ unacceptable and turn to its alternatives (Figure~\ref{fig:ensumble_simulation} (b)). 
To meet real-time decision-making requirements of industrial packing, we leverage the Isaac Gym simulator~\citep{makoviychuk2021isaac}, which features batch parallelism and  GPU-based simulation acceleration.
For each placement, we sample $k_d$ disturbance sets, and evaluate $k_l$ such placements simultaneously, as exhibited in Figure~\ref{fig:ensumble_simulation} (c).  We select the top $k_l$ placements based on probabilities output by policy $\pi_\theta$.

\begin{figure}
    \centering
    \includegraphics[width=\linewidth]{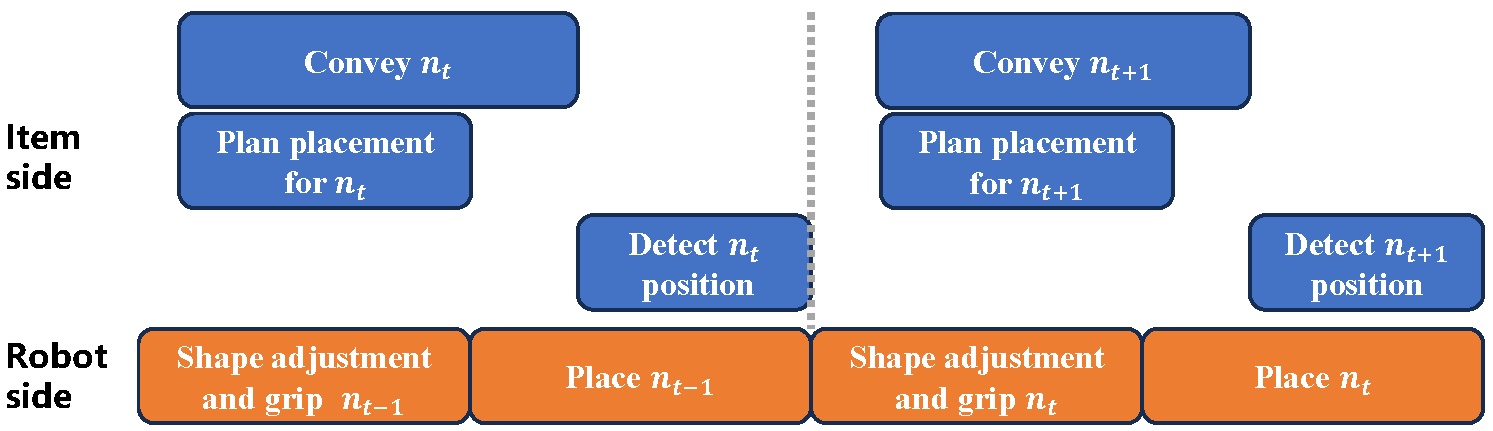}
    \caption{
        Our asynchronous packing pipeline. Blue blocks represent item-side operations, while orange blocks represent robot-side operations. This design ensures the robot is always prepared to pack the next box, maximizing efficiency.
    }
    \label{fig:time_optimization}
    \vspace{-10pt}
\end{figure}

A promising approach to leverage physics-based stability evaluation is to incorporate it as training guidance  (e.g., serving as a reward signal), making the trained policies physics-aware and eliminating test-side simulation costs. 
However, we do not adopt this approach for two practical reasons. First, Isaac Gym's parallelism offers little benefit for online packing, which requires generating new objects and dynamically modifying the simulated scene—an expensive operation due to scene data being preloaded to CUDA memory and the high cost of CPU-GPU synchronization.
Second, \cite{ZhaoS0Y021} find that purely relying on training-side constraints inevitably leads to occasional packing instability during testing. Hence, we conclude that simulation for test-time stability evaluation is always necessary, whereas training can proceed without it.
For encouraging physics awareness in the base PCT policy $\pi_\theta$, we incorporate the fast quasi-static equilibrium estimation method proposed by \citet{zhao2021learning} for training.

Even powered by batch parallelism and GPU-based acceleration, the test-time simulation inevitably increases computational costs and affects packing cycles/manufacturing throughputs.  
We propose an asynchronous decision pipeline for efficiency optimization.
As presented in Figure~\ref{fig:time_optimization}, while the robot is packing box \(n_{t-1}\), the system simultaneously prepares box \(n_t\) by transporting it on the conveyor, retrieving its dimensions from the central controller, calculating its placement and stability, and capturing RGB-D images for location detection. 
Since the robot-side execution cost cannot be optimized, we focus on minimizing the item-side computing time to ensure it is shorter than the robot execution time. This keeps the robot always ready to pack the next box, maximizing efficiency.

\begin{figure}
    \centering
    \includegraphics[width=\linewidth]{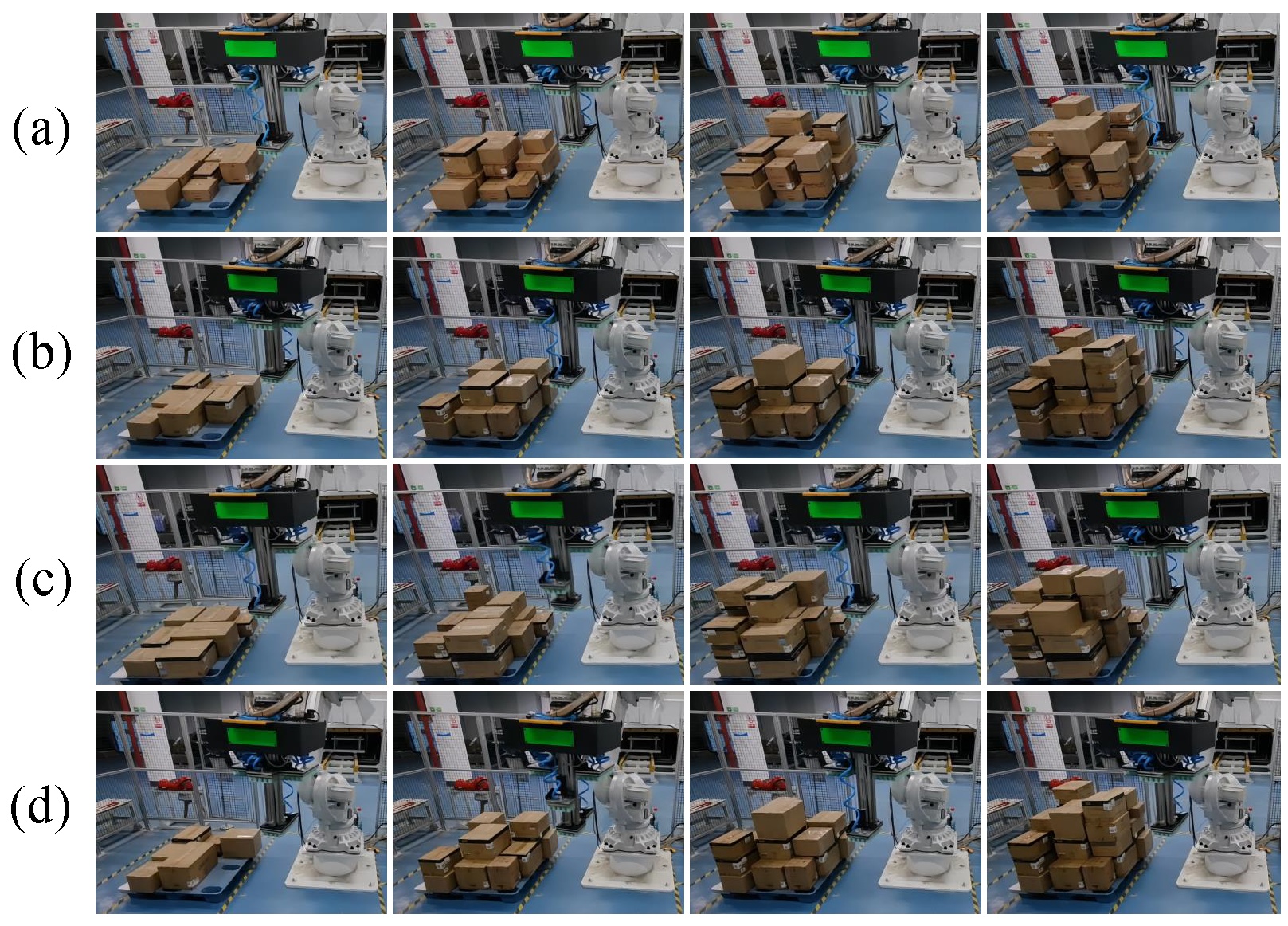}
    \vspace{-20pt}
    \caption{Real-world packing experiments conducted on a warehouse scenario with a maximum buffer size of 3.}
    \vspace{-10pt}
    \label{fig:packing_process}
\end{figure} 

The conveyor can hold up to 3 boxes for the robot to grip, modeling the general setting of packing with $ 1 \leq s \leq 3$ and $0 \leq p \leq 2$.
We adopt ToP for solving different BPP variations with no additional adaptation. The task terminates when the stack reaches the height limitation or no stable placement is available.
AGVs or forklifts then transport the stack, and the packing restarts on a new pallet. 
During placement, the robot maintains a 1.5cm gap between each box. This is necessary due to the sensitivity of packing to operation accuracy.

\input{table/stability.tex}

After box $n$ is placed, the RGB-D camera (HIKROBOT MV-DB1300A) on the pallet detects its position. 
If any deviation from the planned position is observed, the internal description of $n$ is updated accordingly to inform subsequent placement decisions. 
\rev{The average packing cycle is 9.8 seconds per box, with a variance of 0.17, indicating consistent performance across repeated trials.
The minor variations in cycle time result from differing object placements, which lead to varying transportation distances during execution.
}
Without container protection, 363 packing episodes are conducted in real-world industrial production, packing 6891 boxes with an average of 19 boxes per pallet and a  57.4\% space utilization for large-size boxes.
Throughout the packing production, no robot-object collisions occurred, and no stacks collapsed during transportation.
We summarize how our physics-aware randomization improves transportation stability in Table~\ref{tab:stability}, and take the disturbance number $k_d = 8$ for test-time stability evaluation. 
A visual example of real-world packing is shown in Figure~\ref{fig:packing_process}, with additional results provided in Appendix~\ref{subsection:visulization}.  A video showcasing the dynamic packing process, featuring active gripper adjustment and stable transportation, is included in Supplementary Material.

%% file: table/performance_comparison.tex
\begin{table*}[h]
    \caption{Performance comparisons. 
     ``Random'' refers to placements selected randomly from all coordinates. DBL, LSAH, MACS, BR, and HM are heuristics proposed by \citet{KarabulutI04}, \citet{hu2017solving}, \citet{HuXCG0020}, \citet{ZhaoS0Y021}, and \citet{WangH19a}. CDRL presents the constrained deep reinforcement learning method proposed by \citet{zhao2021learning}. 
    } 
    \label{tab:baselines}
    \centering
    \footnotesize
    \resizebox{1\textwidth}{!}{
    \begin{tabular}{ll|cccc|cccc|cccc}
    \toprule
     & \multirow{2}{*}{Method} &  \multicolumn{4}{c|}{$Setting\,1$} & \multicolumn{4}{c|}{$Setting\,2$} & \multicolumn{4}{c}{$Setting\,3$} \\
     &  &  \multicolumn{1}{c}{Uti. $\uparrow$} &  \multicolumn{1}{c}{Var. $\downarrow$} & \multicolumn{1}{c}{Num. $\uparrow$} & \multicolumn{1}{c|}{Gap $\downarrow$} & \multicolumn{1}{c}{Uti. $\uparrow$} &  \multicolumn{1}{c}{Var. $\downarrow$} & \multicolumn{1}{c}{Num. $\uparrow$} & \multicolumn{1}{c|}{Gap $\downarrow$} & \multicolumn{1}{c}{Uti. $\uparrow$} &  \multicolumn{1}{c}{Var. $\downarrow$}  & \multicolumn{1}{c}{Num. $\uparrow$} & \multicolumn{1}{c}{Gap $\downarrow$} \\
    \midrule
    \multirow{8}{*}{\rotatebox[origin=c]{90}{Heuristic}}
     &  Random  &  $36.7\%$ & $10.3$ & $14.9$ & $51.7\%$ & $38.6\%$ & $8.3$ & $15.7$ & $55.1\%$  & $36.8\%$ & $10.6$ & $14.9$ & $51.4\%$  \\
     &  BR  &  $49.0\%$ &$10.8$& $19.6$ & $35.5\%$ & $56.7\%$ &$6.6$& $22.6$ & $34.1\%$ & $48.9\%$ &$10.7$& $19.5$ & $35.4\%$ \\
     &  OnlineBPH   &  $52.1\%$ &$20.1$& $20.6$ & $31.4\%$ & $59.9\%$ &$10.4$& $23.8$ & $30.3\%$  & $51.9\%$ &$20.2$& $20.6$ & $31.4\%$ \\
     &  LSAH  &  $52.5\%$ &$12.2$& $20.8$ & $30.9\%$ & $65.0\%$ &$6.1$& $25.6$ & $24.4\%$  & $52.4\%$ &$12.2$& $20.7$ & $30.8\%$ \\
     &  HM &  $57.6\%$ &$11.5$& $24.1$ & $24.2\%$ & $66.1\%$ &$8.4$& $25.9$ & $23.1\%$  & $56.5\%$ &$11.2$& $22.3$ & $25.4\%$ \\
     &  MACS  &  $57.7\%$ &$10.5$& $22.6$ & $24.1\%$ & $50.8\%$ &$8.8$& $20.1$ & $40.9\%$  & $57.7\%$ &$10.6$& $22.6$ & $23.8\%$ \\
     & DBL &  $60.5\%$ &$8.8$ & $23.8$ & $20.4\%$ & $70.6\%$ &$7.9$& $27.8$ & $17.9\%$  & $60.5\%$ &$8.9$& $23.8$ & $20.1\%$  \\
    \cmidrule{2-14} 
    \multirow{6}{*}{\rotatebox[origin=c]{90}{Learning-based}}
     &  CDRL &  $70.9\%$ &$6.2$& $27.5$ & $6.7\%$ & $70.3\%$ &$4.3$& $27.4$ & $18.3\%$  & $59.6\%$ &$5.4$& $23.1$ & $21.3\%$ \\
     &  PCT \& CP  &  $69.4\%$ &$5.4$& $26.7$ & $8.7\%$ & $81.8\%$ &$2.0$& $31.3$ & $4.9\%$  & $69.5\%$ &$5.4$& $26.7$ & $8.2\%$ \\
     &  PCT \& EP  &  $71.9\%$ &$6.6$& $27.8$ & $5.4\%$ & $78.1\%$ &$3.8$& $30.3$ & $9.2\%$  & $72.2\%$ &$5.8$& $27.9$ & $4.6\%$ \\
     &  PCT \& FC   &  $72.4\%$ &$4.7$& $28.0$ & $4.7\%$ & $76.9\%$ &$3.3$& $29.7$ & $10.6\%$  & $69.8\%$ &$5.3$& $27.1$ & $7.8\%$ \\
     &  PCT \& EV  & $\textbf{76.0\%}$ &$\textbf{4.2}$& $\textbf{29.4}$ & $\textbf{0.0\%}$ & $85.3\%$ &$2.1$& $32.8$ & $0.8\%$ & $\textbf{75.7\%}$ &$\textbf{4.6}$& $\textbf{29.2}$ & $\textbf{0.0\%}$\\
     &  PCT \& EMS  &  $75.8\%$ &$4.4$& $29.3$ & $0.3\%$ & $\textbf{86.0\%}$ &$\textbf{1.9}$& $\textbf{33.0}$ & $\textbf{0.0\%}$  & $75.5\%$ &$4.7$& $29.2$ & $0.3\%$ \\
    \bottomrule
    \end{tabular}
    }
    \end{table*}

%% file: table/presentation.tex
\begin{table*}[ht!]
    \caption{The graph embedding of complete PCT helps the final performance.}
    \label{tab:graph_embedding}
    \centering
    \footnotesize
    \resizebox{1\textwidth}{!}{
    \begin{tabular}{l|cccc|cccc|cccc}
      \toprule
        \multirow{2}{*}{Presentation}  &  \multicolumn{4}{c|}{$Setting\,1$} & \multicolumn{4}{c|}{$Setting\,2$} & \multicolumn{4}{c}{$Setting\,3$} \\
        &  \multicolumn{1}{c}{Uti. $\uparrow$} &  \multicolumn{1}{c}{Var. $\downarrow$} & \multicolumn{1}{c}{Num. $\uparrow$} & \multicolumn{1}{c|}{Gap $\downarrow$} & \multicolumn{1}{c}{Uti. $\uparrow$} &  \multicolumn{1}{c}{Var. $\downarrow$} & \multicolumn{1}{c}{Num. $\uparrow$} & \multicolumn{1}{c|}{Gap $\downarrow$} & \multicolumn{1}{c}{Uti. $\uparrow$} &  \multicolumn{1}{c}{Var. $\downarrow$}  & \multicolumn{1}{c}{Num. $\uparrow$} & \multicolumn{1}{c}{Gap $\downarrow$} \\
       \midrule
     PointNet&  $69.2\%$ &$6.7$& $26.9$ & $8.9\%$ & $78.9\%$& $3.2$ & $30.5$ & $7.5\%$  & $71.5\%$ &$5.3$& $27.7$ & $5.5\%$ \\
     Ptr-Net  &  $64.1\%$ &$10.0$& $25.1$ & $15.7\%$ & $77.5\%$ &$4.1$& $30.1$ & $9.1\%$  & $63.5\%$ &$7.9$& $24.8$ & $16.1\%$ \\
     PCT ($\mathcal{T} / \textbf{B}$) &  $70.9\%$ &$5.9$& $27.5$ & $6.7\%$ & $84.1\%$ &$2.6$& $32.3$ & $1.4\%$  & $70.6\%$ &$5.3$& $27.4$ & $6.7\%$ \\
     PCT ($\mathcal{T}$)  & $\textbf{76.0\%}$ &$\textbf{4.2}$& $\textbf{29.4}$ & $\textbf{0.0\%}$ & $\textbf{85.3\%}$ &$\textbf{2.1}$& $\textbf{32.8}$ & $\textbf{0.0\%}$ & $\textbf{75.7\%}$ &$\textbf{4.6}$& $\textbf{29.2}$ & $\textbf{0.0\%}$ \\
    \bottomrule
    \end{tabular}
    }
    \vspace{-4pt}
    \end{table*}

%% file: table/performance_continuous.tex
\begin{table*}[ht!]
    \caption{Online 3D-BPP with continuous solution space.
    }
    \label{tab:continuous}
    \centering
    \footnotesize
    \resizebox{1\textwidth}{!}{
    \begin{tabular}{ll|cccc|cccc|cccc}
    \toprule
        & \multirow{2}{*}{Method}  &  \multicolumn{4}{c|}{$Setting\,1$} & \multicolumn{4}{c|}{$Setting\,2$} & \multicolumn{4}{c}{$Setting\,3$} \\
        & &  \multicolumn{1}{c}{Uti. $\uparrow$} &  \multicolumn{1}{c}{Var. $\downarrow$} & \multicolumn{1}{c}{Num. $\uparrow$} & \multicolumn{1}{c|}{Gap $\downarrow$} & \multicolumn{1}{c}{Uti. $\uparrow$} &  \multicolumn{1}{c}{Var. $\downarrow$} & \multicolumn{1}{c}{Num. $\uparrow$} & \multicolumn{1}{c|}{Gap $\downarrow$} & \multicolumn{1}{c}{Uti. $\uparrow$} &  \multicolumn{1}{c}{Var. $\downarrow$}  & \multicolumn{1}{c}{Num. $\uparrow$} & \multicolumn{1}{c}{Gap $\downarrow$} \\
       \midrule
    \multirow{3}{*}{\rotatebox[origin=c]{90}{Heu.}}
    &  BR  &   $40.9\%$ &$7.4$&  $16.1$ & $37.5\%$ & $45.3\%$ &$5.2$& $17.8$ & $31.7\%$  & $40.9\%$ &$7.3$& $16.1$ & $38.6\%$ \\
    &  OnlineBPH  &   $43.9\%$ &$14.2$& $17.2$ & $32.9\%$ & $46.1\%$ &$6.8$& $18.1$ & $30.5\%$  & $43.9\%$ &$14.2$& $17.2$ & $34.1\%$ \\
    &  LSAH  &  $48.3\%$ &$12.1$& $18.7$ & $26.1\%$ & $58.7\%$ &$4.6$& $22.8$ & $11.5\%$  & $48.4\%$ &$12.2$& $18.8$ & $27.3\%$ \\
    \cmidrule{2-14} 
    \multirow{3}{*}{\rotatebox[origin=c]{90}{DRL}}
     &  GD  &   $5.6\%$ &  $-$ & $2.2$ & $91.4\%$ & $7.5\%$ & $-$ & $2.9$ & $88.7\%$  & $5.2\%$ & $-$ & $2.1$ & $92.2\%$ \\
     &  PCT \& EV   &   $\textbf{65.4\%}$ &$\textbf{3.3}$ & $\textbf{25.0}$ & $\textbf{0.0\%}$ & $65.0\%$ &$2.6$& $26.4$ & $2.0\%$  & $65.8\%$ & $3.6$ & $25.1$ & $2.7\%$ \\
     &  PCT \& EMS   &   $65.3\%$ &$4.4$& $24.9$ & $0.2\%$ & $\textbf{66.3\%}$ &$\textbf{2.3}$& $\textbf{27.0}$ & $\textbf{0.0\%}$  & $\textbf{66.6\%}$ &$\textbf{3.3}$& $\textbf{25.3}$ & $\textbf{0.0\%}$ \\
     \bottomrule
    \end{tabular}
    }
    \vspace{-4pt}
    \end{table*}

%% file: table/complex_constraints.tex
\begin{table*}[ht!]
    \caption{Online 3D-BPP with practical constraints. ``Obj.'' means the task-specific objective score, which is the primary focus of this comparison. Whether a higher or lower value is preferable depends on the task, with the preference indicated by arrows next to the task name. For load balancing and height uniformity tasks, the objective score is scaled by $\times 10^{3}$ and $\times 10^{2}$, respectively.
    ``Imp.'' indicates the percentage improvement in ``Obj.'' compared to a random placement policy.
    }
    \label{tab:practical}
    \centering
    \footnotesize
    \resizebox{1\textwidth}{!}{
    \begin{tabular}{l|l|cccc|cccc|cccc}
    \toprule
    \multirow{2}{*}{Constraints} & \multirow{2}{*}{Method} &  \multicolumn{4}{c|}{$Setting\,1$} & \multicolumn{4}{c|}{$Setting\,2$} & \multicolumn{4}{c}{$Setting\,3$} \\
    & &  \multicolumn{1}{c}{Obj.} & \multicolumn{1}{c}{Imp. $\uparrow$} & \multicolumn{1}{c}{Uti. $\uparrow$} & \multicolumn{1}{c|}{Num. $\uparrow$} & \multicolumn{1}{c}{Obj.} & \multicolumn{1}{c}{Imp.$\uparrow$} & \multicolumn{1}{c}{Uti. $\uparrow$} & \multicolumn{1}{c|}{Num. $\uparrow$} & \multicolumn{1}{c}{Obj.} & \multicolumn{1}{c}{Imp.$\uparrow$} & \multicolumn{1}{c}{Uti. $\uparrow$} & \multicolumn{1}{c}{Num. $\uparrow$} \\
    \midrule
    \multirow{2}{*}{Isle Friendliness $\downarrow$}
     & CDRL & 0.20 & $31.0\%$ & $58.3\%$ & $22.5$ & 0.19 & $ 20.8\%$ & $64.2\%$ & $24.8$ & 0.20 & $48.7\%$ & $59.0\%$ & $22.8$ \\
     &  PCT  & \textbf{0.15} & \textbf{48.3\%} & \textbf{72.1\%} & \textbf{29.0} & \textbf{0.08} & \textbf{66.6\%} & \textbf{85.2\%} & \textbf{32.8} & \textbf{0.15} & \textbf{61.5\%} & \textbf{74.6\%} & \textbf{28.8} \\
     \midrule
     \multirow{2}{*}{Load Balancing $ \downarrow$}
     &  CDRL &  $3.32$ & $41.7\%$ & $55.8\%$ & $21.6$  & $3.11$ & $32.1\%$ & $ 58.3\%$ & 22.5 & $1.42$ & $ 21.5\%$ & $ 55.9\%$ & $21.7$ \\
     &  PCT   & \textbf{1.40} & \textbf{75.4\%} & \textbf{71.2\%} & \textbf{27.7}  & \textbf{0.69} & \textbf{84.9\%} & \textbf{83.5\%}  & \textbf{32.3} &  \textbf{0.22} & \textbf{87.8\%} & \textbf{71.2\%} & \textbf{27.7} \\
     \midrule
     \multirow{2}{*}{Height Uniformity $\downarrow$}
     &  CDRL & $6.99$ & $ 26.6\%$ & $53.3\%$ & $20.9$ & $7.22$ & $ 38.4\%$ & $ 54.4\%$ & 21.1 & $7.34$ & $ 21.5\%$ &  $ 54.4\%$ & $21.2$  \\
     &  PCT   & \textbf{3.79} & \textbf{60.2\%} & \textbf{74.3\%} & \textbf{28.8} & \textbf{2.01} & \textbf{82.8\%} & \textbf{83.5\%} & \textbf{32.3} & \textbf{3.81} & \textbf{59.3\%} & \textbf{73.1\%} & \textbf{28.4}  \\
     \midrule
     \multirow{2}{*}{Kinematic Constraints $\uparrow$}
     &  CDRL  & $0.77$ & $ 35.1\%$ & $54.6\%$ & $21.2$ & $0.53$ & $-31.2\%$ & $ 60.0\%$ & 23.1 & $0.79$ & $ 17.9\%$ & $ 56.5\%$ & $22.0$ \\
     &  PCT   & \textbf{0.94} & \textbf{64.9\%} & \textbf{72.8\%} & \textbf{28.2} & \textbf{0.96} & \textbf{24.7\%} & \textbf{84.8\%} & \textbf{32.6} & \textbf{0.93} & \textbf{38.8\%} & \textbf{74.6\%} & \textbf{28.8} \\
     \midrule
     \multirow{2}{*}{Load Bearing Constraints $\downarrow$}
     & CDRL  & $1.95$ & $61.8\%$ & $ 51.8\%$ & $20.7$ & $2.44$ & $32.4\%$ & $ 50.3\%$ & $19.6$ & $1.30$ & $60.0\%$ &  $ 44.7\%$ & $17.0$ \\
     &  PCT   & \textbf{1.43} & \textbf{72.0\%} & \textbf{69.8\%} & \textbf{27.9} & \textbf{1.41} & \textbf{60.9\%} & \textbf{80.6\%} & \textbf{31.8} & \textbf{0.80} & \textbf{75.4\%}  & \textbf{68.7\%} & \textbf{27.3} \\
     \midrule
     \multirow{2}{*}{Bridging Constraints $\uparrow$}
     &  CDRL  & $1.09$ &  $2.8\%$ &  $ 59.3\%$ & $23.0$ & $1.03$ & $1.0\%$ & $60.8\%$ & $23.4$ & $1.07$ & $1.9\%$ & $59.7\%$ & $23.1$  \\
     &  PCT   & \textbf{1.18} & \textbf{11.3\%} & \textbf{69.2\%} & \textbf{26.9} & \textbf{1.30} & \textbf{27.5\%} & \textbf{80.6\%} & \textbf{31.3} & \textbf{1.19} & \textbf{13.3\%} & \textbf{69.0\%} & \textbf{26.9} \\
     \bottomrule
\end{tabular}
}
\vspace{-4pt}
\end{table*}

%% file: table/large_scale.tex
\begin{table*}[t]
    \caption{Performance comparisons on large packing scales. 
    } 
    \label{tab:large_scale}
    \centering
    \footnotesize
    \resizebox{1\textwidth}{!}{
    \begin{tabular}{l|rccc|cccc|cccc}
    \toprule
    \multirow{2}{*}{Method}  & \multicolumn{4}{c|}{$\bar{N} = 200$} & \multicolumn{4}{c|}{$\bar{N} = 500$} & \multicolumn{4}{c}{$\bar{N} = 1000$} \\
     &  \multicolumn{1}{c}{Uti. $\uparrow$} &  \multicolumn{1}{c}{Var. $\downarrow$} & \multicolumn{1}{c}{Num. $\uparrow$} & \multicolumn{1}{c|}{Gap $\downarrow$} & \multicolumn{1}{c}{Uti. $\uparrow$} &  \multicolumn{1}{c}{Var. $\downarrow$} & \multicolumn{1}{c}{Num. $\uparrow$} & \multicolumn{1}{c|}{Gap $\downarrow$} & \multicolumn{1}{c}{Uti. $\uparrow$} &  \multicolumn{1}{c}{Var. $\downarrow$}  & \multicolumn{1}{c}{Num. $\uparrow$} & \multicolumn{1}{c}{Gap $\downarrow$} \\
    \midrule
    BR  &  $ 50.6\%$ & $1.3$ & $101.7$ & $ 34.2\%$ & $ 49.6\%$ & $0.4$ & $248.0$ & $  37.9\%$  & $ 49.3\%$ & $0.2$ & $482.9$ & $ 39.3\%$  \\
    OnlineBPH   &  $40.1\%$ & $1.5$ & $81.6$ & $ 47.9\%$ & $ 38.6\%$ & $0.6$ & $190.0$ & $ 51.7\%$  & $40.1\%$ & $0.3$ & $398.7$ & $ 50.6\%$  \\
    LSAH   &   $ 64.1\%$ & $1.4$ & $128.4$ & $ 16.6\%$ & $ 68.3\%$ & $0.5$ & $341.6$ & $ 14.5\%$  & $69.8\%$ & $0.1$ & $681.7$ & $ 14.0\%$  \\
    \midrule
    PCT$^{*}$  &   $72.3\%$ & $0.8$ & $144.4$ & $ 6.0\%$ & $ 74.4\%$ & $0.5$ & $370.9$ & $ 6.9\%$  & $ 56.4\%$ & $0.4$ & $553.4$ & $ 30.5\%$  \\
    PCT$^{\dag}$ &   $74.4\%$ & $\textbf{0.3}$ & $147.8$ & $ 3.3\%$ & $  74.5\%$ & $0.2$ & $371.1$ & $ 6.8\%$  & $73.6\%$ & $ 0.1$ & $719.1$ & $9.4\%$  \\
    Recursive Packing  &   $\textbf{76.9}\%$ & $0.5$ & $\textbf{153.1}$ & $\textbf{0.0}\%$ & $ \textbf{79.9}\%$ &$\textbf{0.1}$& $\textbf{397.6}$ & $ \textbf{0.0}\%$  & $ \textbf{81.2}\%$ &$\textbf{0.01}$& $\textbf{792.5}$ & $ \textbf{0.0}\%$  \\
    \bottomrule
    \end{tabular}
    }
    \vspace{-4pt}
    \end{table*}

%% file: table/fusion_function.tex
\begin{table*}[h]
    \caption{Performance of different solution integration functions for ToP.
    }

    \label{tab:funsion_function}
    \centering
    \footnotesize
    \resizebox{1\textwidth}{!}{
    \begin{tabular}{l|cccc|cccc|cccc}
    \toprule
        \multirow{2}{*}{Integration Method}  &  \multicolumn{4}{c|}{$\bar{N} = 200$} & \multicolumn{4}{c|}{$\bar{N} = 500$} & \multicolumn{4}{c}{$\bar{N} = 1000$} \\
        &  \multicolumn{1}{c}{Uti. $\uparrow$} &  \multicolumn{1}{c}{Var. $\downarrow$} & \multicolumn{1}{c}{Num. $\uparrow$} & \multicolumn{1}{c|}{Gap $\downarrow$} & \multicolumn{1}{c}{Uti. $\uparrow$} &  \multicolumn{1}{c}{Var. $\downarrow$} & \multicolumn{1}{c}{Num. $\uparrow$} & \multicolumn{1}{c|}{Gap $\downarrow$} & \multicolumn{1}{c}{Uti. $\uparrow$} &  \multicolumn{1}{c}{Var. $\downarrow$}  & \multicolumn{1}{c}{Num. $\uparrow$} & \multicolumn{1}{c}{Gap $\downarrow$} \\
    \midrule
    Maximum State Value & $ 60.5\%$ & $1.5$ & $121.2$ & $ 21.3\%$ & $ 46.9\%$ &$1.3$& $234.5$ & $41.3\%$  & $41.4\%$ &$1.2$& $411.3$ & $ 49.0\%$  \\
    Maximum Return   & $66.2\%$ & $3.2$ & $132.4$ & $ 13.9\%$ & $ 50.0\%$ &$2.5$& $249.5$ & $ 37.4\%$  & $45.4\%$ &$2.4$& $446.6$ & $ 44.1\%$  \\
    Maximum Volume & $66.6\%$ & $1.4$ & $133.2$ & $ 13.4\%$ & $61.5\%$ &$4.7$& $307.6$ & $ 23.0\%$  & $48.7\%$ &$8.1$& $479.3$ & $ 40.0\%$  \\
    Minimum Surface Area  & $65.4\%$ & $2.3$ & $130.8$ & $ 15.0\%$ & $ 55.5\%$ &$2.1$& $277.6$ & $ 30.5\%$  & $49.7\%$ &$4.1$& $489.0$ & $ 38.8\%$  \\
     Spatial Ensemble 
     & $\textbf{76.9}\%$ & $\textbf{0.5}$ & $\textbf{153.1}$ & $\textbf{0.0}\%$ & $ \textbf{79.9}\%$ &$\textbf{0.1}$& $\textbf{397.6}$ & $ \textbf{0.0}\%$  & $ \textbf{81.2}\%$ &$\textbf{0.01}$& $\textbf{792.5}$ & $ \textbf{0.0}\%$  \\
     \bottomrule
    \end{tabular}
    }
    \vspace{-10pt}
\end{table*}

%% file: table/various_settings.tex
\begin{table*}[t!]
    \caption{
    \rev{Performance comparison across different 3D-BPP variations with online properties.}}
    \label{tab:various_setting}
    \centering
    \footnotesize
    \setlength{\tabcolsep}{0.52em}
    \resizebox{1\textwidth}{!}{
    \begin{tabular}{l|ccc|ccccc|ccccc|ccccc}
    \toprule
        \multirow{2}{*}{Method} & \multirow{2}{*}{Sel.} &  \multirow{2}{*}{Prev.} &  \multirow{2}{*}{Un.} & 
        \multicolumn{5}{c|}{$Setting\,1$} & 
        \multicolumn{5}{c|}{$Setting\,2$} & 
        \multicolumn{5}{c}{$Setting\,3$} \\
         &  & &  
         & \multicolumn{1}{c}{Uti. $\uparrow$} & \multicolumn{1}{c}{Var. $\downarrow$} & \multicolumn{1}{c}{Num. $\uparrow$} & \multicolumn{1}{c}{Gap $\downarrow$} & \multicolumn{1}{c|}{Time  $\downarrow$} & 
           \multicolumn{1}{c}{Uti. $\uparrow$} & \multicolumn{1}{c}{Var. $\downarrow$} & \multicolumn{1}{c}{Num. $\uparrow$} & \multicolumn{1}{c}{Gap $\downarrow$} & \multicolumn{1}{c|}{Time  $\downarrow$} & 
           \multicolumn{1}{c}{Uti. $\uparrow$} & \multicolumn{1}{c}{Var. $\downarrow$} & \multicolumn{1}{c}{Num. $\uparrow$} & \multicolumn{1}{c}{Gap $\downarrow$} & \multicolumn{1}{c}{Time  $\downarrow$} \\
       \midrule
       CDRL & $1$ & $9$ & $>0$ & $82.5\%$ &$6.0$& $32.3$ & $0.7\%$ & 5.3 & $75.0\%$& $1.3$ & $29.0$ & $17.4\%$ & 3.1 & $68.4\%$ &$5.3$& $27.3$ & $19.2\%$ & 5.5 \\
       ToP & $1$ & $9$  & $>0$ & \textbf{83.1\%} & \textbf{5.6} & \textbf{32.5} & \textbf{0.0\%} & \textbf{1.8} & \textbf{90.8\%} & \textbf{1.1} & \textbf{35.3} & \textbf{0.0\%} & \textbf{1.1} & \textbf{84.7\%} & \textbf{1.9} & \textbf{33.1} & \textbf{0.0\%} & \textbf{1.8} \\
       \midrule
       O3DBP & $5$ & $5$ & $>0$ & $53.1\%$ &$7.5$& $26.4$ & $39.0\%$ & 3.7 & $60.4\%$& $3.6$ & $29.5$ & $35.3\%$ & 3.8 & $53.4\%$ &$7.4$& $26.5$ & $39.5\%$ & 3.8 \\
       ToP & $5$ & $5$  & $>0$ & \textbf{87.1\%} & \textbf{1.2} & \textbf{34.2} & $\textbf{0.0\%}$ & \textbf{2.0} & \textbf{93.3\%} & \textbf{0.9} & \textbf{36.3} & $\textbf{0.0\%}$ & \textbf{1.3} & \textbf{88.3\%} & \textbf{1.3} & \textbf{34.4} & $\textbf{0.0\%}$ & \textbf{2.1} \\
       \midrule
       TAP-NET++ & $10$ & $0$ & $>0$ & $66.8\%$ &$3.6$& $20.6$ & $24.8\%$ & $\mathbf{6.2 \times 10^{-2}}$ & $77.0\%$& $2.1$ & $26.6$ & $18.9\%$ & $\mathbf{4.6 \times 10^{-2}}$ 
       & $68.2\%$ & $3.0$& $21.2$ & $23.5\%$ & 
       $\mathbf{6.3 \times 10^{-2}}$ 
       \\ 
       ToP & $10$ & $0$ & $>0$ & $\textbf{88.8\%}$ &$\textbf{1.2}$& $\textbf{34.8}$ & $\textbf{0.0\%}$ & 2.1 & $\textbf{95.0\%}$& $\textbf{0.3}$ & $\textbf{37.1}$ & $\textbf{0.0\%}$ & 1.5 & $\textbf{89.1\%}$ &$\textbf{1.6}$& $\textbf{35.1}$ & $\textbf{0.0\%}$ & 2.1 \\
      \bottomrule
\end{tabular}
    }
\vspace{-4pt}
\end{table*}

%% file: table/offline_setting.tex
\begin{table}[t!]
    \caption{\rev{Performance comparisons on offline 3D-BPP.}}
    \label{tab:offline_setting}
    \centering
    \footnotesize
    \setlength{\tabcolsep}{0.6em}
    \resizebox{0.49\textwidth}{!}{
    \begin{tabular}{l|ccc|ccccc}
        \toprule
        \multirow{2}{*}{Method}  & \multirow{2}{*}{Sel.}  & \multirow{2}{*}{Prev.} &  \multirow{2}{*}{Un.} & 
        \multicolumn{5}{c}{$Setting\,2$} \\
         &  &  & & 
         \multicolumn{1}{c}{Uti. $\uparrow$} &  
         \multicolumn{1}{c}{Var. $\downarrow$} & 
         \multicolumn{1}{c}{Num. $\uparrow$} & 
         \multicolumn{1}{c}{Gap $\downarrow$} & 
         \multicolumn{1}{c}{Time $\downarrow$} \\
        \midrule
        Gurobi & $50$ & $0$ & $0$ & $76.8\%$ & 11.5 & $29.9$ & 19.3\% & 20.1 \\
        RCQL & $50$ & $0$ & $0$ & $77.8\%$ & $3.4$ & $19.9$ & $18.3\%$ & $\mathbf{1.3\times 10^{-1}}$ \\
        Attend2Pack & $50$ & $0$ & $0$ & $87.6\%$ & $4.4$ & $26.0$ & $8.0\%$ & 3.6$\times 10^{-1}$ \\
        TAP-NET++ & $50$ & $0$ & $0$ & $89.2\%$ & 1.2 & $28.0$ & $6.3\%$ & 5.8$\times 10^{-1}$\\
        ToP & $50$ & $0$ & $0$ & $\textbf{95.2\%}$ & $\textbf{0.2}$ & $\textbf{36.3}$ & $\textbf{0.0\%}$ & 1.8 \\
        \bottomrule
\end{tabular}
    }
\vspace{-10pt}
\end{table}

%% file: table/stability.tex
\begin{table}[t!]
    \caption[font = \small]{Real-world transportation  stability (\%). Test-side evaluation is conducted with quasi-static equilibrium~\citep{zhao2021learning} and our physics-based verification. Each is evaluated with 20 packing episodes.
    } 
    \label{tab:stability}
    \centering
    \footnotesize
    \resizebox{0.49\textwidth}{!}{
    \begin{tabular}{c|cccc}
        \toprule
        Quasi-Static Equilibrium & Ours ($k_d = 1$) &  Ours $(k_d = 2) $
         & Ours $(k_d = 4)$ & Ours $(k_d = 8)$  \\
        \midrule
        $55\%$ & $70\%$ & $85\%$ & $95\%$ & $\textbf{100.0\%}$  \\
        \bottomrule
\end{tabular}
    }
\vspace{-14pt}
\end{table}

%% file: conclusion.tex
\section{Conclusions and Discussions}

We formulate the online 3D-BPP as a novel hierarchical representation—packing configuration tree (PCT). 
PCT is a full-fledged description of the state and action space of bin packing, which makes the DRL agent easy to train and well-performing.  
We extract state features from PCT using graph attention networks which encode the spatial relations of all space configuration nodes. 
The graph representation of PCT  helps the agent with handling online 3D-BPP with complex practical constraints, while the finite leaf nodes prevent the action space from growing explosively. 
We further discover the potential of PCT as a tree-based planner to deliberately solve packing problems of industrial significance, including large-scale packing and different BPP variations.

Our method surpasses all other online packing competitors and is the first learning-based method that solves online 3D-BPP with continuous solution space. It performs well even on item sampling distributions that are different from the training one.  We also give demonstrations to prove that PCT is versatile in terms of incorporating various practical constraints. 
The PCT-driven planning excels across large problem scales and different BPP variations, with performance improving as the problem scales and decision variables increase.
Our real-world packing robot operates reliably on unprotected pallets, densely packing at 10 seconds per box with an average of 19 boxes and a  57.4\% space utilization for large-size boxes.

\paragraph*{\textbf{Limitation and future work}}
We see several important opportunities for future research.
First, we use the default physical simulation parameters~\citep{makoviychuk2021isaac} to evaluate the stability of real-world packing,  where the Sim2Real gap can lead to incorrect solutions being either discarded or retained. From both academic and industrial perspectives, we are highly interested in estimating physical parameters from real-world packing processes to achieve system identification~\citep{ljung1998system}, support more accurate stability evaluation and Real2Sim2Real packing policy learning~\citep{li2025pinwm}. 
\rev{Additionally, this work proves from a geometric perspective that the PCT candidates cover placements that are locally optimal in a tightness measure; however, it does not provide tight performance bounds for the parameterized policies.
Previous study~\citep{yang2021competitive} provides only a loose bound on the competitive ratio for the online 3D-BPP, showcasing that the performance gap between the offline optimum and any online methods is no smaller than $\Omega(\log \theta\alpha)$, where $\theta$  is the ratio between the maximum and minimum item value, and $\alpha$  is the ratio of the aggregate capacity to the minimum capacity. This bound primarily depends on the distribution of the input problem data, while tighter bounds are algorithm-specific.
Establishing tight performance guarantees for parameterized PCT policies whose behavior evolves throughout training remains a challenging yet meaningful future direction, as it helps clarify the theoretical positioning  of PCT.}
Last but not least, this work expands the industrial applicability of PCT through planning, with its capability still bounded by the pre-trained base model. 
 Developing PCT as a foundation model~\citep{Foundation_model} trained on diverse packing data, with zero-shot generalization to new packing scenarios, is an exciting and promising avenue for future research.

 \section{Acknowledgments}

 This work was supported in part by the NSFC (62325211, 62132021, 62502347, 62522219, 62372457, 62322207, 62225113), the Postdoctoral Fellowship Program and China Postdoctoral Science Foundation  (BX20250386), the Major Program of Xiangjiang Laboratory (23XJ01009), Key R\&D Program of Wuhan (2024060702030143).

%% file: appendix.tex
\appendix
\section*{Appendix}
In this appendix, we provide more details and statistical results of our PCT method.
Our real-world packing demo is also submitted with the supplemental material.

\section{Implementation Details}
\label{section:appendix_para}

\paragraph{\textbf{Deep Reinforcement Learning}} 
We formulate online 3D-BPP as a Markov Decision Process and solve it with the deep reinforcement learning method.
A DRL agent seeks a policy $\pi$ to maximize the accumulated discounted reward:
\begin{equation}
J(\pi)=E_{\psi \sim \pi}[\sum_{t=0}^{\infty}\gamma^tR(s_t,a_t)] 
\end{equation}	
Where $\gamma \in [0,1]$ is the discount factor, and $\psi = (s_0,a_0,s_1,\ldots)$ is a trajectory sampled based on the policy $\pi$. 
We extract the feature of state $s_t = (\mathcal{T}_t, n_t)$ using graph attention networks~\citep{VelickovicCCRLB18} for encoding the spatial relations of all space configuration nodes. The context feature is fed to two key components of our pipeline: an actor network and a critic network. The actor network, designed based on a pointer mechanism, weighs the leaf nodes of PCT, which is written as $\pi(a_t|s_t)$. The action $a_t$ is an index of selected leaf node $l \in \textbf{L}_t$, denoted as $a_t = index(l)$.
The critic network maps the context feature into a state value prediction  $V(s_t)$, which helps the training of the actor network.
The whole network is trained via a composite loss $L = \alpha \cdot L_{actor} + \beta \cdot L_{critic}$ ($\alpha = \beta = 1$ in our implementation), which consists of actor loss $L_{actor}$ and critic loss $L_{critic}$. These two loss functions are defined as:
\begin{equation}
    \begin{array}{ll}
        \displaystyle L_{actor} & \displaystyle =  (
            r_{t}+ \gamma  V(s_{t+1}) -V(s_t))\log
    \pi(a_t | s_t) \\
        \displaystyle L_{critic} & \displaystyle = (r_{t}+ \gamma  V(s_{t+1})-V(s_t))^2  \\
    \end{array}
\end{equation}	
Where $\displaystyle r_t = c_r \cdot w_t $ is our reward signal, and we set $\gamma$ as 1 since the packing episode is finite. We adopt a step-wise reward $r_t = c_{r} \cdot w_t$ once $n_t$ is inserted into PCT as an internal node successfully. Otherwise, $r_t = 0$ and the packing episode ends. The choice of item weight $w_t$ depends on the packing preferences. In the general sense, we set $w_t$ as the volume occupancy $v_t= s^x_n \cdot s^y_n \cdot s^z_n$ of $n_t$, and the constant $c_r$ is $10/(S^x \cdot S^y \cdot S^z)$. For online 3D-BPP with additional packing constraints, this weight can be set as $w_t = max(0, v_t - c \cdot O(s_t,a_t))$. While the term $v_t$  ensures that space utilization is still the primary concern, the objective function $O(s_t,a_t)$ guides the agent to satisfy additional constraints like isle friendliness and load balancing. 
We adopt the ACKTR~\citep{wu2017scalable} method for training our DRL agent.
\cite{ZhaoS0Y021} have demonstrated that this method has a  superiority on online 3D-BPP over other model-free DRL algorithms like SAC~\citep{HaarnojaZAL18}. 

\paragraph{\textbf{Feature extraction}} Specifically, ACKTR runs multiple parallel processes (64 here) to interact with their respective environments and gather samples. The different processes may have different packing time steps $t$ and deal with different packing sequences; the space configuration node number N also changes. To combine these data with irregular shapes into one batch, we fulfill $\textbf{B}_t$ and $\textbf{L}_t$ to fixed lengths,  $80$ and $25\cdot|\textbf{O}|$ respectively, with dummy nodes. The descriptors for dummy nodes are all-zero vectors and have the same size as the internal nodes or the leaf nodes. The relation weight logits $u_{ij}$ of dummy node $j$ to arbitrary node $i$ is replaced with $-inf$  to eliminate these dummy nodes during the feature calculation of GAT. The global context feature $\bar{h}$ is aggregated only on the eligible nodes $\textbf{h}$: $\bar{h} = \frac{1}{N} \sum_{i = 1}^N h_i$. All space configuration nodes are embedded by \text{GAT} as a fully connected graph as Figure~\ref{fig:fullConnect} (b), without any inner mask operation. 

We only provide the packable leaf nodes that satisfy placement constraints for DRL agents. For $setting\,2$, we check in advance if a candidate placement satisfies Equation~(\ref{eq:non-overlapping}) and~(\ref{eq:containment}). For $setting\,1$ and $setting\,3$, where the mass of item $n_t$ is $v_t$ and $\rho \cdot v_t$ respectively, we will additionally check if one placement meets the constraints of packing stability. 
Benefits from the fast stability estimation method proposed by \cite{zhao2021learning}, this pre-checking process can be completed in a very short time, and our DRL agent samples data at a frequency of more than 400 FPS. 

The node-wise MLPs $\phi_{\theta_B}, \phi_{\theta_L}$, and $\phi_{\theta_n}$ used to embed raw space configuration nodes are two-layer linear networks with LeakyReLU activation function.  $\phi_{FF}$ is a two-layer linear structure activated by ReLU.
The feature dimensions $d_h,d_k,\text{and}\,d_v$ are 64. The hyperparameter $c_{clip}$ used to control the range of clipped compatibility logits is set to 10  in our GAT implementation.

\paragraph{\textbf{Choice of PCT Length}}
Since PCT allows discarding some valid leaf nodes and this will not harm our performance, we randomly intercept a subset $\textbf{L}_{sub_t}$ from $\textbf{L}_t$ if $|\textbf{L}_t|$ exceeds 
a certain length.
Determining the suitable PCT length for different bin configurations is important, and we give our recommendations for finding this hyperparameter. 
For training, we find that the performance of learned policies is more sensitive to the number of allowed orientations $|\textbf{O}|$.  Thus, we set the PCT length as  $c \cdot|\textbf{O}|$ where $c$ can be determined by a grid search near $c=25$ for different bin configurations.  For our experiments, $c=25$ works quite well.
During the test, the PCT length can be different from the training one. We suggest searching for this interception length with a validation dataset via a grid search, which ranges from 50 to 300 with a step length of 10.

\section{Leaf Node Expansion Schemes}
\label{section:appendix_expanding}

\begin{figure}
    \centering
    \includegraphics[width=0.95\linewidth]{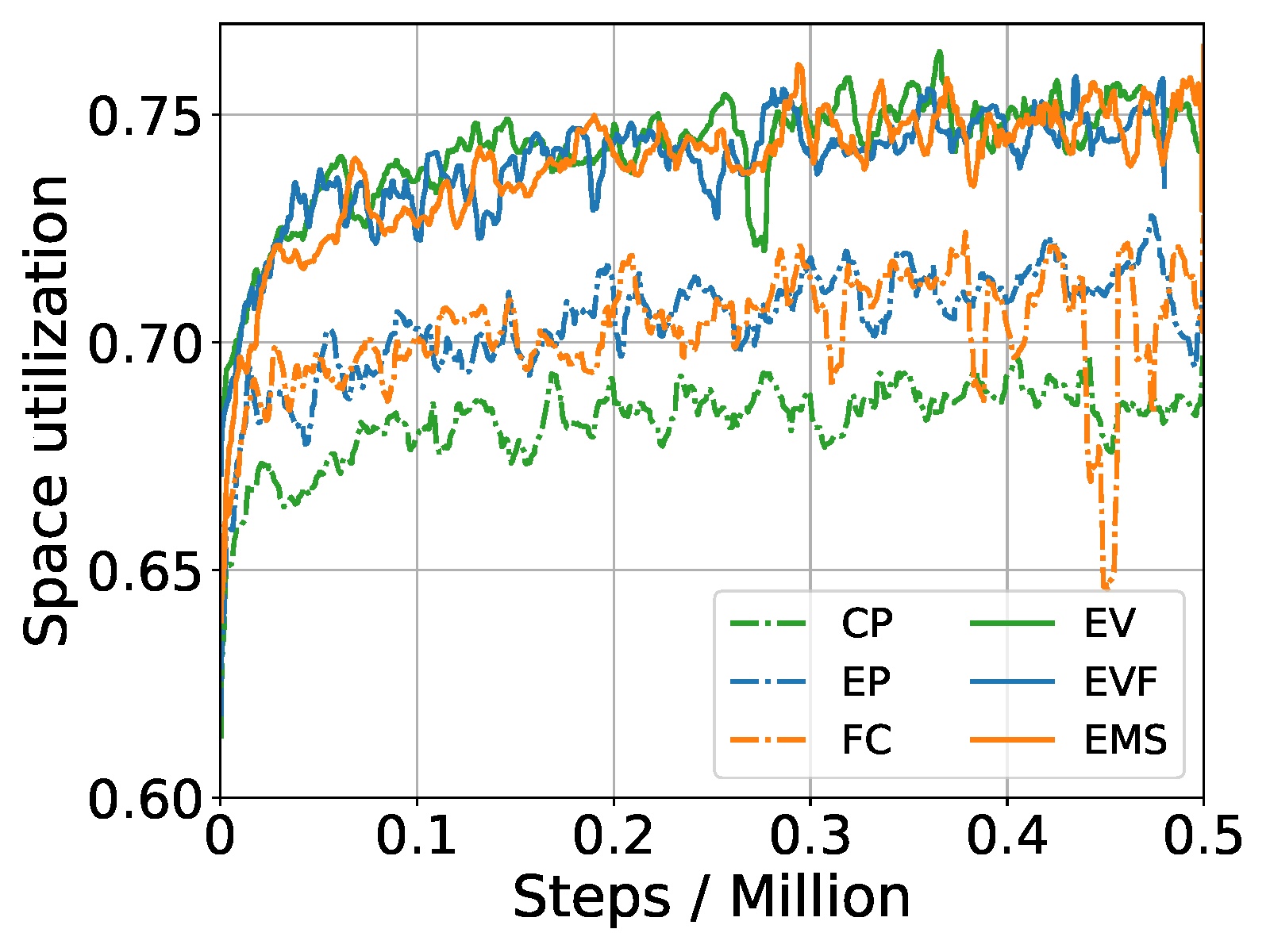}
    \caption{Learning curves on $setting\,1$.    
    A good expansion scheme for PCT reduces the complexity and helps DRL methods for more efficient learning and better performance. EVF means the full EV leaf node set without an interception.
     }
    \label{fig:curve}
\end{figure}  
We introduce the leaf node expansion schemes adopted in our PCT implementation here.  These schemes are used to incrementally calculate new candidate placements introduced by the just-placed item $n_t$. 
A good expansion scheme should reduce the number of solutions to be explored while not missing too many feasible packings. Meanwhile, polynomials computability is also expected.
As shown in Figure~\ref{fig:curve}, the policies guided by suitable leaf node expansion schemes outperform the policy trained on a full coordinate (FC) space in the whole training process.
We extend three existing heuristic placement rules, which have proven to be both accurate and efficient, to our PCT expansion, i.e., \textit{Corner Point}, \textit{Extreme Point}, and \textit{Empty Maximal Space}. Since all these schemes are related to boundary points of packed items, we combine the start/end points of $n_t$ with these boundary points as a superset, namely \textit{Event Point}.

\begin{figure*}[t]
    \begin{center}
    \centerline{\includegraphics[width=\textwidth]{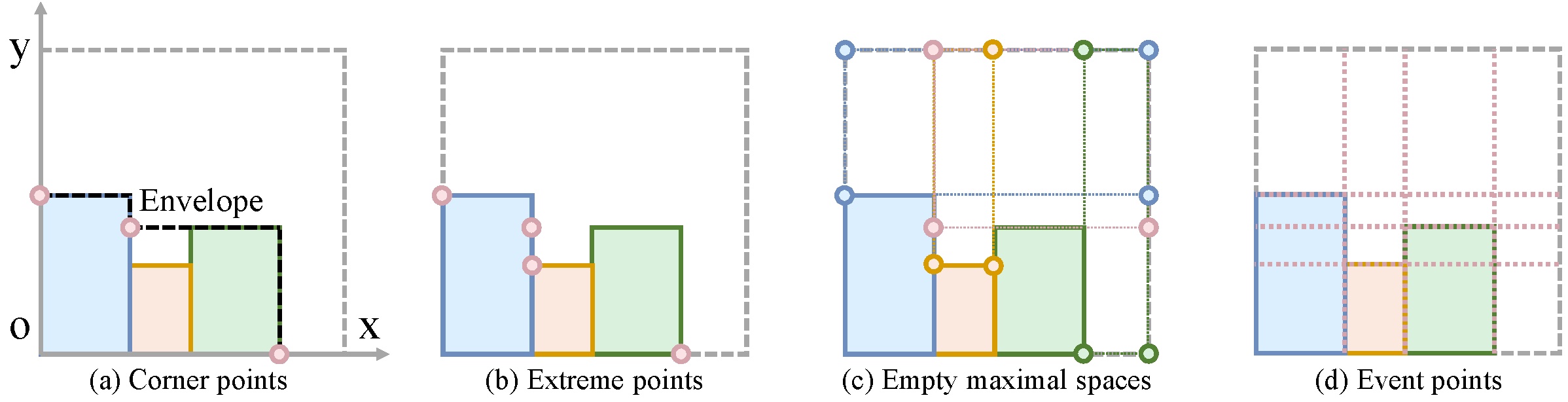}}
    \caption{Full candidate positions generated by different PCT expansion schemes (all in $xoy$ plane). The gray dashed lines are the boundaries of the bin. 
    Circles in (a) and (b) represent corner points and extreme points, respectively. (c) The candidate positions (circles) introduced by different EMSs are rendered with different colors. All intersections of two dashed lines in (d) constitute event points. 
    }
\label{fig:schemes}
\end{center}
\end{figure*}

\paragraph{\textbf{Corner Point}} 
\cite{MartelloPV00} first introduce the concept of Corner Point (CP) for their branch-and-bound methods.  
Given 2D packed items in the $xoy$ plane, the corner points can be found where the envelope of the items in the bin changes from vertical to horizontal, as shown in Figure~\ref{fig:schemes} (a). 
The past corner points that no longer meet this condition will be deleted.

Extend this 2D situation to 3D cases, the new candidate 3D positions introduced by the just placed item $n_t$ are a subset of  $\{(p_n^x + s_n^x,  p_n^y,  p_n^z),(p_n^x,  p_n^y+  s_n^y,  p_n^z),
(p_n^x,  p_n^y,  p_n^z + s_n^z)
\}$ if the envelope of the corresponding 2D plane, i.e.$\,$$xoy,yoz,\text{and }xoz$, is changed by $n_t$. 
The time complexity of finding 3D corner points incrementally is $O(c)$ with an easy-to-maintained bin height map data structure to detect the change of envelope on each plane, where $c$ is a constant.

\paragraph{\textbf{Extreme Point}} 
\cite{CrainicPT08} extend the concept of Corner Point to Extreme Point (EP) and claim their method reaches the best offline performance of that era.
Its insight is to provide the means to exploit the free space defined inside a packing by the shapes of the items that already exist. When the current item $n_t$ is added, new EPs are incrementally generated by projecting the coordinates $\{(p_n^x + s_n^x,  p_n^y,  p_n^z),(p_n^x,  p_n^y+  s_n^y,  p_n^z),
(p_n^x,  p_n^y,  p_n^z + s_n^y)
\}$ on the orthogonal axes, e.g.,  project $(p_n^x + s_n^x,  p_n^y,  p_n^z)$ in the directions of the $y$ and $z$ axes to find  intersections with all items lying between item $n_t$ and the boundary of the bin. The nearest intersection in the respective direction is an extreme point. Since we stipulate a top-down loading direction, the 3D extreme points in the strict sense may exist a large item blocking the loading direction. So we find the 2D extreme points (see Figure~\ref{fig:schemes} (b)) in the $xoy$ plane and repeat this operation on each distinct $p^z$ value (i.e.$\,$start/end $z$ coordinate of a packed item) which satisfies $p^z_n \le p^z \le p^z_n + s^z_n $. The time complexity of this method is $O(m \cdot |\textbf{B}_{2D}|)$, where $\textbf{B}_{2D}$ is the packed items that exist in the corresponding $z$ plane and $m$ is the number of related $z$ scans.

\begin{figure*}[h]
    \begin{center}
    \centerline{\includegraphics[width=\textwidth]{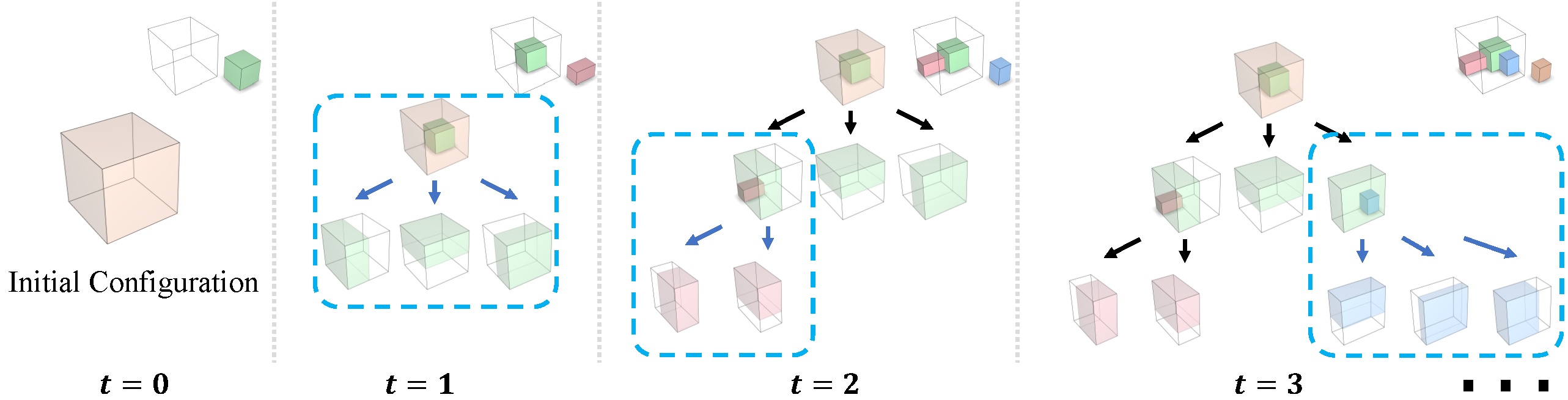}}
    \caption{
        A 3D PCT expansion schematic diagram. This PCT grows under the guidance of the EMS expansion scheme. For simplicity, we only choose the bottom-right-up corners of each EMS as candidate positions, and we set $|\textbf{O}| = 1$ here. 
    }
\vspace{-14pt}
\label{fig:PackingTree3D}
\end{center}
\end{figure*}

\paragraph{\textbf{Empty Maximal Space}} 
Empty Maximal Spaces (EMSs)~\citep{ha2017online} are the largest empty orthogonal spaces 
whose sizes cannot extend more along the coordinate axes from their front-left-bottom (FLB) corner.
This is a simple and effective placement rule. An EMS $e$ is presented by its FLB corner $(p_e^x, p_e^y, p_e^z)$ and sizes $(s_e^x, s_e^y, s_e^z)$. When the current item $n_t$ is placed into $e$ on its FLB corner, this EMS  is split into three smaller EMSs with positions $(p_e^x + s_n^x, p_e^y, p_e^z),(p_e^x, p_e^y + s_n^y, p_e^z),(p_e^x, p_e^y, p_e^z + s_n^z)$ and sizes $(s_e^x - s_n^x, s_e^y, s_e^z), (s_e^x, s_e^y - s_n^y, s_e^z),(s_e^x, s_e^y, s_e^z - s_n^z)$, respectively. 
If the item $n_t$ only partially intersects with $e$, we can apply a similar volume subtraction to the intersecting part for splitting $e$.
For each ems, we define the left-up $(p_e^x, p_e^y +  s_e^y, p_e^z)$, right-up $(p_e^x + s_e^x, p_e^y + s_e^y, p_e^z)$, left-bottom $(p_e^x, p_e^y, p_e^z)$, and right-bottom $(p_e^x + s_e^x, p_e^y, p_e^z)$ corners of its vertical bottom as candidate positions, as shown in Figure~\ref{fig:schemes} (c). These positions also need to be converted to the FLB corner coordinate for placing item $n_t$. The  left-up, right-up, left-bottom, and right-bottom corners of $e$ should be 
converted to  $(p_e^x, p_e^y + s_e^y - s_n^y, p_e^z)$,  $(p_e^x + s_e^x - s_n^x, p_e^y + s_e^y - s_n^y, p_e^z)$,  $(p_e^x, p_e^y, p_e^z)$, and $(p_e^x + s_e^x - s_n^x, p_e^y, p_e^z)$ respectively.
Since all EMSs $e\in\textbf{E}$  in the bin need to detect intersection with $n_t$, the time complexity of finding 3D EMSs incrementally is $O(|\textbf{E}|)$. A 3D schematic diagram of PCT expansion guided by EMSs is provided in Figure~\ref{fig:PackingTree3D}.

\paragraph{\textbf{Event Point}} 
It’s not difficult to find that all schemes mentioned above are related to boundary points of a packed item along $d\in\{x,y\}$ axes (we assume the initial empty bin is also a special packed item here).
When the current item $n_t$ is packed, we update the existing PCT leaf nodes by scanning all distinct $p^z$ values that satisfy $p^z_n \le p^z \le p^z_n + s^z_n $ and combine the start/end points of $n_t$ with the boundary points that exist in this $z$ plane to get the superset (see Figure~\ref{fig:schemes} (d)), which is called \textit{Event Points}. 
The time complexity for detecting event points is $O(m\cdot|B_{2D}|^2)$.

\rev{
\section{Theoretical Analysis of PCT}
\label{sec:theory}

Although PCT is guided by heuristic rules that also restrict the solution space, we provide a theoretical proof that certain heuristic-generated candidates are locally optimal. Reinforcement learning then builds upon these local optima to ultimately make a global decision.

Given an item $n$, the goal is to identify a finite set of candidate placements where $n$ is tightly positioned against obstacles (i.e., the container or packed items), such that it permits no further movement and leaves maximal space for future placements.
Since we adopt a top-down placement manner that determines item positions on the $xoy$ plane, the 3D packing can be reduced to a 2.5D formulation, i.e.,  analyzing optimal placement on each horizontal 2D plane with the same $z$ value.
For an item $n$ with a known orientation, we denote its horizontal projection $G\subset\mathbb{R}^2$, which is a 2D polygon.
Accordingly, we extract 2D No-Fit Polygons (NFP) $E$ from each $z$ plane.
The NFP~\citep{burke2007complete} is a geometric representation of all relative positions where one polygon can touch another without overlapping, as shown in Figure~\ref{fig:NFP} (a).  
Let $O\subset\mathbb{R}^2$ denote the 2D  projection of packed items; we can get NFP
$E$  with $E\oplus G \subset C$ and  $E \oplus G \cap O = \emptyset$, where $\oplus$ is the Minkowski sum~\citep{mark2008computational}. 
A 2D theorem can be established to show that the convex vertices of $E$ are local optimum for tight packing:
\begin{theorem}
\label{theorem:convex}
For a boundary point
$p\in\partial E$, if $p$ is a convex vertex, then $p = \text{argmax}_{q\in \mathcal{N}(p)}\psi(q)$ for an open neighborhood $\mathcal{N}(p)$ of $p$ and a tightness measure $\psi(\cdot)$.
\end{theorem}

\begin{figure}[t]
\centering
\centerline{\includegraphics[width=0.48\textwidth]{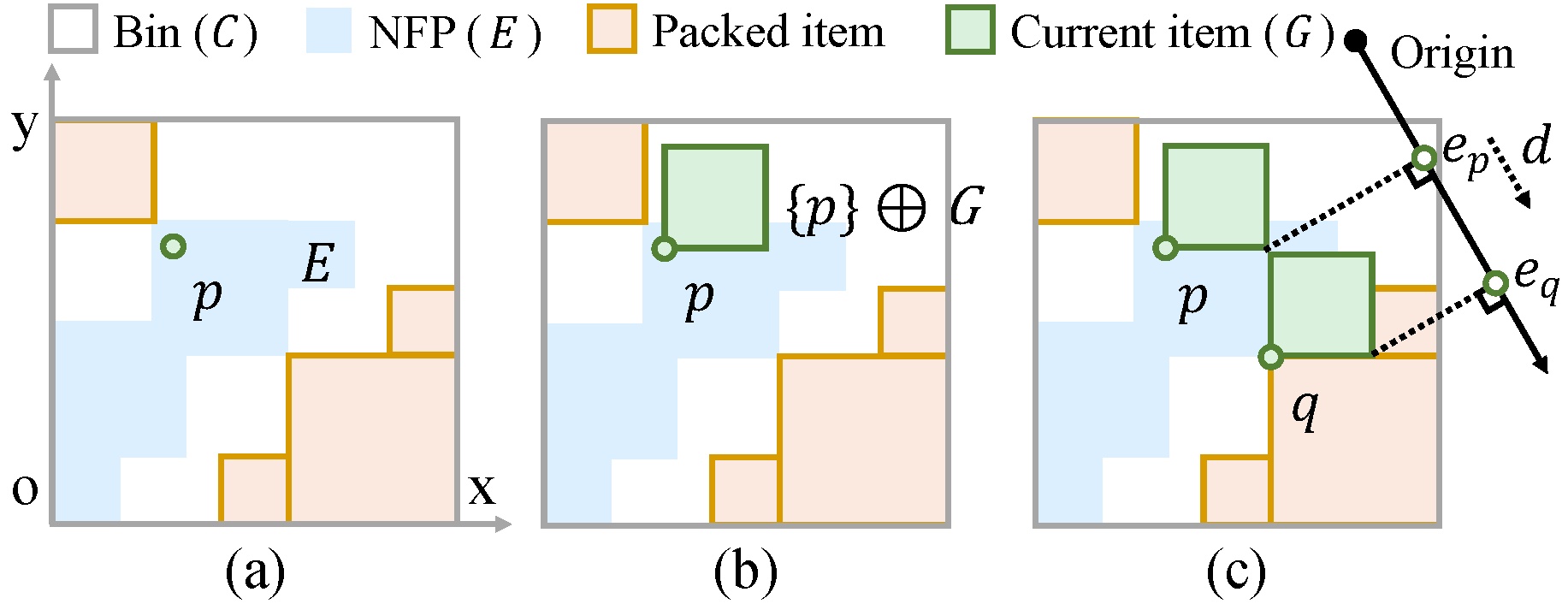}}
\vspace{-10pt}
\caption{\label{fig:NFP}
\rev{Select a reference point $p$ from the NFP $E$ (a), its corresponding placement is $\{p\} \oplus G$ (b). (c):
In direction $d$, the projection $e_q = e(G,q,d)$ of  $\{q\} \oplus G$ exceeds   $e_p$ of  $\{p\} \oplus G$,
indicating  $\{q\} \oplus n$ is a more extreme placement in $d$.
}}
\vspace{-10pt}
\end{figure}

We formally define our tightness measure $\psi(\cdot)$ and provide a proof of Theorem~\ref{theorem:convex}. 
The placement of item $n$ refers to selecting a reference point $p$ from $E$ with $\{p\} \oplus G$ being the corresponding placement, as illustrated in Figure~\ref{fig:NFP} (b). 
We prefer reference points that result in $n$ being tightly packed against obstacles. 
However, quantifying this tightness is nontrivial. 
\citet{mark2008computational} introduce the notion of \textit{extremeness} for polygons:  a polygon is considered more extreme in direction $d$ if its furthest point along $d$ lies farther than that of another. Inspired by this, we define the tightness measure as $e(G,p,d)=\max_{g\in\{p\}\oplus G}d^Tg$, which is the extreme value of $G$'s projection for the packing task.
Placement at $p$ is more extreme than at $q$ if $e(G,p,d) > e(G,q,d)$.
When $G$ is tightly packed against obstacles along $d$, $e(G,p,d)$ achieves a local maximum, as illustrated in Figure~\ref{fig:NFP} (c).
This leads to an intuitive tightness criterion:
\begin{align}
\label{eq:extreme_maintext}
p=\text{argmax}_{q\in \mathcal{N}(p)}\;e(G,q,d)
\end{align}
for some open neighborhood $\mathcal{N}(p)\subset E$  and direction $d\in\mathbb{R}^2$.
The following property arises directly from the Minkowski sum, establishing a geometric connection between tight packing and feasible reference points in $E$:
\begin{lemma}
    If~Equation~(\ref{eq:extreme_maintext}) holds for some open neighborhood $\mathcal{N}(p)$, then:
    \begin{align}
       p=\text{argmax}_{q\in \mathcal{N}(p)} d^Tq
        \label{eq:target_maintext} 
        \end{align}
\end{lemma}
In other words, if $d^Tp$ is a local maximum, i.e., $d^Tp \ge d^Tq$ for all $ q \in \mathcal{N}(p)$, then  $p$ represents the most extreme placement of $G$  in direction $d$ within its neighborhood. 
With this formulation, we can assess the packing tightness of point $p$ by evaluating the range of directions $d$ for which $p$ satisfies Equation~(\ref{eq:target_maintext}).  This range corresponds to the spanning angle of the 2D normal cone \citep{boyd2004convex}.
The normal cone of a set $C$ at a boundary point $x_0$ is the set of all vectors $y$ such that $y^T (x - x_0) \le 0$ for all $x \in C$. Based on this, we define the tightness measure:
\begin{lemma}
For a convex polygonal set $E$ and $p\in\partial E$,  the tightness measure $\psi(p)$ is defined as the spanning angle of the normal cone at $p$, given by $\psi(p)\triangleq  \pi-\theta$, where $\theta$ is the interior angle of $E$ at $p$. 
\end{lemma}

\begin{figure}[h]
    \begin{center}
    \centerline{\includegraphics[width=0.46
    \textwidth]{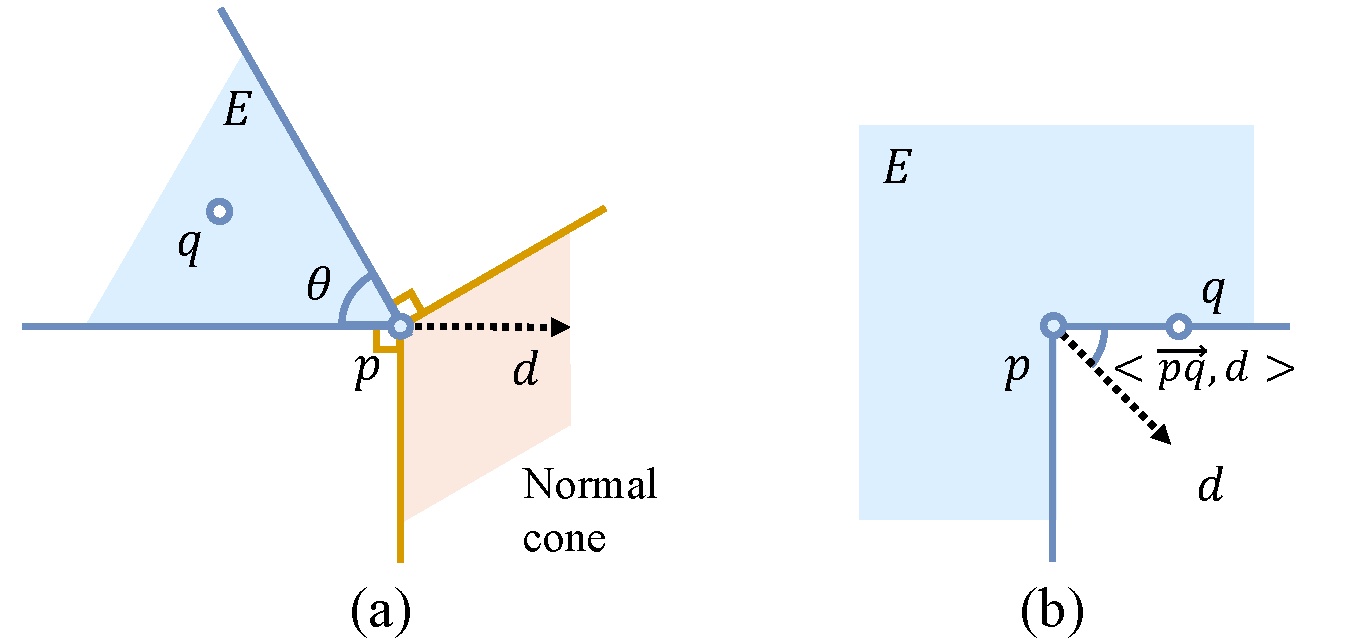}}
    \caption{ 
    \rev{(a): For a boundary point $p$ on a convex cone $E$, any direction $d$ in its normal cone satisfies  $d^Tp \ge d^Tq$, where $q\in E$. (b): If $p$ lies on a concave segment, then for  $\forall d \in \mathbb{R}^2$, there  exists $q\in E$ that the angle $<\vec{pq},d> < \pi/2$, i.e., $d^Tp < d^Tq$. 
    }}
\vspace{-10pt}
\label{fig:proof_maintext}
\end{center}
\end{figure}

Figure~\ref{fig:proof_maintext} (a) illustrates an example of a normal cone.
For concave polygons, points on concave boundaries have no normal cone,  as shown in Figure~\ref{fig:proof_maintext} (b).  We assign  $\psi(p)=0$ in these cases.
Using $\psi(p)$ as a tightness metric, we compare different candidate placements.
 The proof for Theorem~\ref{theorem:convex} is obvious:
\begin{proof}
Normal cones exist only at boundary points, not interior ones.
Consider a neighborhood $\mathcal{N}(p)$ that includes $p$ and its two neighboring half-open edges.
Within $\mathcal{N}(p)$, point  $p$ is convex with  $\psi(p) = \pi  - \theta$.  All other points $q$ 
are with $\psi(q)=0$. 
\end{proof}

\begin{figure}[h]
    \centering
    \includegraphics[width=0.48\textwidth]{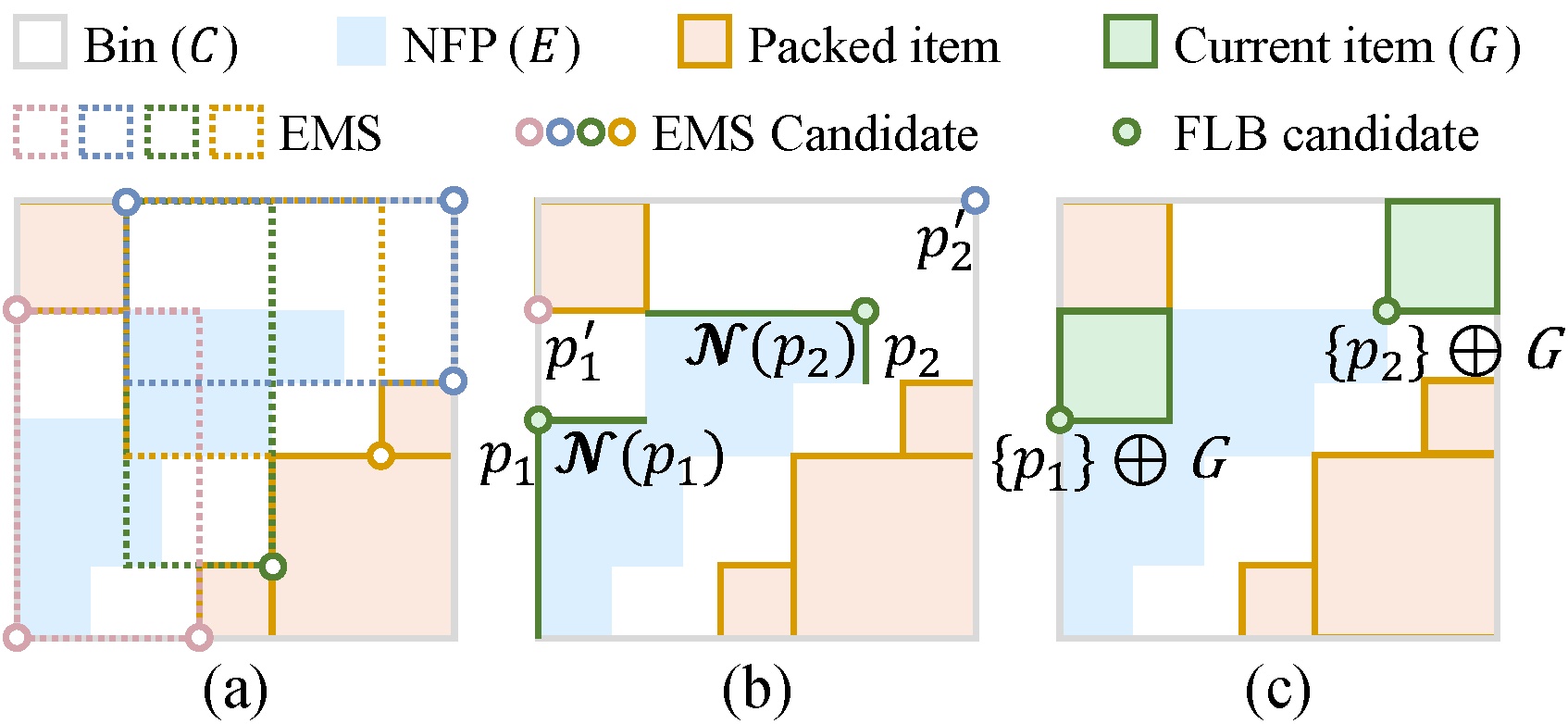}
    \caption{
        \rev{(a): Candidates that are locally optimal under the tightness measure  $\psi$ are covered by the EMSs, with (b) and (c) illustrating their  FLB coordinates and final placements, respectively. $\mathcal{N}(p_1)$ and  $\mathcal{N}(p_2)$ are  the neighborhoods for convex vertices  $p_1$ and $p_2$.
        }
        }
    \label{fig:neighborhood_maintext_new}
\end{figure}

This proof provides a geometric interpretation of why PCT performs well under heuristic-based leaf node expansion schemes.
Some schemes, such as EMS and EV,  operate along the boundaries of already packed items in the $d\in\{x, y\}$ directions. 
The intersecting boundaries cover the convex vertices of the NFP, thereby preserving local optimality while pruning much of the solution space to reduce unnecessary exploration.
As illustrated in  Figure~\ref{fig:neighborhood_maintext_new} (a), each open circle represents an EMS candidate that can be transformed into a locally optimal FLB candidate. Examples of such transformed FLB candidate (filled circle) and their corresponding placements are exhibited in Figure~\ref{fig:neighborhood_maintext_new} (b) and (c), respectively. 
The DRL policy then selects global placements from these high-quality local candidates, enhancing both learning efficiency and performance. 

}

\section{More Results}

\label{section:moreResults}
In this section, we report more results of our method. Section~\ref{section:generalization} further discusses the generalization ability of our method on disturbed item sampling distributions and unseen items. 
Section~\ref{section:behavior} visualizes packing sequences to analyze model behaviors.
Section~\ref{section:cost} reports the running cost of each method. 

\subsection{Generalization Performance}
\label{section:generalization}

\begin{figure*}[h]
    \begin{center}
    \centerline{\includegraphics[width=\textwidth]{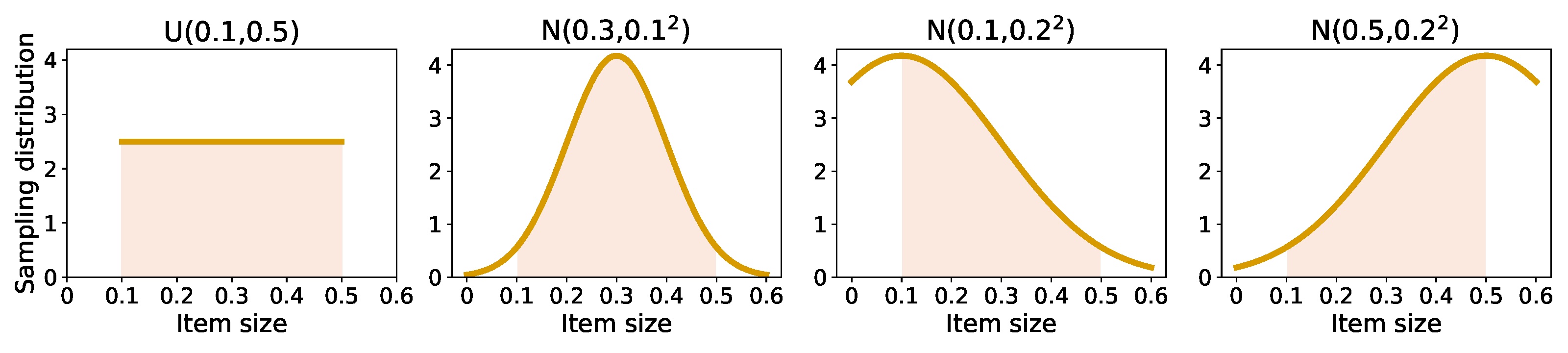}}
    \vspace{-10pt}
    \caption{
        The probability distribution for sampling item sizes. The area of the colored zone is normalized to 1.
    }
    \vspace{-20pt}
\label{fig:different_sample_dist}
\end{center}
\end{figure*}

The generalization ability of learning-based methods has always been a concern.  Here we demonstrate that our method has a good generalization performance on item size distributions different from the training one.
We conduct this experiment with continuous solution space. 
We sample item size $s^d$ from normal distributions $\mathcal{N}(\mu,\sigma^2)$ for generating test sequences where $\mu$ and $\sigma$ are the expectation and the standard deviation.
Three normal distributions are adopted here, namely $\mathcal{N}(0.3, 0.1^2)$, $\mathcal{N}(0.1, 0.2^2)$, and $\mathcal{N}(0.5, 0.2^2)$, as shown in Figure~\ref{fig:different_sample_dist}.
The larger $\mu$ of the normal distribution, the larger the average size of sampled items. We still control $s^d$ within the range of $[0.1, 0.5]$. 
If the sampled item sizes are not within this range, we will resample until they meet the condition.

\input{./table/difDist.tex}

We directly transfer our policies trained on $\mathcal{U}(0.1, 0.5)$ to these new datasets without any fine-tuning. We use the best-performing heuristic method LSAH~\citep{hu2017solving} in Section~\ref{subsection:continous}  as a baseline. 
The test results are summarized in Table~\ref{tab:difDist}. Our method performs well on distributions different from the training one and always surpasses the LSAH method.

We demonstrate that our algorithm has a good generalization performance on disturbed item transitions, i.e., $\mathcal{P}(s_{t+1}|s_t)$.  Generalizing to a new transition is a classic challenge for reinforcement learning \citep{TaylorS09}.
We conduct this experiment in the discrete setting where the item set is finite ($|\mathcal{I}|=125$).
For each item  $n\in\mathcal{I}$, we add a random non-zero disturbance $\delta_i$ on its original sample probability $p_i$, e.g., $p_i=p_i\cdot(1-\delta_i)$.
We normalize the disturbed $p_i$ as the final item sampling probability. 
Note that $\delta_i$ is fixed during sampling one complete sequence. 
We test different ranges of $\delta_i$ and the results are summarized in Table~\ref{tab:Disturbance}.

\input{./table/disturb.tex}

Benefits from the efficient guidance of heuristic leaf node expansion schemes, our method maintains its performance under various amplitude disturbances.
Our method even behaves well with a strong disturbance $\delta_i\in[-100\%,100\%]$ applied, which means some items may never be sampled by some distributions when $\delta_i = 1$ and $p_i\cdot(1-\delta_i)=0$ in a specific sequence.

Beyond experiments on generalization to disturbed distributions, we also test our method with unseen items. 
We conduct this experiment in the discrete setting.
We randomly delete 25 items from $\mathcal{I}$ and train PCT policies with $|\mathcal{I}_{sub}| = 100$. Then we test the trained policies on full $\mathcal{I}$. See Table~\ref{tab:unseen} for results. Our method still performs well on datasets where unseen items exist in all settings.
\input{./table/unseen.tex}

\subsection{Understanding of Model Behaviors}
\label{section:behavior}
\begin{figure*}[t]
    \begin{center}
    \centerline{\includegraphics[width=\textwidth]{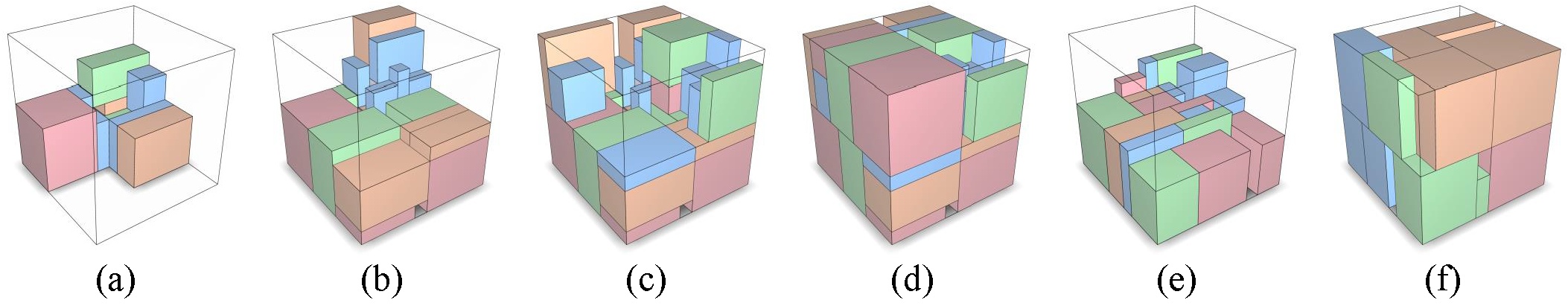}}
    \caption{
        (a)$\sim$(d) Different packing stages of the same sequence. The learned policies assign little items (colored in blue) to gaps and save room for future uncertainty. 
        (e) Online 3D-BPP where load balancing is considered. (f) Online 3D-BPP with isle-friendliness, and different colors mean different item categories.
    }
\vspace{-20pt}
\label{fig:behavior}
\end{center}
\end{figure*}

The qualitative understanding of model behaviors 
is important, especially for practical concerns. We visualize our packing sequences to give our analysis. The behaviors of learned models differ with the packing constraints.
If there is no specific packing preference, our learned policies will start packing near a fixed corner (Figure~\ref{fig:behavior} (a)). 
The learned policies tend to 
combine items of different heights together to form a plane for supporting future ones
(Figure~\ref{fig:behavior} (b)). Meanwhile, it prefers to assign little items to gaps and make room (Figure~\ref{fig:behavior} (c)) for future large ones (Figure~\ref{fig:behavior} (d)).
If additional packing preference is considered, the learned policies behave differently. 
For online 3D-BPP with load balancing, the model will keep the maximum height in the bin as low as possible and pack items layer by layer (Figure~\ref{fig:behavior}(e)).
For online 3D-BPP with isle friendliness, our model tends to pack the same category of items near the same bin corner (Figure~\ref{fig:behavior} (f)). 

\subsection{Performance on ICRA  Stacking Challenge}

The IEEE International Conference on Robotics and Automation (ICRA) organized the Stacking Challenge in 2023~\citep{icra2023stacking}. 
This competition primarily focuses on packing problem variants with \( u > 0 \), including online packing problems with \( p = s = 1 \), forward-looking packing problems with \( p > s = 1 \), and general packing problems with \( p > s > 1 \). 
The bin dimensions are set to \( S_x = 2.141, S_y = 1.076, S_z = 0.99 \).
The item dimensions follow normal distributions: length \( s_x \sim \mathcal{N}(0.45, 0.09^2) \), width \( s_y \sim \mathcal{N}(0.3, 0.05^2) \), and height \( s_z \sim \mathcal{N}(0.17, 0.03^2) \). 
Since the details and code of the participating algorithms have not been publicly released by the organizers, we compare our method with continuous-domain packing algorithms introduced in Section \ref{subsection:continous}.
The test results are summarized in Table~\ref{tab:ICRA_stacking_challange}. In the continuous domain, our method continues to effectively utilize the operational properties of selectable and previewed items for efficient planning, with performance improving as decision variables increases.

\input{./table/icra_stacking_challange.tex}

\subsection{Running Costs}
\label{section:cost}
For 3D-BPP of online needs, the running cost for placing each item is especially important. 
We count the running costs of the experiments in Section~\ref{subsection:action_result} and Section~\ref{subsection:continous} and summarize them in Table~\ref{tab:time}.
Each running cost at time step $t$ is counted from putting down the previous item $n_{t-1}$ until the current item $n_t$ is placed, which includes the time to make placement decisions, the time to check placement feasibility, and the time to interact with the packing environment.
The running cost of our method is comparable to most baselines. Our method can meet real-time packing requirements in both the discrete solution space and continuous solution space.

\input{./table/time.tex}

\subsection{Scalability}
\label{section:scale}

The number of PCT nodes changes constantly with the generation and removal of leaf nodes during the packing process.  
To verify whether our method can solve packing problems with a larger scale $|\textbf{B}|$, 
 we conduct a stress test on $setting\,2$ where the most orientations are allowed and the most leaf nodes are generated. We limit the maximum item sizes $s^d$ to $S^d/5$ so that more items can be accommodated.
We transfer the best-performing policies on $setting\,2$ (trained with EMS) to these new datasets without any fine-tuning. The results are summarized in Table~\ref{tab:scale}.

\input{./table/scale.tex}

PCT size will not grow exponentially with packing scale $|\textbf{B}|$ since invalid leaf nodes will be removed from leaf nodes $\textbf{L}$ during the packing process, both discrete and continuous cases.
For continuous cases,  $|\textbf{L}|$ is more sensitive to $|\textbf{B}|$ due to the diversity of item sizes (i.e.\,$|\mathcal{I}|=\infty$), however, $|\textbf{L}|$ still doesn't explode with $|\textbf {B}|$ and it grows in a sub-linear way.  Our method can execute packing decisions at a real-time speed with controllable PCT sizes, even if the item scale is around two hundred.

\rev{
\subsection{Computational Cost of Attention}

PCT employs attention mechanisms to flexibly encode varying state and action spaces, including Graph Attention Network (GAT)~\citep{VelickovicCCRLB18} and a pointer mechanism~\citep{VinyalsFJ15}.  GAT incurs a computational complexity of \(O(N^2)\), while the pointer mechanism scales as $O(|\textbf{L}_t|)$. Here $N$ is the total number of nodes in the PCT, and $|\textbf{L}_t|$ is the number of leaf nodes at time step $t$.  We present quantitative measurements to illustrate how their computational costs grow with increasing node count, as visualized in Figure \ref{fig:computational_cost}.

During training, the policy launches \(k_p\) parallel packing environments to collect experience and performs a forward rollout of \(k_s\) steps per update. We recommend \(k_p = 64\) and \(k_s = 5\) as a practical setting. Under this 
batch size, both the inference time and memory usage of GAT scale quadratically with the number of nodes, whereas the pointer mechanism scales linearly, as shown in Figure~\ref{fig:computational_cost} (a) and (b). The quadratic cost of GAT also explains the difficulty of directly training packing policies at large scales.
During testing, i.e., \(k_p = k_s = 1\), the running time of both GAT and pointer mechanisms remains stable as the number of nodes increases, as shown in Figure~\ref{fig:computational_cost} (c). This is due to their lightweight design and low CPU memory footprint, as further demonstrated in Figure~\ref{fig:computational_cost} (d), which helps prevent GPU compute unit saturation.

\begin{figure}[!t]
    \centering
    \begin{minipage}{0.50\textwidth}
        \centering
        \subfigure{
            \includegraphics[width=0.475\textwidth, clip, trim=0 0 40 0]{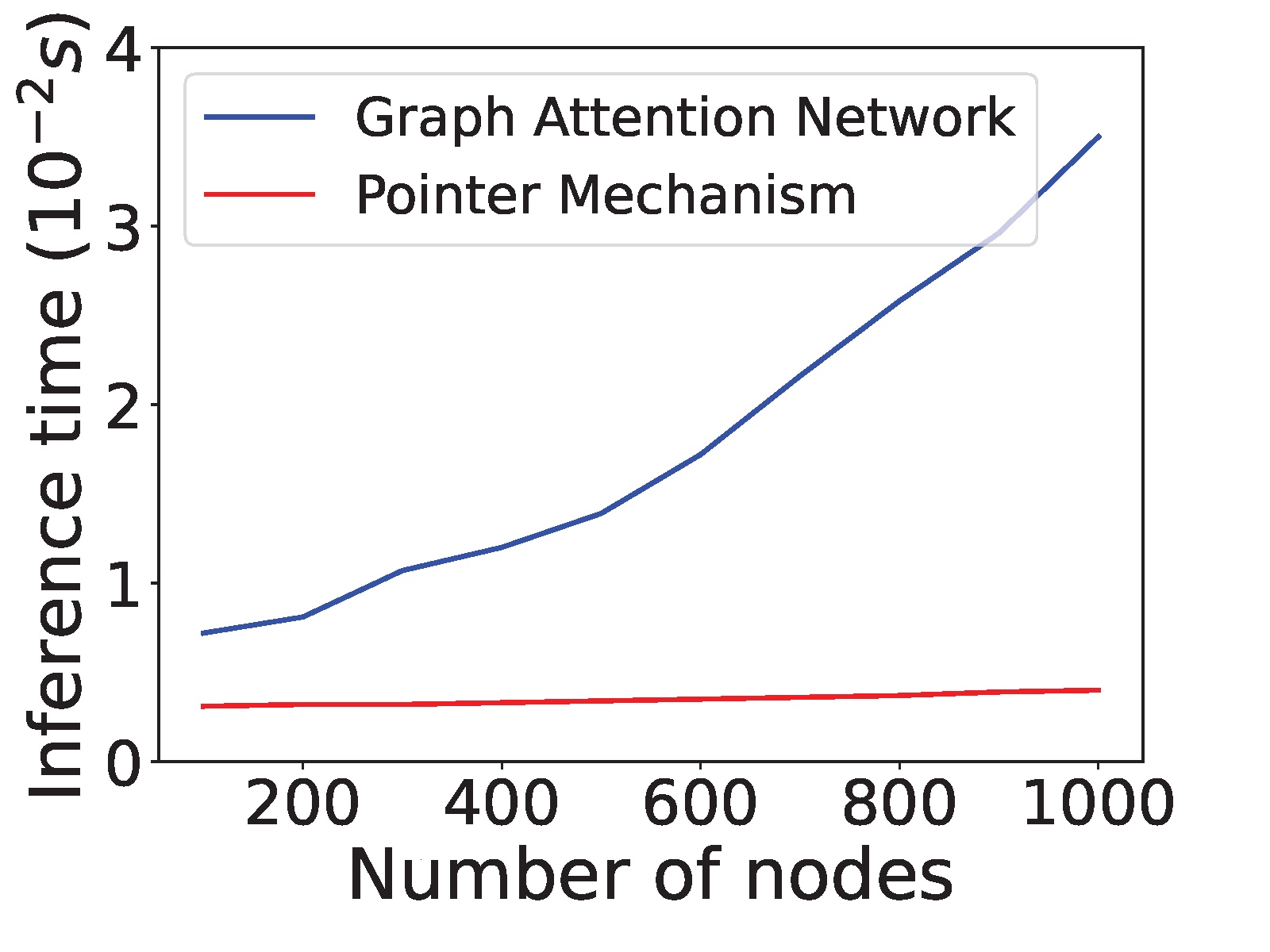}
        }
        \vspace{-4pt} 
        \subfigure{
            \includegraphics[width=0.475\textwidth, clip, trim=0 0 40 0]{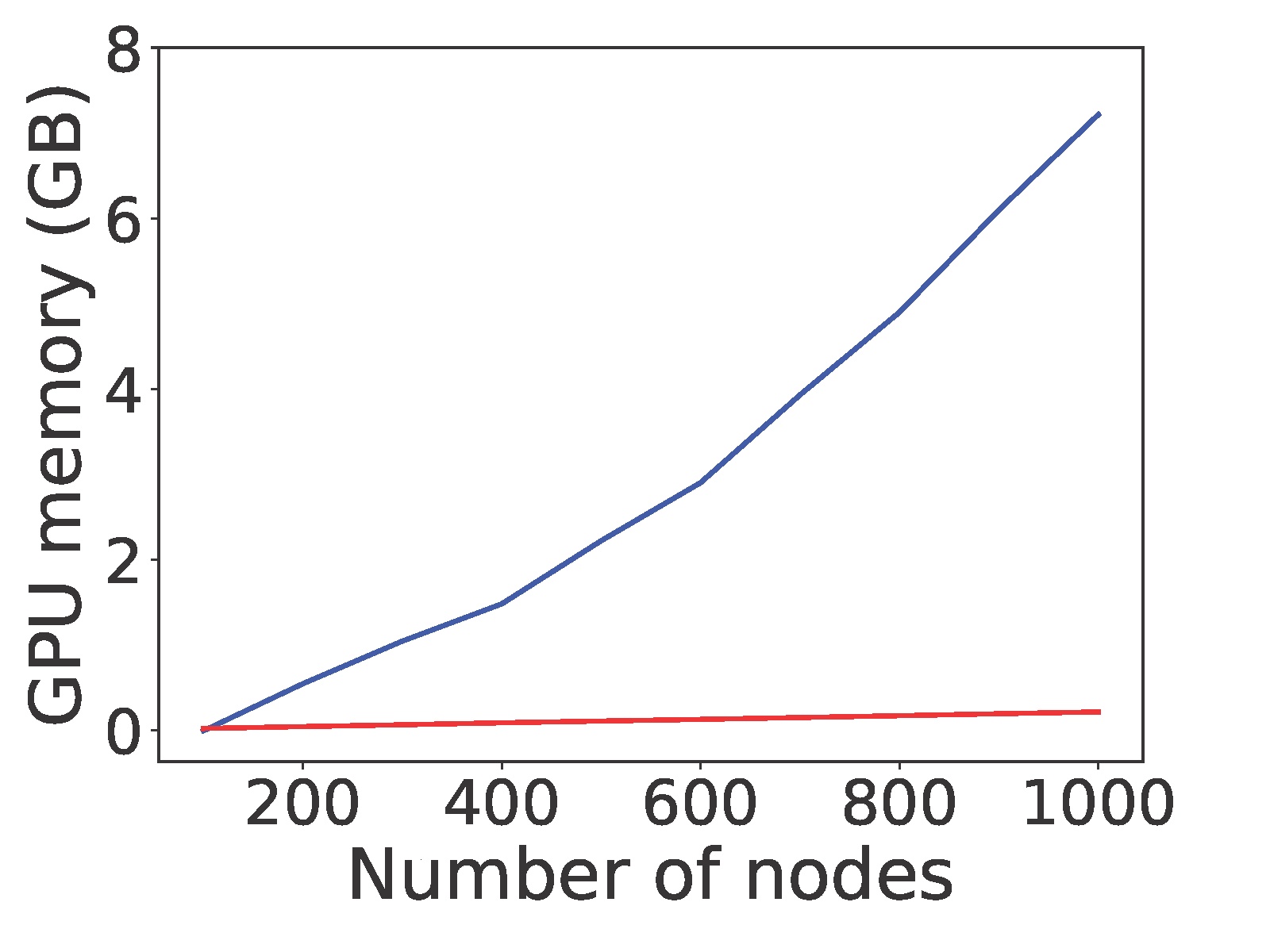}
        }
        \subfigure{
            \includegraphics[width=0.475\textwidth, clip, trim=0 0 40 0]{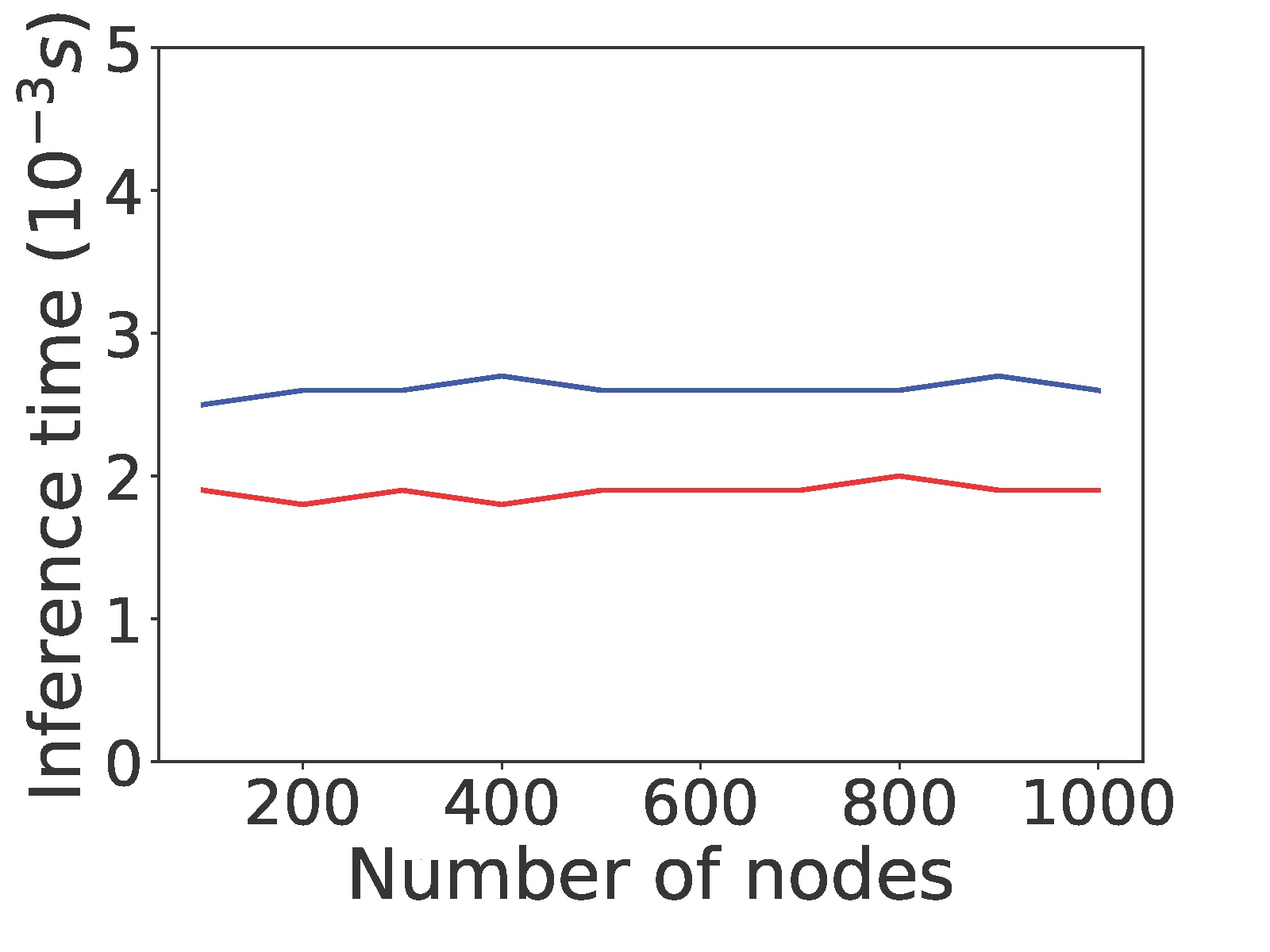}
        }
        \subfigure{
            \includegraphics[width=0.475\textwidth, clip, trim=0 0 40 0]{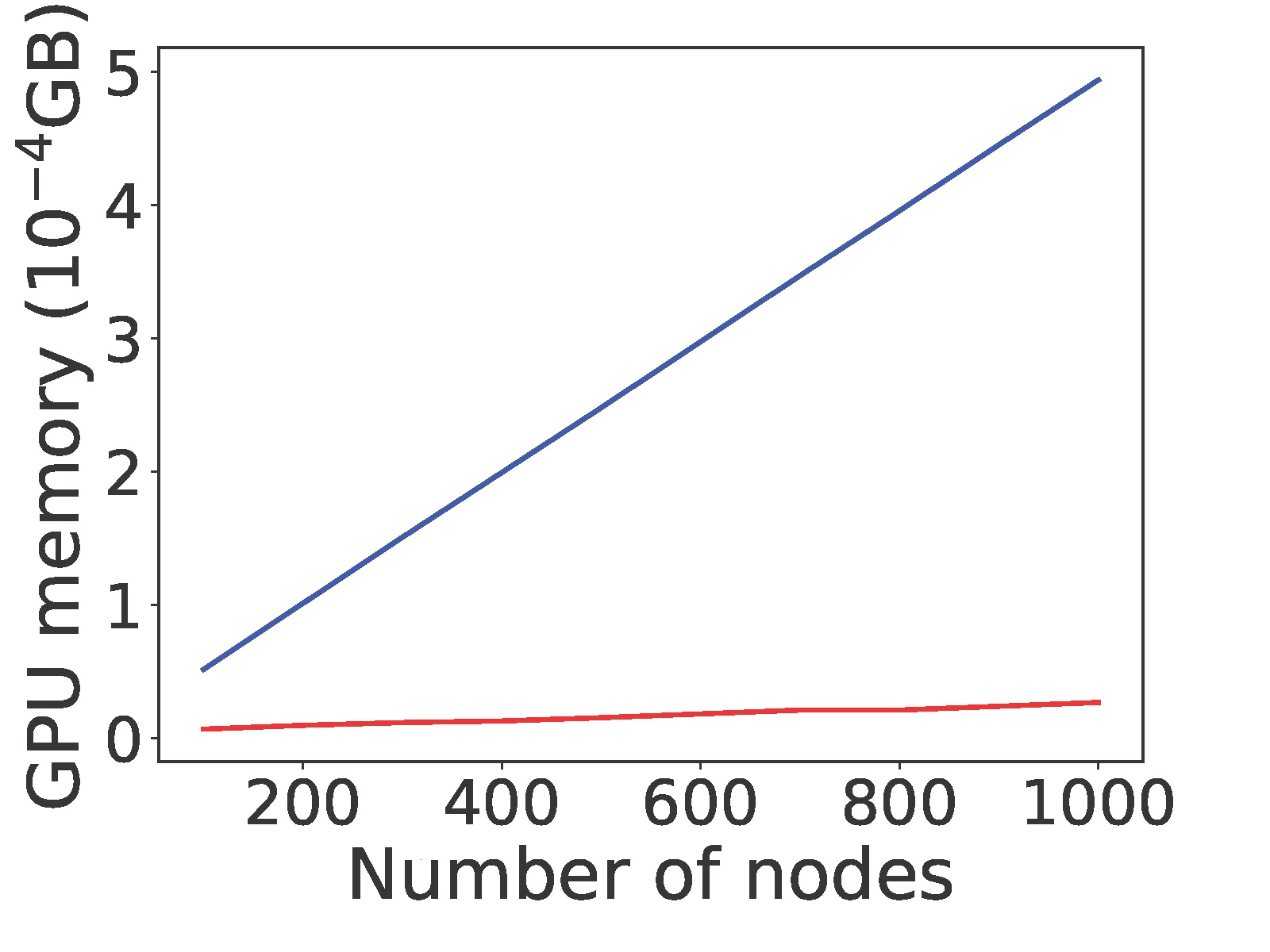}
        }
    \end{minipage}
    \put(-180, 7){\makebox(0,0){\footnotesize (a) Inference time for training }} 
    \put(-56,  7){\makebox(0,0){\footnotesize (b) Memory cost for training}}
    \put(-180,-100){\makebox(0,0){\footnotesize (c) Inference time for test }}
    \put(-56, -100){\makebox(0,0){\footnotesize (d) Memory cost for test}}
    \caption{\rev{Computational cost of attention mechanisms. 
    The reported node count refers to the total node number in PCT for GAT, and to the leaf node number for the pointer mechanism.
    }
    }
    \label{fig:computational_cost}
\end{figure}

}

\rev{
\subsection{Sensitivity of Constraint Reward Weights }
\label{section:sensitivity}
We introduce constraint rewards ${f}(\cdot)$ into the packing reward function to handle additional complex constraints, as described in Section~\ref{subsection:morecomplex}, forming a multi-objective optimization problem. 
The item weight is revised as:   
\begin{equation} \label{eq:reward_weight} 
    w_t = \text{max}(0, v_t + c \cdot \hat{f}(\cdot)),
\end{equation}
 with  $\hat{f}(\cdot)$ representing the normalized constraint reward and \( c \) being a constant to balance the importance of the constraint reward with respect to space utilization.
We perform a parameter sensitivity analysis for \( c \in \{0.1, 1.0, 10.0 \} \), 
corresponding to constraint rewards weighted below, the same level as, or above the space utilization objective. 
Meanwhile, we remove the max operator in Equation~(\ref{eq:reward_weight}) so that both rewards can function at full effect.  The results are summarized in Table~\ref{tab:sensitivity}.

\input{./table/parameter_analysis.tex}

We can observe that increasing the weight of constraint rewards inevitably reduces space utilization.
Some constraints conflict mildly with this objective.
For example, \textit{Kinematic Constraint} aims to minimize the impact of placed items on subsequent robot motions by favoring horizontally distal placements for early items.
Optimizing this placement order has a relatively minor impact on space utilization, which reaches the highest utilization at $c = 10.0$. 
 In contrast, for strongly conflicting constraints like \textit{Load-bearing Constraint}, where minimizing per-object load can be achieved by placing fewer items, increasing $c$ leads to a more significant utilization decline, i.e., the  lowest utilization at $c = 10.0$.
 }

\rev{
\section{Additional Real-World Specifications}
\label{subsection:Hardware}

We adopt a modular gripper that can actively adjust its shape to avoid robot-object collisions.
The gripper comprises five grasping modules of varying sizes, capable of reconfiguring themselves based on the dimensions of the target box, as shown in Figure~\ref{fig:gripper_design} (a). Each module is equipped with multiple suction cups, totaling 165 units (each with a diameter of 39 mm),  providing a maximum suction force of $260kg/m^2$.
The central module is fixed in place and supports the vertical motion of the four surrounding modules, as demonstrated in Figure~\ref{fig:gripper_design} (b). 
These movable modules are actuated by rodless pneumatic cylinders, allowing them to be raised or lowered as needed.
The cylinders are powered by vacuum pumps, which are also shared to generate the suction force for gripping items, thereby eliminating the need for an additional power source for gripper adjustment.

To further prevent collisions in real-world factories,  the robot avoids moving directly above the target position to perform a vertical drop. Instead, it follows a simplified yet effective placement manner by
planning an oblique insertion point as an intermediate waypoint along the motion trajectory, as illustrated in Figure~\ref{fig:motion_planning}.  
This oblique approach reduces the risk of collisions during vertical descent caused by interference from adjacent boxes.
To determine valid oblique insertion directions, we examine eight candidates $d \in \{(c_1, c_2, 0)\}$ with $c_1, c_2 \in \{-1, 0, 1\}$, subject to the constraint $c_1 \neq 0$ or $c_2 \neq 0$. Directions without obstruction from taller boxes are considered valid, and we prioritize those with non-zero values in both axes (i.e., $c_1 \cdot c_2 \neq 0$) 
to maximize spatial clearance.

\begin{figure}
    \centering
    \includegraphics[width=\linewidth]{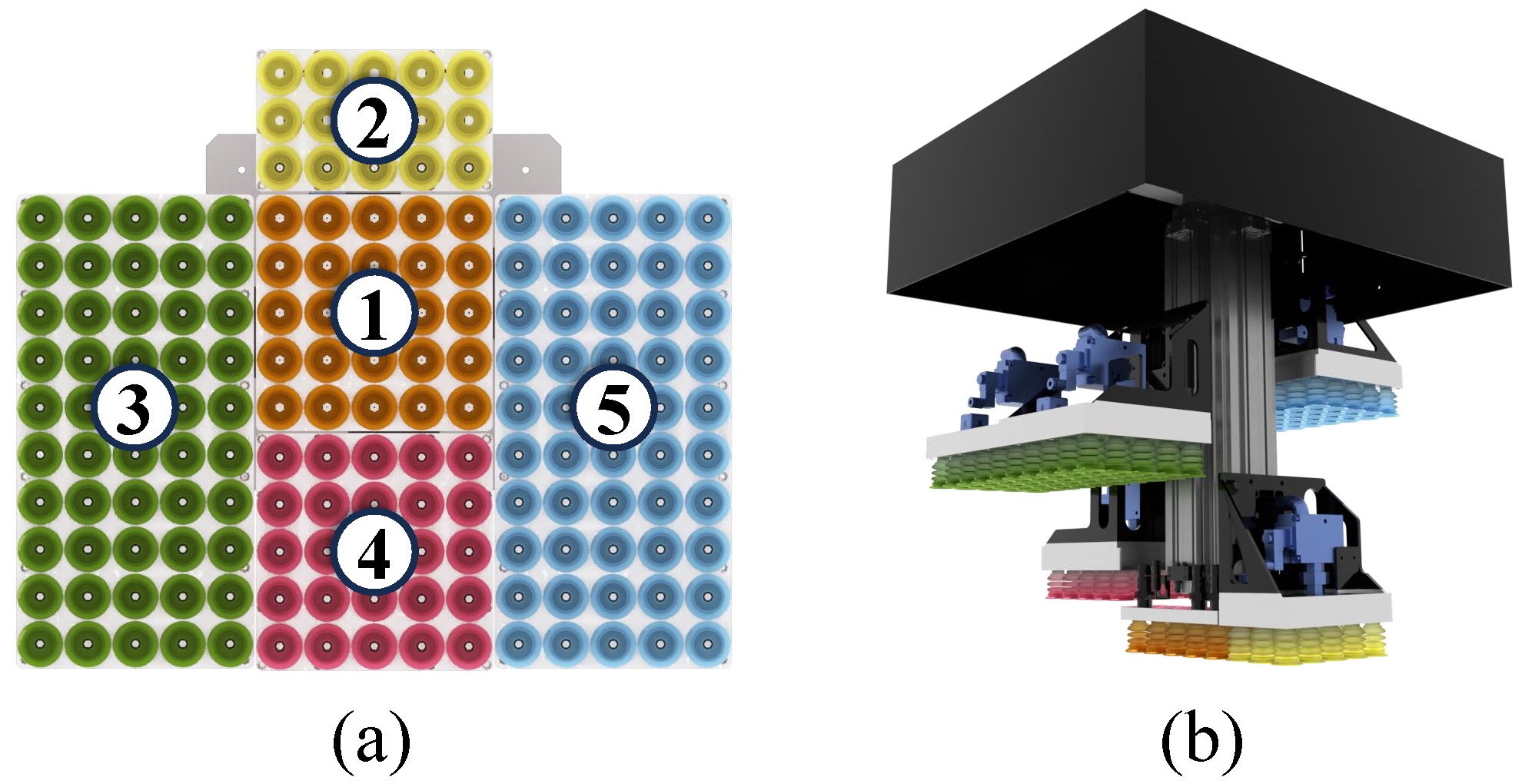}
    \caption{\rev{The modular designed gripper for constraint placement in packing scenarios. This flexible gripper consists of five suction cup modules (a) which can move vertically and online adjust its shape (b) based on the target item size. The central module  1 is fixed and serves as a guide, supporting the vertical movement of the surrounding modules.}}
    \label{fig:gripper_design}
\end{figure}

\begin{figure}
    \centering
    \includegraphics[width=\linewidth]{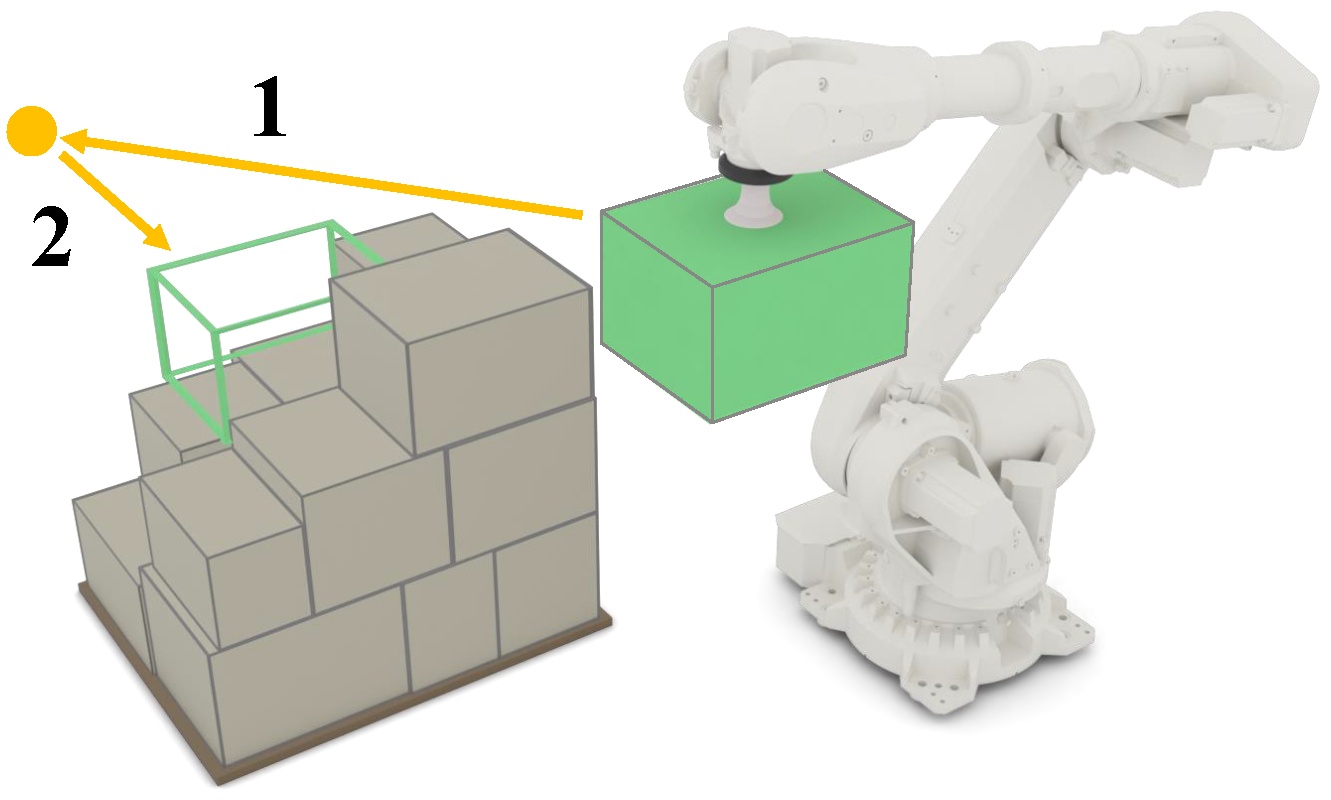}  
    \caption{\rev{Simplified robot packing trajectory. The green box is to be placed into the target placement outlined by the green bounding box. The yellow circle marks the waypoint for oblique insertion. The robot first moves to this waypoint, then proceeds to the final placement.}
    }
    \label{fig:motion_planning}
\end{figure} 

}

\section{Visualized Results}
\label{subsection:visulization}
\begin{figure*}
    \centering
    \includegraphics[width=0.9\linewidth]{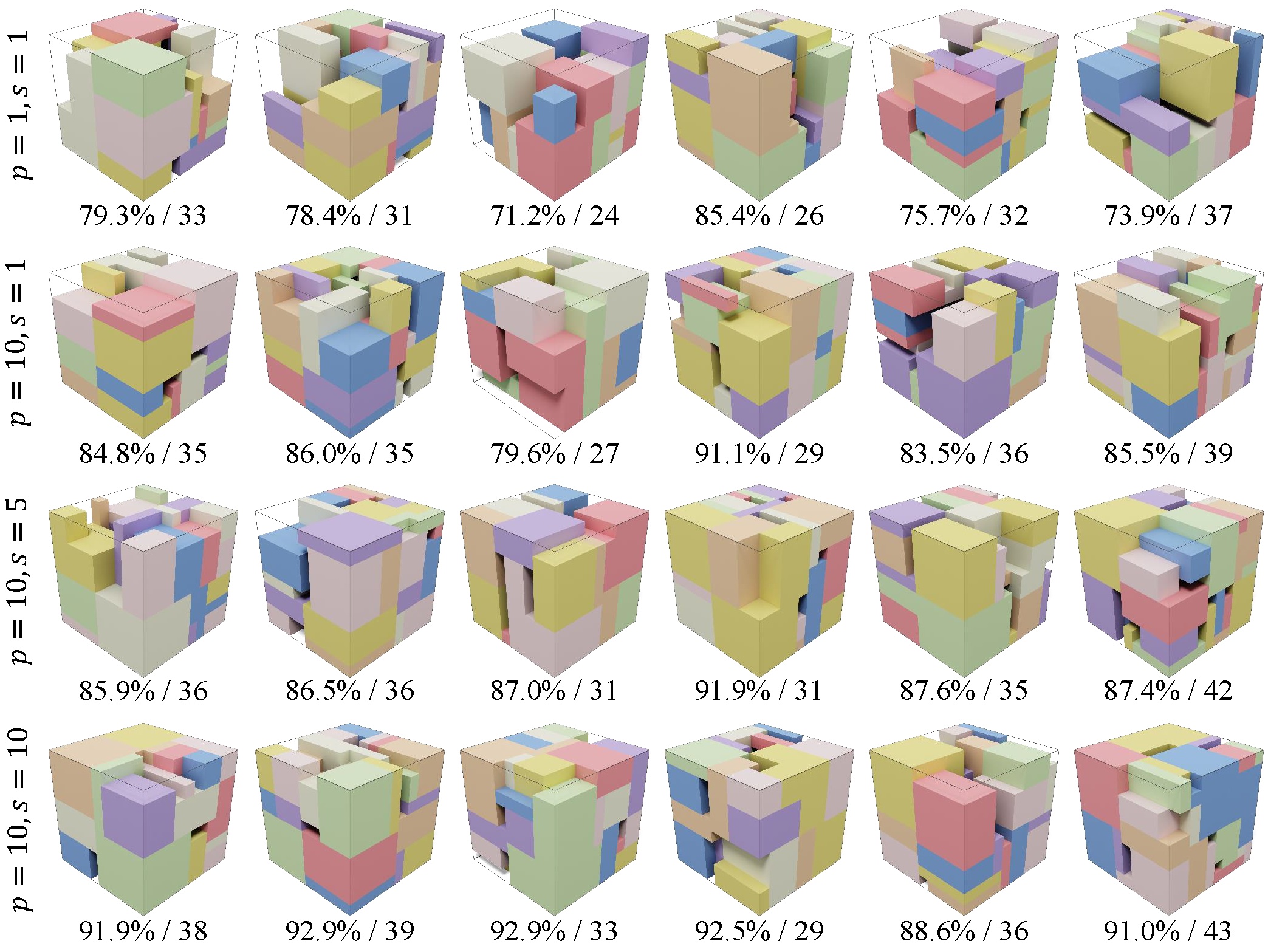}
    \caption{Visualized results of different BPP variations on setting 1, with space utilization and packed item number labeled below. Each column corresponds to the same test data.}
    \label{fig:packing_variants_vis_1}
\end{figure*}

\begin{figure*}
    \centering
    \includegraphics[width=0.9\linewidth]{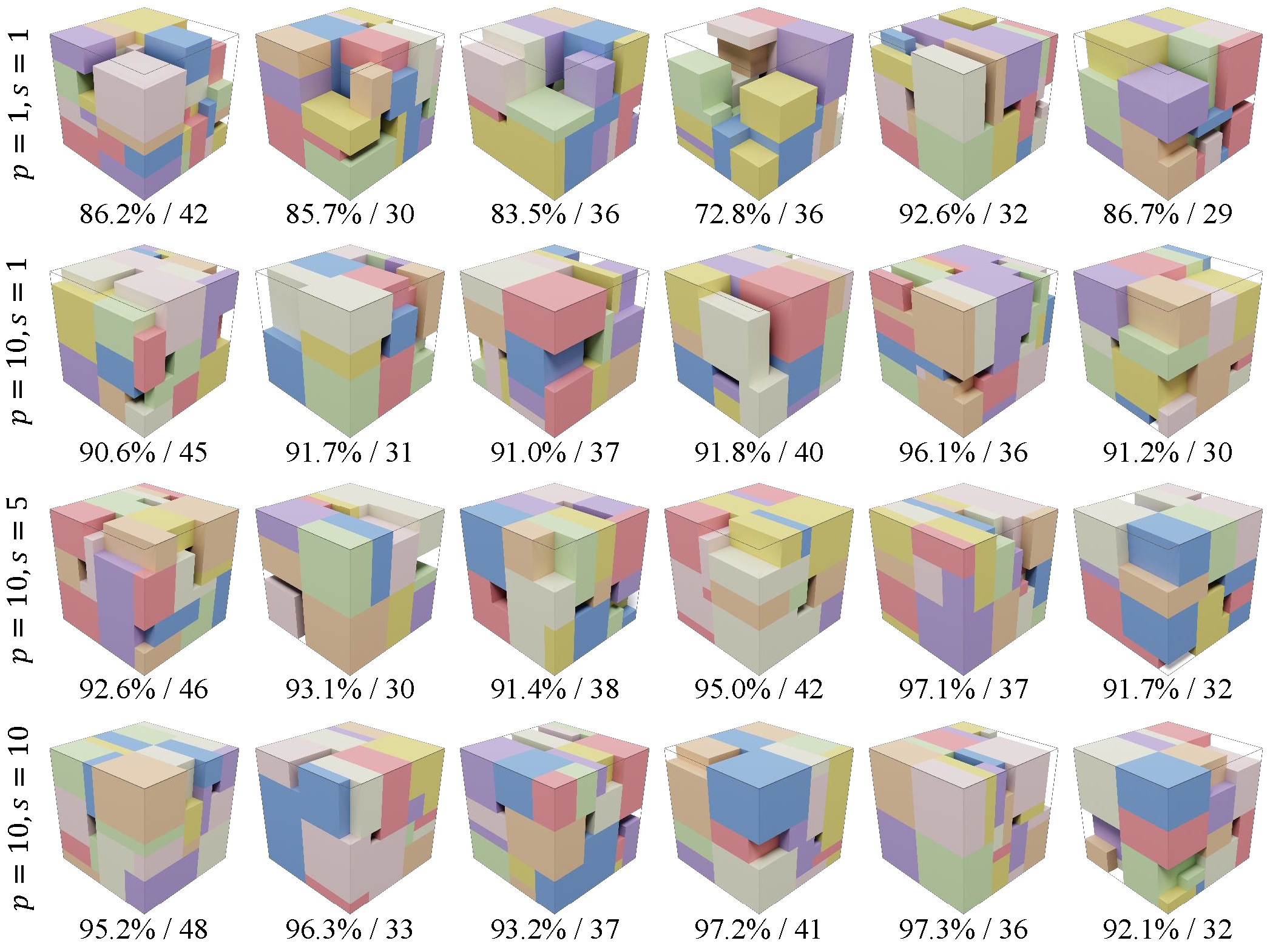}
    \caption{Visualized results of different BPP variations on setting 2. Each column corresponds to the same test data.}
    \label{fig:packing_variants_vis_2}
\end{figure*}

\begin{figure*}
    \centering
    \includegraphics[width=0.9\linewidth]{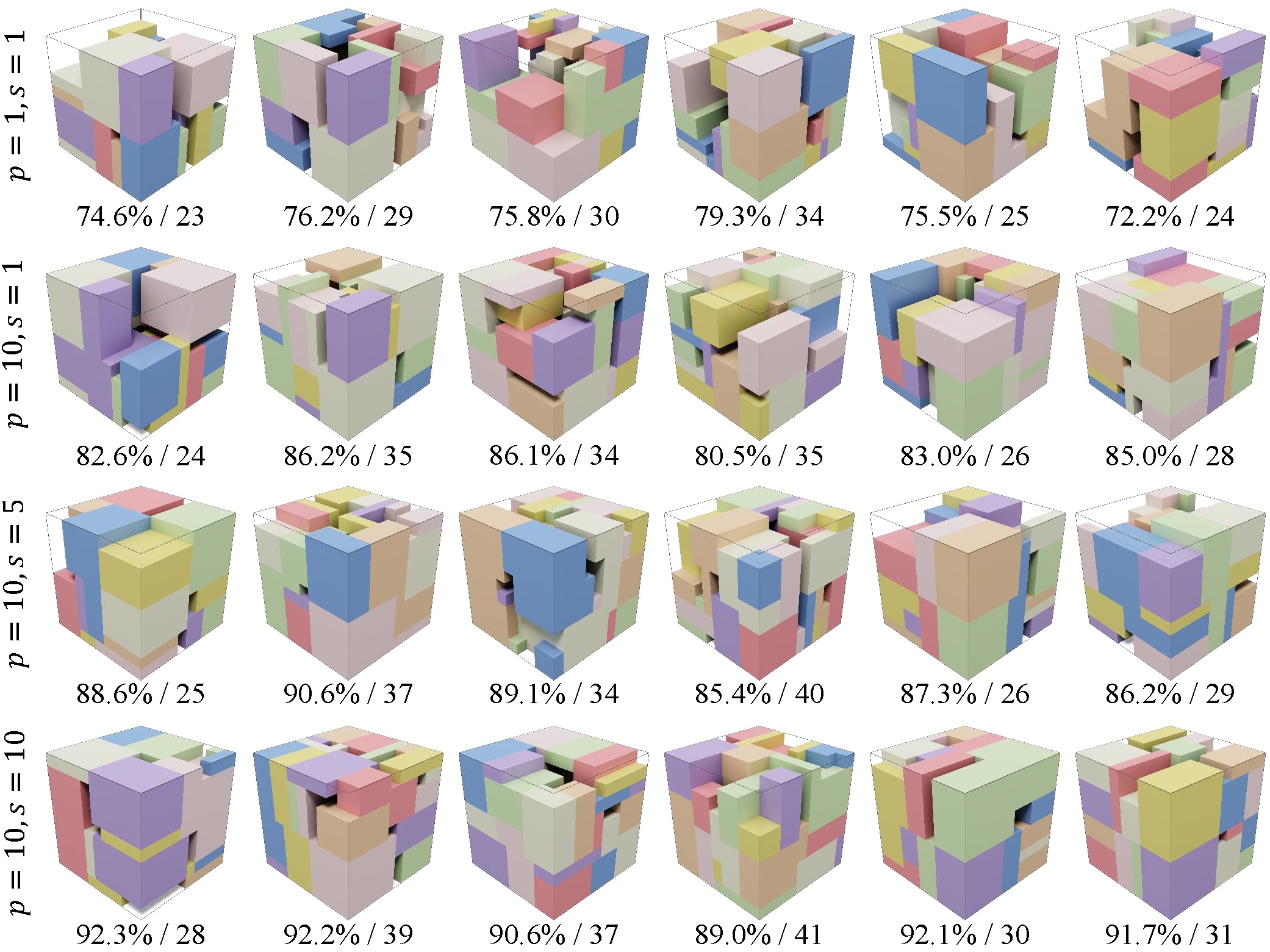}
    \caption{Visualized results of different BPP variations on setting 3.
    Each column corresponds to the same test data.}
    \label{fig:packing_variants_vis_3}
\end{figure*}

\begin{figure*}
    \centering
    \includegraphics[width=0.9\linewidth]{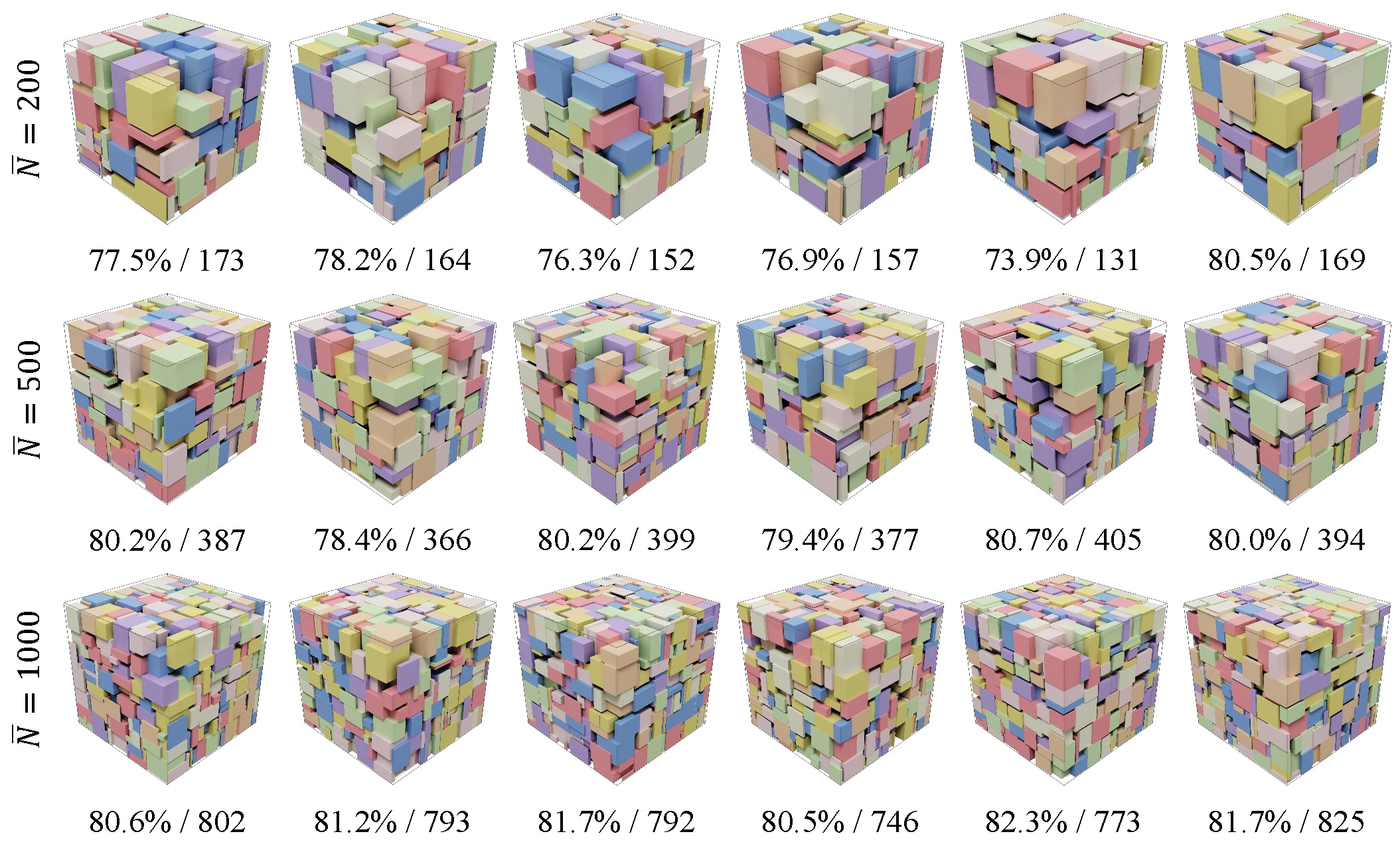}
    \caption{Visualized results of large-scale packing at various problem scales, with space utilization and item number labeled.}
    \label{fig:packing_large_scale}
\end{figure*}

\begin{figure*}
    \centering
    \includegraphics[width=0.9\linewidth]{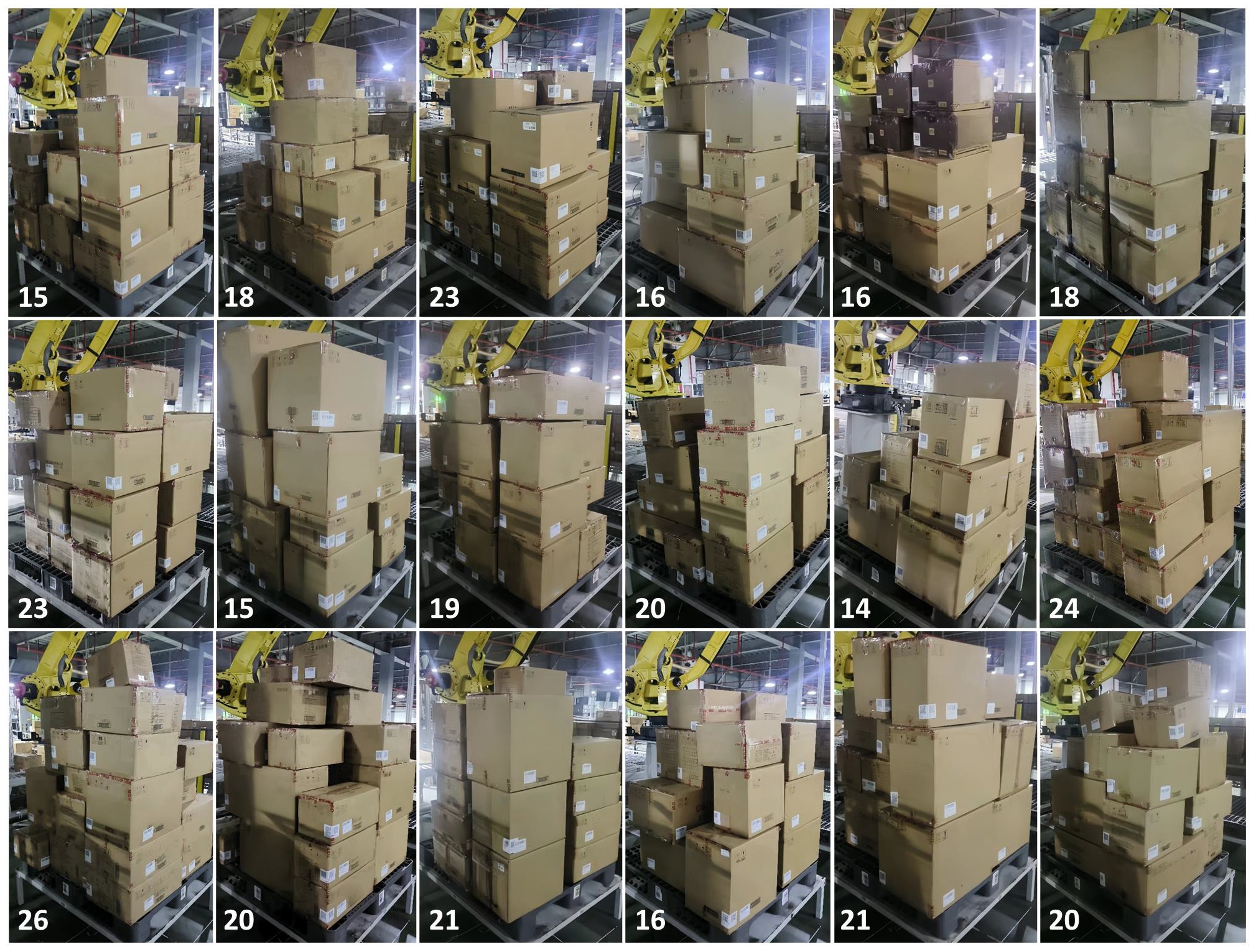}
    \caption{Real-world packing results. The number of packed items on each pallet is labeled in the bottom left corner.}
    \label{fig:real_world_packing}
\end{figure*}

We visualize the experimental results of different BPP variations on three settings in Figure~\ref{fig:packing_variants_vis_1}, Figure~\ref{fig:packing_variants_vis_2}, and Figure~\ref{fig:packing_variants_vis_3}. 
It is clearly observed that as the number of packing decision variables increases, the packing results become more compact.
We also provide the visualized results of large-scale packing in Figure~\ref{fig:packing_large_scale}.
Each plot is about results tested by a randomly generated item sequence.
The real-world packing results conducted via industrial production are provided in Figure~\ref{fig:real_world_packing}.

\clearpage

%% file: table/difDist.tex
\begin{table*}[ht!]
    \caption{ Generalization performance on different kinds of item sampling distributions. 
    }
    \label{tab:difDist}
    \centering
    \footnotesize
    \resizebox{1\textwidth}{!}{
    \begin{tabular}{l|l|ccc|ccc|ccc}
    \toprule
        \multirow{2}{*}{Test Distribution}  & \multirow{2}{*}{Method} &  \multicolumn{3}{c|}{$Setting\,1$} & \multicolumn{3}{c|}{$Setting\,2$} & \multicolumn{3}{c}{$Setting\,3$} \\
        &  &  \multicolumn{1}{c}{Uti. $\uparrow$} &  \multicolumn{1}{c}{Var. $\downarrow$} & \multicolumn{1}{c|}{Num. $\uparrow$}  & \multicolumn{1}{c}{Uti. $\uparrow$} &  \multicolumn{1}{c}{Var. $\downarrow$} & \multicolumn{1}{c|}{Num. $\uparrow$}  & \multicolumn{1}{c}{Uti. $\uparrow$} &  \multicolumn{1}{c}{Var. $\downarrow$}  & \multicolumn{1}{c}{Num. $\uparrow$} \\
       \midrule
    & LSAH  & $48.3\%$ &$12.1$& $18.7$ &  $58.7\%$ &$4.6$& $22.8$ &  $48.4\%$ &$12.2$& $18.8$  \\
    $s^d\sim U(0.1, 0.5)$ & PCT \& EMS  &   $65.3\%$ &$4.4$& $24.9$ & $\textbf{66.3\%}$ &$\textbf{2.3}$& $\textbf{27.0}$ & $\textbf{66.6\%}$ &$\textbf{3.3}$& $\textbf{25.3}$  \\
    & PCT \& EV  &   $\textbf{65.4\%}$ &$\textbf{3.3}$ & $\textbf{25.0}$ & $65.0\%$ &$2.6$& $26.4$ & $65.8\%$ & $3.6$ & $25.1$  \\
    \cmidrule{1-11} 
    & LSAH  &   $49.2\%$ &$11.1$& $18.9$ & $60.0\%$ &$4.1$& $22.9$ & $49.2\%$ &$11.0$& $18.9$  \\
    $s^d\sim N(0.3, 0.1^2)$ &  PCT \& EMS  &   $\textbf{66.1\%}$ &$3.6$& $\textbf{25.1}$ & $\textbf{64.3\%}$ &$3.5$& $\textbf{25.6}$ & $\textbf{66.4\%}$ &$3.0$& $\textbf{25.2}$  \\
    & PCT \& EV  &   $65.1\%$ &$\textbf{2.8}$ & $24.7$ & $63.7\%$ &$\textbf{2.6}$& $25.3$ & $66.2\%$ & $\textbf{2.9}$ & $25.1$  \\
    \cmidrule{1-11} 
    & LSAH  &   $52.4\%$ &$8.9$& $30.3$ & $62.9\%$ &$\textbf{2.4}$& $44.3$ & $52.3\%$ &$8.9$& $30.2$  \\
    $s^d\sim N(0.1, 0.2^2)$ &  PCT \& EMS  &   $\textbf{68.5\%}$ &$\textbf{2.5}$& $\textbf{39.0}$ & $\textbf{66.4\%}$ &$3.0$& $\textbf{49.7}$ & $\textbf{69.2\%}$ &$2.5$& $\textbf{39.4}$  \\
    &  PCT \& EV  &   $66.5\%$ &$2.7$ & $38.0$ & $64.9\%$ &$2.7$& $48.3$ & $67.4\%$ & $\textbf{2.4}$ & $38.5$  \\
    \cmidrule{1-11} 
    & LSAH  &   $47.3\%$ &$12.6$& $13.0$ & $56.0\%$ &$5.5$& $12.9$ & $47.3\%$ &$12.6$& $13.0$  \\
    $s^d\sim N(0.5, 0.2^2)$ & PCT \& EMS  &   $63.5\%$ &$5.0$& $17.3$ & $\textbf{64.5\%}$ &$\textbf{2.8}$& $\textbf{15.4}$ & $\textbf{65.2\%}$ &$3.8$& $\textbf{17.7}$  \\
    & PCT \& EV  &   $\textbf{65.1\%}$ &$\textbf{3.3}$ & $\textbf{17.7}$ & $64.5\%$ &$2.8$& $15.3$ & $65.1\%$ & $\textbf{3.7}$ & $17.7$  \\
    \bottomrule
    \end{tabular}
    }
\end{table*}

%% file: table/disturb.tex
\begin{table*}[ht!]
    \caption{
        Transfer the best-performing PCT policies directly to the disturbed item sampling distributions.    
     ``Dif."$\,$means how much the generalization performance drops from the undisturbed case ($\delta_i=0$).}
    \label{tab:Disturbance}
    \centering
    \footnotesize
    \resizebox{1\textwidth}{!}{
    \begin{tabular}{l|cccc|cccc|cccc}
        \toprule
        \multirow{2}{*}{Disturbance} &  \multicolumn{4}{c|}{$Setting\,1$} & \multicolumn{4}{c|}{$Setting\,2$} & \multicolumn{4}{c}{$Setting\,3$} \\
         &  \multicolumn{1}{c}{Uti. $\uparrow$} &  \multicolumn{1}{c}{Var. $\downarrow$} & \multicolumn{1}{c}{Num. $\uparrow$} & \multicolumn{1}{c|}{Dif. $\downarrow$} & \multicolumn{1}{c}{Uti. $\uparrow$} &  \multicolumn{1}{c}{Var. $\downarrow$} & \multicolumn{1}{c}{Num. $\uparrow$} & \multicolumn{1}{c|}{Dif. $\downarrow$} & \multicolumn{1}{c}{Uti. $\uparrow$} &  \multicolumn{1}{c}{Var. $\downarrow$}  & \multicolumn{1}{c}{Num. $\uparrow$} & \multicolumn{1}{c}{Dif. $\downarrow$} \\
       \midrule
    $\delta_i = 0$ & $76.0\%$ &$4.2$& $29.4$ & $0.0\%$ & $86.0\%$ &$1.9$& $33.0$ & $0.0\%$ & $75.7\%$ &$4.6$& $29.2$ & $0.0\%$ \\
    $\delta_i\in[-20\%, 20\%]$ &  $75.6\%$ &$4.6$& $29.1$ & $0.5\%$ & $85.7\%$& $2.1$ & $32.8$ & $0.3\%$  & $75.3\%$ &$4.5$& $29.0$ & $0.5\%$ \\
    $\delta_i\in[-40\%, 40\%]$ &  $75.5\%$ &$4.5$& $29.0$ & $0.7\%$ & $85.6\%$& $2.1$ & $32.8$ & $0.5\%$  & $75.6\%$ &$4.8$& $29.3$ & $0.1\%$ \\
    $\delta_i\in[-60\%, 60\%]$ &  $75.5\%$ &$4.3$& $28.9$ & $0.7\%$ & $85.8\%$& $2.1$ & $32.8$ & $0.2\%$  & $75.5\%$ &$4.8$& $28.9$ & $0.3\%$ \\
    $\delta_i\in[-80\%,80\%]$ &  $75.7\%$ &$4.5$& $29.2$ & $0.4\%$ & $85.6\%$& $2.2$ & $32.9$ & $0.5\%$  & $75.4\%$ &$4.9$& $29.3$ & $0.4\%$ \\
    $\delta_i\in[-100\%,100\%]$ &  $75.8\%$ &$4.4$& $29.0$ & $0.3\%$ & $85.5\%$& $2.2$ & $32.6$ & $0.6\%$  & $75.3\%$ &$4.7$& $29.3$ & $0.5\%$ \\
    \bottomrule
    \end{tabular}
    }
    \end{table*}

%% file: table/unseen.tex
\begin{table*}[h]
    \caption{Generalization performance on unseen items. 
    }
    \label{tab:unseen}
    \centering
    \footnotesize
    \resizebox{1\textwidth}{!}{
    \begin{tabular}{l|l|rrr|rrr|rrr}
        \toprule
        \multirow{2}{*}{Train} & \multirow{2}{*}{Test} &  \multicolumn{3}{c|}{$Setting\,1$} & \multicolumn{3}{c|}{$Setting\,2$} & \multicolumn{3}{c}{$Setting\,3$} \\
        &  &  \multicolumn{1}{c}{Uti. $\uparrow$} &  \multicolumn{1}{c}{Var. $\downarrow$} & \multicolumn{1}{c|}{Num. $\uparrow$}  & \multicolumn{1}{c}{Uti.$\uparrow$} &  \multicolumn{1}{c}{Var. $\downarrow$} & \multicolumn{1}{c|}{Num. $\uparrow$}  & \multicolumn{1}{c}{Uti. $\uparrow$} &  \multicolumn{1}{c}{Var. $\downarrow$}  & \multicolumn{1}{c}{Num.$\uparrow$} \\
       \midrule
    $|\mathcal{I}| = 125$ &  $|\mathcal{I}|  = 125$  &   $76.0\%$ &$4.2$ & $29.4$ & $85.3\%$ &$2.1$& $32.8$ & $75.7\%$ & $4.6$ & $29.2$  \\
    $|\mathcal{I}_{sub}| = 100$ & $|\mathcal{I}_{sub}|  = 100$  & $74.4\%$ &$5.1$& $29.4$ &  $86.3\%$ &$1.7$& $33.8$ &  $74.2\%$ &$4.7$& $29.3$  \\
    $|\mathcal{I}_{sub}| = 100$ & $|\mathcal{I}|  = 125$ & $74.6\%$ &$5.4$& $28.9$ &  $85.6\%$ &$2.6$& $33.0$ &  $74.4\%$ &$5.2$& $28.8$  \\
    \bottomrule 
    \end{tabular}
    }
    \end{table*}

%% file: table/icra_stacking_challange.tex
\begin{table*}[ht!]
    \caption{Performance comparisons on ICRA stacking challange benchmark~\citep{icra2023stacking}.}
    \label{tab:ICRA_stacking_challange}
    \centering
    \footnotesize
    \setlength{\tabcolsep}{0.36em}
    \resizebox{1\textwidth}{!}{
    \begin{tabular}{l|ccc|cccc|cccc|cccc}
        \toprule
        \multirow{2}{*}{Method} & \multirow{2}{*}{Prev.} & \multirow{2}{*}{Sel.} & \multirow{2}{*}{Un.} &  \multicolumn{4}{c|}{$Setting\,1$} & \multicolumn{4}{c|}{$Setting\,2$} & \multicolumn{4}{c}{$Setting\,3$} \\
        & & & &  \multicolumn{1}{c}{Uti. $\uparrow$} &  \multicolumn{1}{c}{Var. $\downarrow$} & \multicolumn{1}{c}{Num. $\uparrow$} & \multicolumn{1}{c|}{Gap $\downarrow$} & \multicolumn{1}{c}{Uti. $\uparrow$} &  \multicolumn{1}{c}{Var. $\downarrow$} & \multicolumn{1}{c}{Num. $\uparrow$} & \multicolumn{1}{c|}{Gap $\downarrow$} & \multicolumn{1}{c}{Uti. $\uparrow$} &  \multicolumn{1}{c}{Var. $\downarrow$}  & \multicolumn{1}{c}{Num. $\uparrow$} & \multicolumn{1}{c}{Gap $\downarrow$} \\
    \midrule
    BR  & $p=1$ & $s=1$  & $u>0$ &  $53.4\%$ & $2.7$ & $53.4$ & $ 16.3\%$ & $ 61.6\%$ &$0.7$& $61.6$ & $ 12.7\%$  & $53.7\%$ &$2.5$& $53.7$ & $ 17.1\%$ \\
    OnlineBPH  & $p=1$ & $s=1$  & $u>0$ &  36.4$\%$ & $12.5$ & $36.4$ & $ 42.9\%$ & $ 46.7\%$ &$1.3$& $46.8$ & $ 33.9\%$  & $ 36.7\%$ &$ 12.7$& $36.8$ & $ 43.4\%$ \\
    LSAH  & $p=1$ & $s=1$  & $u>0$ &   $43.6\%$ & $11.7$ & $43.6$ & $ 31.7\%$ & $ 66.4\%$ &$0.6$& $66.4$ & $ 5.9\%$  & $43.9\%$ &$11.9$& $43.9$ & $ 32.3\%$ \\
    PCT   & $p=1$ & $s=1$  & $u>0$ &  $\textbf{63.8\%}$ & $\textbf{1.5}$ & $\textbf{63.6}$ & $ \textbf{0.0\%}$ & $ \textbf{70.6\%}$ &$\textbf{0.3}$& $\textbf{70.5}$ & $ \textbf{0.0\%}$  & $ \textbf{64.8\%}$ &$\textbf{2.1}$& $ \textbf{64.7}$ & $ \textbf{0.0\%}$ \\
    \midrule
    ToP & $p=10$ & $s=1$  & $u>0$ &   $70.2\%$ & $0.4$ & $70.8$ & $-$ & $72.9\%$ &$0.4$& $73.3$ & $-$  & $68.7\%$ &$1.3$& $69.1$ & $-$ \\
    ToP & $p=10$ & $s=5$  & $u>0$ &   $70.6\%$ & $0.5$ & $71.2$ & $-$ & $74.3\%$ &$0.3$& $74.7$ & $-$  & $74.7\%$ &$0.4$& $74.9$ & $-$ \\
    ToP & $p=10$ & $s=10$  & $u>0$ &   $72.2\%$ & $0.2$ & $72.5$ & $-$ & $76.2\%$ &$0.2$& $76.8$ & $-$  & $76.0\%$ &$0.3$& $77.0$ & $-$ \\
    \bottomrule
    \end{tabular}
    }
    \end{table*}

%% file: table/time.tex
\begin{table*}[ht!]
    \caption{Running costs ($seconds$) tested on online 3D-BPP with discrete solution space (Section~\ref{subsection:action_result}) and continuous solution space (Section~\ref{subsection:continous}). The running costs of the latter are usually more expensive since checking Constraints~(\ref{eq:non-overlapping}) and~(\ref{eq:containment}) in the continuous domain is more time-consuming.
    } 
    \label{tab:time}
    \centering
    \footnotesize
    \setlength{\tabcolsep}{1.16em}
    \renewcommand{\arraystretch}{0.9}
    \resizebox{1\textwidth}{!}{
    \begin{tabular}{ll|cc|cc|cc}
    \toprule
     & \multirow{2}{*}{Method} &  \multicolumn{2}{c|}{$Setting\,1$} & \multicolumn{2}{c|}{$Setting\,2$} & \multicolumn{2}{c}{$Setting\,3$} \\
     &  &  \multicolumn{1}{c}{Discrete}  & \multicolumn{1}{c|}{Continuous} & \multicolumn{1}{c}{Discrete}  & \multicolumn{1}{c|}{Continuous} & \multicolumn{1}{c}{Discrete}  & \multicolumn{1}{c}{Continuous} \\
    \midrule
    \multirow{8}{*}{\rotatebox[origin=c]{90}{Heuristic}}
     &  $Random$   & $4.59\times 10^{-2}$ & \multicolumn{1}{c|}{$-$}  & $2.03 \times 10^{-2}$ & \multicolumn{1}{c|}{$-$} & $4.62\times 10^{-2}$ &\multicolumn{1}{c}{$-$}  \\
     &  HM  & $4.76\times 10^{-2}$ & \multicolumn{1}{c|}{$-$}  & $3.01\times 10^{-2}$ & \multicolumn{1}{c|}{$-$} & $4.55\times 10^{-2}$ & \multicolumn{1}{c}{$-$}  \\
     &  DBL   & $5.58\times 10^{-2}$ & \multicolumn{1}{c|}{$-$}  & $1.87\times 10^{-2}$ & \multicolumn{1}{c|}{$-$} & $5.44\times 10^{-2}$ & \multicolumn{1}{c}{$-$}  \\
     &  BR   & $1.50\times 10^{-2}$ & $1.69\times 10^{-2}$  & $1.74\times 10^{-2}$ & $1.76\times 10^{-2}$ & $1.42\times 10^{-2}$ & $1.62\times 10^{-2}$  \\
     &  OnlineBPH   & $5.89\times 10^{-3}$ & $1.48\times 10^{-2}$  & $3.39\times 10^{-3}$ & $7.17\times 10^{-3}$ & $4.86\times 10^{-3}$ & $1.38\times 10^{-2}$  \\
     & LSAH    & $1.22\times 10^{-2}$ & $1.44\times 10^{-2}$  & $4.98\times 10^{-3}$ & $7.02\times 10^{-3}$ & $1.14\times 10^{-2}$ & $1.33\times 10^{-2}$  \\
     &  MACS   & $2.68\times 10^{-2}$ & \multicolumn{1}{c|}{$-$}  & $3.00\times 10^{-2}$ & \multicolumn{1}{c|}{$-$} & $2.79\times 10^{-2}$ & \multicolumn{1}{c}{$-$}  \\
    \cmidrule{2-8} 
    \multirow{5}{*}{\rotatebox[origin=c]{90}{DRL}}
     &   CDRL  & $5.51\times 10^{-2}$ & \multicolumn{1}{c|}{$-$}  & $1.33\times 10^{-2}$ & \multicolumn{1}{c|}{$-$} & $3.31\times 10^{-2}$ & \multicolumn{1}{c}{$-$}  \\
     &   PCT \& CP   & $8.43\times 10^{-3}$ & $1.61\times 10^{-2}$  & $7.36\times 10^{-3}$ & $1.52\times 10^{-2}$ & $8.79\times 10^{-3}$ & $1.73\times 10^{-2}$  \\
     &   PCT \& EP   & $1.22\times 10^{-2}$ & $3.73\times 10^{-2}$  & $1.13\times 10^{-2}$ & $1.57\times 10^{-2}$ & $1.25\times 10^{-2}$ & $3.65\times 10^{-2}$  \\
     &   PCT \& EMS   & $1.77\times 10^{-2}$ & $4.11\times 10^{-2}$  & $9.49\times 10^{-3}$ & $2.36\times 10^{-2}$ & $1.80\times 10^{-2}$ & $3.08\times 10^{-2}$  \\
     &   PCT \& EV   & $2.66\times 10^{-2}$ & $4.46\times 10^{-2}$  & $1.25\times 10^{-2}$ & $3.21\times 10^{-2}$ & $2.61\times 10^{-2}$ & $4.38\times 10^{-2}$  \\
     \bottomrule
    \end{tabular}
    }
    \end{table*}

%% file: table/scale.tex
\begin{table*}[h]
    \caption{Scalability on larger packing problems.  $|\textbf{L}|$ is the average number of leaf nodes per step. 
    $|\textbf{L}|$ will not increase exponentially with $|\textbf{B}|$ since invalid leaf nodes will be removed.
    }
    \label{tab:scale}
    \centering
    \footnotesize
    \setlength{\tabcolsep}{0.86em}
    \renewcommand{\arraystretch}{0.9}
    \resizebox{1\textwidth}{!}{
    \begin{tabular}{l|ccccc|ccccc}
        \toprule
        \multirow{2}{*}{Item sizes} & \multicolumn{5}{c|}{Discrete} & \multicolumn{5}{c}{Continuous}\\

        & \multicolumn{1}{c}{$|\textbf{B}|$} & \multicolumn{1}{c}{$|\textbf{L}|$} & \multicolumn{1}{c}{Uti.} &  \multicolumn{1}{c}{Time} & \multicolumn{1}{c|}{$|\textbf{L}| / |\textbf{B}|$} & \multicolumn{1}{c}{$|\textbf{B}|$} & \multicolumn{1}{c}{$|\textbf{L}|$} & \multicolumn{1}{c}{Uti.} &  \multicolumn{1}{c}{Time} & \multicolumn{1}{c}{$|\textbf{L}| / |\textbf{B}|$} \\
        \midrule
        $ [S^d / 10, S^d / 2]$ & 33.0 & 51.5 & $86.0\%$ & $9.5\times10^{-2}$ & 1.6 & 27.0 & 197.5 & $66.3\%$ & $2.4\times10^{-2}$ & 7.3 \\
        $ [S^d / 10, S^d / 5]$ & 241.3 & 67.2 & $81.3\%$ & $9.8\times10^{-3}$ & 0.3 & 185.4 & 956.5 & $61.9\%$ & $3.7\times10^{-2}$ & 5.2 \\
        \bottomrule
    \end{tabular}
    }
\end{table*}

%% file: table/parameter_analysis.tex
\begin{table*}[ht!]
    \caption{\rev{Parameter sensitivity study of the constraint reward weight $c$.}} 
    \label{tab:sensitivity}
    \centering
    \footnotesize
    \resizebox{1\textwidth}{!}{
    \begin{tabular}{l|c|ccc|ccc|ccc}
    \toprule
    \multirow{2}{*}{Constraints} & \multirow{2}{*}{ Weight $c$} &  \multicolumn{3}{c|}{$Setting\,1$} & \multicolumn{3}{c|}{$Setting\,2$} & \multicolumn{3}{c}{$Setting\,3$} \\
    & &  \multicolumn{1}{c}{Obj.} & \multicolumn{1}{c}{Uti. $\uparrow$} & \multicolumn{1}{c|}{Num. $\uparrow$} & \multicolumn{1}{c}{Obj.} & \multicolumn{1}{c}{Uti. $\uparrow$} & \multicolumn{1}{c|}{Num. $\uparrow$} & \multicolumn{1}{c}{Obj.} & \multicolumn{1}{c}{Uti. $\uparrow$} & \multicolumn{1}{c}{Num. $\uparrow$} \\
    \midrule
    \multirow{3}{*}{Isle Friendliness $\downarrow$}
     &  0.1  & 0.14 & \textbf{71.8\%} & \textbf{28.7} & 0.08 & \textbf{85.0\%} & \textbf{32.6} & 0.14 & \textbf{74.2\%} & \textbf{28.6} \\
     &  1.0  & 0.13 & 66.2\% & 26.4 & 0.07 & 75.4\% & 29.0  & 0.13 &  68.5\% & 26.4 \\
     &  10.0 & \textbf{0.10} & 59.5\% & 23.9 & \textbf{0.05} & 63.1\% & 24.1 & \textbf{0.09} & 62.7\% & 24.0 \\
     \midrule
     \multirow{3}{*}{Load Balancing $ \downarrow$}
     &  0.1   & 1.38 & \textbf{70.9\%} & \textbf{27.5}  & 0.68 & \textbf{83.3\%}  & \textbf{32.2} &  0.22 & \textbf{71.1\%} & \textbf{27.7} \\
     &  1.0    & 1.16 & 65.2\% & 25.4 & 0.55 & 75.6\% & 29.2 & 0.18 & 64.5\%
      & 25.2 \\
     &  10.0   & \textbf{0.90} & 59.0\% & 22.9 & \textbf{0.49} & 68.2\% & 26.4 & \textbf{0.14} & 58.5\% & 22.8 \\
     \midrule
     \multirow{3}{*}{Height Uniformity $\downarrow$}
     &  0.1   & 3.74 & \textbf{73.8\%} & \textbf{28.5} & 1.99 & \textbf{83.3\%} & \textbf{32.2} & 3.77 & \textbf{72.8\%} & \textbf{28.2}  \\
     &  1.0   & 2.57 &  73.2\%  & 28.1  & 1.36 & 82.1\% & 31.8 & 2.54 & 71.9\% & 27.9 \\
     &  10.0  & \textbf{2.18} & 64.5\% & 25.0 & \textbf{1.19} & 72.5\% & 28.1 & \textbf{2.24} & 65.4\% & 25.3 \\
     \midrule
     \multirow{3}{*}{Kinematic Constraints $\uparrow$}
     &  0.1  & 0.95 & \textbf{72.7\%} & \textbf{28.2} & 0.96 & \textbf{84.7\%} & \textbf{32.5} & 0.94 & \textbf{74.4\%} & \textbf{28.7} \\
     &  1.0   & 1.03 & 70.2\% & 27.2 & 1.06 & 81.3\% & 31.3 & 1.03 & 71.3\% & 27.6 \\
     &  10.0  & \textbf{1.12} & 68.5\% & 26.6 & \textbf{1.17} & 76.2\% & 29.3 & \textbf{1.11} & 69.2\% & 26.7 \\
     \midrule
     \multirow{3}{*}{Load Bearing Constraints $\downarrow$}
     &  0.1  & 1.40 & \textbf{69.6\%} & \textbf{27.8} & 1.36 & \textbf{80.3\%} & \textbf{31.6} & 0.78 & \textbf{68.4\%} & \textbf{27.2} \\
     &  1.0   & 1.22 & 61.9\% & 24.7 & 1.25 & 74.1\% & 29.0 & 0.72 & 63.1\% & 25.2 \\
     &  10.0  & \textbf{1.13} & 57.2\% & 23.0 & \textbf{1.03} & 62.2\% & 24.5 & \textbf{0.66} & 58.5\% & 23.2 \\
     \midrule
     \multirow{3}{*}{Bridging Constraints $\uparrow$}
     &  0.1  & 1.19 & \textbf{68.8\%} & \textbf{26.8} & 1.30 & \textbf{80.5\%} & \textbf{31.3} & 1.19 & \textbf{68.8\%} & \textbf{26.8} \\
     &  1.0   & 1.23 & 66.2\% & 25.7 & 1.33 & 78.3\% & 30.4 & 1.25 & 65.6\% & 25.6 \\
     &  10.0  & \textbf{1.31}  & 62.0\% & 24.1 & \textbf{1.42} & 70.1\% & 27.0 & \textbf{1.31} & 61.5\% &  24.0
     \\
     \bottomrule
\end{tabular}
}
\vspace{-4pt}
\end{table*}

%% file: PCT_IJRR.bbl
\begin{thebibliography}{74}
\providecommand{\natexlab}[1]{#1}
\providecommand{\url}[1]{\texttt{#1}}
\providecommand{\urlprefix}{URL }
\expandafter\ifx\csname urlstyle\endcsname\relax
  \providecommand{\doi}[1]{DOI:\discretionary{}{}{}#1}\else
  \providecommand{\doi}{DOI:\discretionary{}{}{}\begingroup
  \urlstyle{rm}\Url}\fi

\bibitem[{Bello et~al.(2017)Bello, Pham, Le, Norouzi and Bengio}]{BelloPL0B17}
Bello I, Pham H, Le QV, Norouzi M and Bengio S (2017) Neural combinatorial
  optimization with reinforcement learning.
\newblock In: \emph{International Conference on Learning Representations}.

\bibitem[{Boyd et~al.(2004)Boyd, Boyd and Vandenberghe}]{boyd2004convex}
Boyd S, Boyd SP and Vandenberghe L (2004) \emph{Convex optimization}.
\newblock Cambridge university press.

\bibitem[{Bruna et~al.(2014)Bruna, Zaremba, Szlam and LeCun}]{BrunaZSL13}
Bruna J, Zaremba W, Szlam A and LeCun Y (2014) Spectral networks and locally
  connected networks on graphs.
\newblock In: \emph{International Conference on Learning Representations}.

\bibitem[{Burke et~al.(2007)Burke, Hellier, Kendall and
  Whitwell}]{burke2007complete}
Burke EK, Hellier RS, Kendall G and Whitwell G (2007) Complete and robust
  no-fit polygon generation for the irregular stock cutting problem.
\newblock \emph{European Journal of Operational Research} .

\bibitem[{Choset et~al.(2005)Choset, Lynch, Hutchinson, Kantor and
  Burgard}]{choset2005principles}
Choset H, Lynch KM, Hutchinson S, Kantor GA and Burgard W (2005)
  \emph{Principles of robot motion: theory, algorithms, and implementations}.
\newblock MIT press.

\bibitem[{Coumans and Bai(2016)}]{coumans2016pybullet}
Coumans E and Bai Y (2016) Pybullet, a python module for physics simulation for
  games, robotics and machine learning.
\newblock \emph{URL http://pybullet.org} .

\bibitem[{Crainic et~al.(2008)Crainic, Perboli and Tadei}]{CrainicPT08}
Crainic TG, Perboli G and Tadei R (2008) Extreme point-based heuristics for
  three-dimensional bin packing.
\newblock \emph{{INFORMS} Journal on Computing} .

\bibitem[{Crainic et~al.(2009)Crainic, Perboli and Tadei}]{CrainicPT09}
Crainic TG, Perboli G and Tadei R (2009) Ts2pack: {A} two-level tabu search for
  the three-dimensional bin packing problem.
\newblock \emph{European Journal of Operational Research} .

\bibitem[{De~Castro~Silva et~al.(2003)De~Castro~Silva, Soma and
  Maculan}]{de2003greedy}
De~Castro~Silva J, Soma N and Maculan N (2003) A greedy search for the
  three-dimensional bin packing problem: the packing static stability case.
\newblock \emph{International Transactions in Operational Research} .

\bibitem[{Dekel et~al.(2020)Dekel, Harenstam-Nielsen and
  Caccamo}]{dekel2020optimal}
Dekel A, Harenstam-Nielsen L and Caccamo S (2020) Optimal least-squares
  solution to the hand-eye calibration problem.
\newblock In: \emph{IEEE Conference on Computer Vision and Pattern
  Recognition}.

\bibitem[{Demisse et~al.(2012)Demisse, Mihalyi, Okal, Poudel, Schauer and
  N{\"u}chter}]{demisse2012mixed}
Demisse G, Mihalyi R, Okal B, Poudel D, Schauer J and N{\"u}chter A (2012)
  Mixed palletizing and task completion for virtual warehouses.
\newblock In: \emph{Virtual Manufacturing and Automation Competition Workshop
  at the International Conference of Robotics and Automation}.

\bibitem[{Dijkstra(2022)}]{dijkstra2022note}
Dijkstra EW (2022) A note on two problems in connexion with graphs.
\newblock In: \emph{Edsger Wybe Dijkstra: His Life, Work, and Legacy}.

\bibitem[{Duan et~al.(2019)Duan, Hu, Qian, Gong, Zhang, Wei and
  Xu}]{DuanHQGZWX19}
Duan L, Hu H, Qian Y, Gong Y, Zhang X, Wei J and Xu Y (2019) A multi-task
  selected learning approach for solving {3D} flexible bin packing problem.
\newblock In: \emph{International Conference on Autonomous Agents and
  MultiAgent Systems}.

\bibitem[{Egeblad et~al.(2007)Egeblad, Nielsen and Odgaard}]{egeblad2007fast}
Egeblad J, Nielsen BK and Odgaard A (2007) Fast neighborhood search for two-and
  three-dimensional nesting problems.
\newblock \emph{European Journal of Operational Research} .

\bibitem[{Faroe et~al.(2003)Faroe, Pisinger and Zachariasen}]{FaroePZ03}
Faroe O, Pisinger D and Zachariasen M (2003) Guided local search for the
  three-dimensional bin-packing problem.
\newblock \emph{{INFORMS} Journal on Computing} .

\bibitem[{Firoozi et~al.(2025)Firoozi, Tucker, Tian, Majumdar, Sun, Liu, Zhu,
  Song, Kapoor, Hausman, Ichter, Driess, Wu, Lu and
  Schwager}]{Foundation_model}
Firoozi R, Tucker J, Tian S, Majumdar A, Sun J, Liu W, Zhu Y, Song S, Kapoor A,
  Hausman K, Ichter B, Driess D, Wu J, Lu C and Schwager M (2025) Foundation
  models in robotics: Applications, challenges, and the future.
\newblock \emph{The International Journal of Robotics Research} .

\bibitem[{Gori et~al.(2005)Gori, Monfardini and Scarselli}]{1555942}
Gori M, Monfardini G and Scarselli F (2005) A new model for learning in graph
  domains.
\newblock In: \emph{Proceedings. 2005 IEEE International Joint Conference on
  Neural Networks, 2005.}

\bibitem[{G{\"o}rner et~al.(2019)G{\"o}rner, Haschke, Ritter and
  Zhang}]{gorner2019moveit}
G{\"o}rner M, Haschke R, Ritter H and Zhang J (2019) Moveit! task constructor
  for task-level motion planning.
\newblock In: \emph{International Conference on Robotics and Automation}.

\bibitem[{Grove(1995)}]{grove1995online}
Grove EF (1995) Online bin packing with lookahead.
\newblock In: \emph{Annual ACM-SIAM Symposium on Discrete Algorithms}.

\bibitem[{Gzara et~al.(2020)Gzara, Elhedhli and Yildiz}]{GzaraEY20}
Gzara F, Elhedhli S and Yildiz BC (2020) The pallet loading problem:
  Three-dimensional bin packing with practical constraints.
\newblock \emph{European Journal of Operational Research} .

\bibitem[{Ha et~al.(2017)Ha, Nguyen, Bui and Wang}]{ha2017online}
Ha CT, Nguyen TT, Bui LT and Wang R (2017) An online packing heuristic for the
  three-dimensional container loading problem in dynamic environments and the
  physical internet.
\newblock In: \emph{Applications of Evolutionary Computation}.

\bibitem[{Haarnoja et~al.(2018)Haarnoja, Zhou, Abbeel and
  Levine}]{HaarnojaZAL18}
Haarnoja T, Zhou A, Abbeel P and Levine S (2018) Soft actor-critic: Off-policy
  maximum entropy deep reinforcement learning with a stochastic actor.
\newblock In: \emph{International Conference on Machine Learning}.

\bibitem[{Hart et~al.(1968)Hart, Nilsson and Raphael}]{hart1968formal}
Hart PE, Nilsson NJ and Raphael B (1968) A formal basis for the heuristic
  determination of minimum cost paths.
\newblock \emph{Transactions on Systems Science and Cybernetics} .

\bibitem[{He et~al.(2017)He, Gkioxari, Doll{\'{a}}r and Girshick}]{HeGDG17}
He K, Gkioxari G, Doll{\'{a}}r P and Girshick RB (2017) Mask {R-CNN}.
\newblock In: \emph{{IEEE} International Conference on Computer Vision}.

\bibitem[{Hof et~al.(2005)Hof, Gazendam and Sinke}]{hof2005condition}
Hof AL, Gazendam M and Sinke W (2005) The condition for dynamic stability.
\newblock \emph{Journal of Biomechanics} .

\bibitem[{Hong et~al.(2020)Hong, Kim and Lee}]{HongKL20}
Hong Y, Kim Y and Lee K (2020) Smart pack: Online autonomous object-packing
  system using {RGB-D} sensor data.
\newblock \emph{Sensors} .

\bibitem[{Hu et~al.(2017)Hu, Zhang, Yan, Wang and Xu}]{hu2017solving}
Hu H, Zhang X, Yan X, Wang L and Xu Y (2017) Solving a new {3D} bin packing
  problem with deep reinforcement learning method.
\newblock \emph{arXiv preprint arXiv:1708.05930} .

\bibitem[{Hu et~al.(2020)Hu, Xu, Chen, Gong, Zhang and Huang}]{HuXCG0020}
Hu R, Xu J, Chen B, Gong M, Zhang H and Huang H (2020) Tap-net:
  transport-and-pack using reinforcement learning.
\newblock \emph{{ACM} Transactions on Graphics} .

\bibitem[{Kagerer et~al.(2023)Kagerer, Beinhofer, Stricker and
  Nüchter}]{BED-BPP}
Kagerer F, Beinhofer M, Stricker S and Nüchter A (2023) Bed-bpp: Benchmarking
  dataset for robotic bin packing problems.
\newblock \emph{The International Journal of Robotics Research} .

\bibitem[{Kang et~al.(2012)Kang, Moon and Wang}]{KangMW12}
Kang K, Moon I and Wang H (2012) A hybrid genetic algorithm with a new packing
  strategy for the three-dimensional bin packing problem.
\newblock \emph{Applied Mathematics and Computation} .

\bibitem[{Karabulut and Inceoglu(2004)}]{KarabulutI04}
Karabulut K and Inceoglu MM (2004) A hybrid genetic algorithm for packing in
  {3D} with deepest bottom left with fill method.
\newblock In: \emph{Advances in Information Systems}.

\bibitem[{Kool et~al.(2019)Kool, van Hoof and Welling}]{KoolHW19}
Kool W, van Hoof H and Welling M (2019) Attention, learn to solve routing
  problems!
\newblock In: \emph{International Conference on Learning Representations}.

\bibitem[{Korf(1998)}]{korf1998complete}
Korf RE (1998) A complete anytime algorithm for number partitioning.
\newblock \emph{Artificial Intelligence} .

\bibitem[{Li et~al.(2022)Li, Gu, Wang, Ren and Lau}]{li2022one}
Li D, Gu Z, Wang Y, Ren C and Lau FC (2022) One model packs thousands of items
  with recurrent conditional query learning.
\newblock \emph{Knowledge-Based Systems} .

\bibitem[{Li et~al.(2025)Li, Zhao, Yu, Du, Zou, Hu and Xu}]{li2025pinwm}
Li W, Zhao H, Yu Z, Du Y, Zou Q, Hu R and Xu K (2025) Pin-wm: Learning
  physics-informed world models for non-prehensile manipulation.
\newblock In: \emph{Robotics: Science and Systems}.

\bibitem[{Ljung(1998)}]{ljung1998system}
Ljung L (1998) System identification.
\newblock In: \emph{Signal Analysis and Prediction}.

\bibitem[{{LLC Gurobi Optimization}(2018)}]{gurobi2018}
{LLC Gurobi Optimization} (2018) Gurobi optimizer reference manual.
\newblock \url{https://tinyurl.com/4uxkx7nw}.

\bibitem[{Makoviychuk et~al.(2021)Makoviychuk, Wawrzyniak, Guo, Lu, Storey,
  Macklin, Hoeller, Rudin, Allshire, Handa and State}]{makoviychuk2021isaac}
Makoviychuk V, Wawrzyniak L, Guo Y, Lu M, Storey K, Macklin M, Hoeller D, Rudin
  N, Allshire A, Handa A and State G (2021) Isaac gym: High performance
  gpu-based physics simulation for robot learning.
\newblock \emph{arXiv preprint arXiv:2108.10470} .

\bibitem[{Mark et~al.(2008)Mark, Otfried, Marc and
  Mark}]{mark2008computational}
Mark dB, Otfried C, Marc vK and Mark O (2008) \emph{Computational geometry
  algorithms and applications}.
\newblock Spinger.

\bibitem[{Martello et~al.(2000)Martello, Pisinger and Vigo}]{MartelloPV00}
Martello S, Pisinger D and Vigo D (2000) The three-dimensional bin packing
  problem.
\newblock \emph{Operations Research} .

\bibitem[{Martello et~al.(2007)Martello, Pisinger, Vigo, Boef and
  Korst}]{martello2007algorithm}
Martello S, Pisinger D, Vigo D, Boef ED and Korst J (2007) Algorithm 864:
  General and robot-packable variants of the three-dimensional bin packing
  problem.
\newblock \emph{ACM Transactions on Mathematical Software} .

\bibitem[{Mayne et~al.(2000)Mayne, Rawlings, Rao and
  Scokaert}]{mayne2000constrained}
Mayne DQ, Rawlings JB, Rao CV and Scokaert PO (2000) Constrained model
  predictive control: Stability and optimality.
\newblock \emph{Automatica} .

\bibitem[{Newell et~al.(1959)Newell, Shaw and Simon}]{newell1959report}
Newell A, Shaw JC and Simon HA (1959) Report on a general problem solving
  program.
\newblock In: \emph{IFIP Congress}.

\bibitem[{Page(1981)}]{page1981biaxial}
Page A (1981) The biaxial compressive strength of brick masonry.
\newblock \emph{Proceedings of the Institution of Civil Engineers} .

\bibitem[{Pan et~al.(2023)Pan, Chen and Lin}]{pan2023adjustable}
Pan Y, Chen Y and Lin F (2023) Adjustable robust reinforcement learning for
  online {3D} bin packing.
\newblock \emph{Advances in Neural Information Processing Systems} .

\bibitem[{Puche and Lee(2022)}]{puche2022online}
Puche AV and Lee S (2022) Online {3D} bin packing reinforcement learning
  solution with buffer.
\newblock In: \emph{IEEE International Conference on Intelligent Robots and
  Systems}.

\bibitem[{Qi et~al.(2017)Qi, Su, Mo and Guibas}]{QiSMG17}
Qi CR, Su H, Mo K and Guibas LJ (2017) Pointnet: Deep learning on point sets
  for {3D} classification and segmentation.
\newblock In: \emph{{IEEE} Conference on Computer Vision and Pattern
  Recognition}.

\bibitem[{Qiu et~al.(2022)Qiu, Sun and Yang}]{qiu2022dimes}
Qiu R, Sun Z and Yang Y (2022) Dimes: A differentiable meta solver for
  combinatorial optimization problems.
\newblock \emph{Advances in Neural Information Processing Systems} 35.

\bibitem[{Ramos et~al.(2016)Ramos, Oliveira and Lopes}]{RamosOL16}
Ramos AG, Oliveira JF and Lopes MP (2016) A physical packing sequence algorithm
  for the container loading problem with static mechanical equilibrium
  conditions.
\newblock \emph{International Transactions in Operational Research} .

\bibitem[{Seiden(2002)}]{Seiden02}
Seiden SS (2002) On the online bin packing problem.
\newblock \emph{Journal of the ACM} .

\bibitem[{Shin et~al.(2016)Shin, Porst, Vouga, Ochsendorf and
  Durand}]{shin2016reconciling}
Shin HV, Porst CF, Vouga E, Ochsendorf J and Durand F (2016) Reconciling
  elastic and equilibrium methods for static analysis.
\newblock \emph{ACM Transactions on Graphics} .

\bibitem[{Silver et~al.(2016)Silver, Huang, Maddison, Guez, Sifre, van~den
  Driessche, Schrittwieser, Antonoglou, Panneershelvam, Lanctot, Dieleman,
  Grewe, Nham, Kalchbrenner, Sutskever, Lillicrap, Leach, Kavukcuoglu, Graepel
  and Hassabis}]{silver2016mastering}
Silver D, Huang A, Maddison CJ, Guez A, Sifre L, van~den Driessche G,
  Schrittwieser J, Antonoglou I, Panneershelvam V, Lanctot M, Dieleman S, Grewe
  D, Nham J, Kalchbrenner N, Sutskever I, Lillicrap TP, Leach M, Kavukcuoglu K,
  Graepel T and Hassabis D (2016) Mastering the game of {Go} with deep neural
  networks and tree search.
\newblock \emph{Nature} .

\bibitem[{Sim4Dexterity(2023)}]{icra2023stacking}
Sim4Dexterity (2023) {ICRA} 2023 virtual manipulation challenge: Stacking.
\newblock \url{https://tinyurl.com/4asfdnex}.

\bibitem[{Spaan(2012)}]{spaan2012partially}
Spaan MT (2012) Partially observable markov decision processes.
\newblock In: \emph{Reinforcement Learning: State-of-the-Art}. Springer.

\bibitem[{Sutton and Barto(2018)}]{sutton2018reinforcement}
Sutton RS and Barto AG (2018) \emph{Reinforcement Learning: An Introduction}.
\newblock MIT press.

\bibitem[{Taylor and Stone(2009)}]{TaylorS09}
Taylor ME and Stone P (2009) Transfer learning for reinforcement learning
  domains: {A} survey.
\newblock \emph{Journal of Machine Learning Research} .

\bibitem[{Vaswani et~al.(2017)Vaswani, Shazeer, Parmar, Uszkoreit, Jones,
  Gomez, Kaiser and Polosukhin}]{VaswaniSPUJGKP17}
Vaswani A, Shazeer N, Parmar N, Uszkoreit J, Jones L, Gomez AN, Kaiser L and
  Polosukhin I (2017) Attention is all you need.
\newblock In: \emph{Advances in Neural Information Processing Systems}.

\bibitem[{Velickovic et~al.(2018)Velickovic, Cucurull, Casanova, Romero,
  Li{\`{o}} and Bengio}]{VelickovicCCRLB18}
Velickovic P, Cucurull G, Casanova A, Romero A, Li{\`{o}} P and Bengio Y (2018)
  Graph attention networks.
\newblock In: \emph{International Conference on Learning Representations}.

\bibitem[{Verma et~al.(2020)Verma, Singhal, Khadilkar, Basumatary, Nayak,
  Singh, Kumar and Sinha}]{Generalized}
Verma R, Singhal A, Khadilkar H, Basumatary A, Nayak S, Singh HV, Kumar S and
  Sinha R (2020) A generalized reinforcement learning algorithm for online {3D}
  bin-packing.
\newblock \emph{arXiv preprint arXiv:2007.00463} .

\bibitem[{Vinyals et~al.(2015)Vinyals, Fortunato and Jaitly}]{VinyalsFJ15}
Vinyals O, Fortunato M and Jaitly N (2015) Pointer networks.
\newblock In: \emph{Advances in Neural Information Processing Systems}.

\bibitem[{Wang and Hauser(2019{\natexlab{a}})}]{WangH19}
Wang F and Hauser K (2019{\natexlab{a}}) Robot packing with known items and
  nondeterministic arrival order.
\newblock In: \emph{Robotics: Science and Systems}.

\bibitem[{Wang and Hauser(2019{\natexlab{b}})}]{WangH19a}
Wang F and Hauser K (2019{\natexlab{b}}) Stable bin packing of non-convex {3D}
  objects with a robot manipulator.
\newblock In: \emph{International Conference on Robotics and Automation}.

\bibitem[{Wang and Hauser(2021)}]{wang2021dense}
Wang F and Hauser K (2021) Dense robotic packing of irregular and novel {3D}
  objects.
\newblock \emph{IEEE Transactions on Robotics} .

\bibitem[{Wu et~al.(2017)Wu, Mansimov, Grosse, Liao and Ba}]{wu2017scalable}
Wu Y, Mansimov E, Grosse RB, Liao S and Ba J (2017) Scalable trust-region
  method for deep reinforcement learning using kronecker-factored
  approximation.
\newblock In: \emph{Advances in Neural Information Processing Systems}.

\bibitem[{Xu et~al.(2023)Xu, Gong, Zhang, Huang and Hu}]{xu2023neural}
Xu J, Gong M, Zhang H, Huang H and Hu R (2023) Neural packing: from visual
  sensing to reinforcement learning.
\newblock \emph{ACM Transactions on Graphics} .

\bibitem[{Yang et~al.(2021{\natexlab{a}})Yang, Zeynali, Hajiesmaili, Sitaraman
  and Towsley}]{yang2021competitive}
Yang L, Zeynali A, Hajiesmaili MH, Sitaraman RK and Towsley D
  (2021{\natexlab{a}}) Competitive algorithms for online multidimensional
  knapsack problems.
\newblock \emph{Proceedings of the ACM on Measurement and Analysis of Computing
  Systems} .

\bibitem[{Yang et~al.(2021{\natexlab{b}})Yang, Yang, Song, Zhang, Song, Cheng
  and Li}]{yang2021packerbot}
Yang Z, Yang S, Song S, Zhang W, Song R, Cheng J and Li Y (2021{\natexlab{b}})
  Packerbot: Variable-sized product packing with heuristic deep reinforcement
  learning.
\newblock In: \emph{International Conference on Intelligent Robots and
  Systems}.

\bibitem[{Yuan et~al.(2023)Yuan, Zhang, Cai and Yan}]{yuan2023towards}
Yuan J, Zhang J, Cai Z and Yan J (2023) Towards variance reduction for
  reinforcement learning of industrial decision-making tasks: A bi-critic based
  demand-constraint decoupling approach.
\newblock In: \emph{ACM SIGKDD Conference on Knowledge Discovery and Data
  Mining}.

\bibitem[{Zhang et~al.(2021)Zhang, Zi and Ge}]{Attend2Pack}
Zhang J, Zi B and Ge X (2021) Attend2pack: Bin packing through deep
  reinforcement learning with attention.
\newblock \emph{arXiv preprint arXiv:2107.04333} .

\bibitem[{Zhao et~al.(2023)Zhao, Pan, Yu and Xu}]{zhao2023learning}
Zhao H, Pan Z, Yu Y and Xu K (2023) Learning physically realizable skills for
  online packing of general {3D} shapes.
\newblock \emph{ACM Transactions on Graphics} .

\bibitem[{Zhao et~al.(2021)Zhao, She, Zhu, Yang and Xu}]{ZhaoS0Y021}
Zhao H, She Q, Zhu C, Yang Y and Xu K (2021) Online {3D} bin packing with
  constrained deep reinforcement learning.
\newblock In: \emph{{AAAI} Conference on Artificial Intelligence}.

\bibitem[{Zhao et~al.(2024)Zhao, Yu, Huang, Yi, Zhu and Xu}]{yu2024disco}
Zhao H, Yu K, Huang Y, Yi R, Zhu C and Xu K (2024) Disco: Efficient diffusion
  solver for large-scale combinatorial optimization problems.
\newblock \emph{Graphical Models} .

\bibitem[{Zhao et~al.(2022{\natexlab{a}})Zhao, Yu and Xu}]{zhao2022learning}
Zhao H, Yu Y and Xu K (2022{\natexlab{a}}) Learning efficient online {3D} bin
  packing on packing configuration trees.
\newblock In: \emph{International Conference on Learning Representations}.

\bibitem[{Zhao et~al.(2022{\natexlab{b}})Zhao, Zhu, Xu, Huang and
  Xu}]{zhao2021learning}
Zhao H, Zhu C, Xu X, Huang H and Xu K (2022{\natexlab{b}}) Learning practically
  feasible policies for online {3D} bin packing.
\newblock \emph{Science China Information Sciences} .

\end{thebibliography}
